\documentclass{article}

\usepackage[numbers,sort&compress]{natbib}

     \usepackage[final]{neurips_2020}

\usepackage[utf8]{inputenc} 
\usepackage[T1]{fontenc}    
\usepackage{hyperref}       
\usepackage{url}            
\usepackage{booktabs}       
\usepackage{amsfonts}       
\usepackage{nicefrac}       
\usepackage{microtype}      

\usepackage{amsmath}
\usepackage{amssymb}
\usepackage{epsfig}
\usepackage{graphicx}
\usepackage{subcaption}
\usepackage{mathtools}
\usepackage{array}
\usepackage{boldline , multirow}
\usepackage{comment}        
\usepackage{times}
\usepackage{appendix}
\usepackage{bbm}
\usepackage{algorithm}
\usepackage{algpseudocode}
\usepackage{varwidth}
\usepackage{enumitem}
\usepackage{adjustbox}
\usepackage[outline]{contour}
\usepackage{multicol}
\usepackage{etoolbox}\AtBeginEnvironment{algorithmic}{\small}
\usepackage[labelformat=simple]{subcaption}

\title{Improving model calibration with accuracy versus uncertainty optimization}

\author{%
  Ranganath Krishnan \\
  Intel Labs\\
  \texttt{ranganath.krishnan@intel.com} \\
  \And
  Omesh Tickoo \\
  Intel Labs\\
  \texttt{omesh.tickoo@intel.com} \\
}

\begin{document}

\maketitle

\begin{abstract}
  Obtaining reliable and accurate quantification of uncertainty estimates from deep neural networks is important in safety-critical applications. A well-calibrated model should be accurate when it is certain about its prediction and indicate high uncertainty when it is likely to be inaccurate. Uncertainty calibration is a challenging problem as there is no ground truth available for uncertainty estimates. We propose an optimization method that leverages the relationship between accuracy and uncertainty as an anchor for uncertainty calibration. We introduce a differentiable \textit{accuracy versus uncertainty calibration} (AvUC) loss function that allows a model to learn to provide well-calibrated uncertainties, in addition to improved accuracy. We also demonstrate the same methodology can be extended to post-hoc uncertainty calibration on pretrained models. We illustrate our approach with mean-field stochastic variational inference and compare with state-of-the-art methods. Extensive experiments demonstrate our approach yields better model calibration than existing methods on large-scale image classification tasks under distributional shift.
  
\end{abstract}

\section{Introduction}
Probabilistic deep neural networks (DNNs) enable quantification of principled uncertainty estimates, which are essential to understand the model predictions for reliable decision making in safety critical applications~\cite{ghahramani2015probabilistic}. In addition to obtaining accurate predictions from the model, it is important for the model to indicate when it is likely to make incorrect predictions. Various probabilistic methods have been proposed to capture uncertainty estimates from DNNs including Bayesian~\cite{graves2011practical,blundell2015weight,kingma2015variational,gal2016dropout,maddox2019simple,rohekar2019modeling,farquhar_radial_2020} and non-Bayesian~\cite{lakshminarayanan2017simple,lee2018simple} formulations. In spite of recent advances in probabilistic deep learning to improve model robustness, obtaining accurate quantification of uncertainty estimates from DNNs is still an open research problem. A well-calibrated model should be confident about its predictions when it is accurate and indicate high uncertainty when making inaccurate predictions. Modern neural networks are poorly calibrated~\cite{guo2017calibration, kumar2018trainable} as they tend to be overconfident on incorrect predictions. Negative log-likelihood (NLL) loss is conventionally used for training the neural networks in multi-class classification tasks. Miscalibration in DNNs has been linked to overfitting of NLL~\cite{guo2017calibration,mukhoti2020calibrating}. 
Probabilistic DNNs fail to provide calibrated uncertainty in between separated regions of observations due to model misspecification and the use of approximate inference~\cite{kuleshov18accurate,foong2019between,heekwell}. Overcoming the problem of poor calibration in modern neural networks is an active area of research~\cite{guo2017calibration,kumar2018trainable,kuleshov18accurate,kumar2019verified,kull2019beyond,thulasidasan2019mixup,hendrycks2019robustness,hendrycks*2020augmix,foong2019between,mukhoti2020calibrating,heekwell}. 

In real-world settings, the observed data distribution may shift from training distribution (dataset shift~\cite{moreno2012unifying}) and there are possibilities of observing novel inputs that are far-off from training data manifold (out-of-distribution). DNN model predictions have been shown to be unreliable under such distributional shift ~\cite{alcorn2019strike,blum2019fishyscapes,hendrycks2019robustness}. Obtaining reliable uncertainties even under distributional shift is important to build robust AI systems for successful deployment in the real-world~\cite{amodei2016concrete,snoek2019can}. Uncertainty calibration will also help in detecting distributional shift to caution AI practitioners, as well-calibrated uncertainty estimates can guide when to trust and when not to trust the model predictions. But uncertainty calibration is a challenging problem due to the unavailability of ground truth uncertainty estimates.

\paragraph{Contribution}In this paper, we introduce the \textit{accuracy versus uncertainty calibration} (AvUC) loss function for probabilistic deep neural networks to derive models that will be confident on accurate predictions and indicate higher uncertainty when likely to be inaccurate. We rely on theoretically sound loss-calibrated approximate inference framework~\cite{lacoste2011approximate,cobb2018loss} with AvUC loss as utilty-dependent penalty term for the task of obtaining well-calibrated uncertainties along with improved accuracy. 
We find that accounting for predictive uncertainty while training the neural network improves model calibration. To evaluate model calibration under dataset shift, we  use various image perturbations and corruptions at different shift intensities~\cite{hendrycks2019robustness} and compare with high-performing baselines provided in uncertainty quantification(UQ) benchmark~\cite{snoek2019can}. In summary, we make the following contributions in this work:


\begin{itemize}[leftmargin=0.5cm]
	\item Propose an optimization method that leverages the relationship between accuracy and uncertainty as anchor for uncertainty calibration while training deep neural network classifiers (Bayesian and non-Bayesian). We introduce differentiable proxy for \textit{Accuracy versus Uncertainty} (AvU) measure and the corresponding \textit{accuracy versus uncertainty calibration} (AvUC) loss function devised to obtain well-calibrated uncertainties, while maintaining or improving model accuracy.
	\item Investigate accounting for predictive uncertainty estimation in the training objective function and its effect on model calibration under distributional shift (dataset shift and out-of-distribution).
	\item Propose a post-hoc model calibration method extending the temperature scaling using AvUC loss.
	\item Empirically evaluate the proposed methods and compare with existing high-performing baselines on large-scale image classification tasks using a wide range of metrics. We demonstrate our method yields state-of-the-art model calibration under distributional shift. We also compare the distributional shift detection performance using predictive uncertainty estimates obtained from different methods.
\end{itemize}


\section{Background}
\label{sec:background}

\paragraph{Related work} Calibration of deep neural networks involves accurately representing predictive probabilities with respect to true likelihood. Existing research to achieve model calibration and robustness in DNNs for multiclass classification tasks can be broadly classified into three categories (i) post-processing calibration 
(ii) training the model with data augmentation for better representation of training data 
(iii) probabilistic methods with Bayesian and non-Bayesian formulation for DNNs towards better representation of model parameters
. Post-hoc calibration includes temperature scaling~\cite{guo2017calibration} and dirichlet calibration~\cite{kull2019beyond}. Though post-processing method like temperature scaling perform well under in-distribution conditions, the calibration on the i.i.d. validation dataset does not guarantee calibration under distributional shift~\cite{snoek2019can}. Also they push the accurate predictions to low confidence regions~\cite{kumar2018trainable}. Data augmentation methods include Mixup~\cite{thulasidasan2019mixup} and AugMix~\cite{hendrycks*2020augmix}. Though data augmentation methods improve model robustness, it is practically difficult to introduce a wide spectrum of perturbations and corruptions during training that comprehensively represents the real-world deployment conditions. Deep-ensembles~\cite{lakshminarayanan2017simple} propose a non-Bayesian approach of training an ensemble of neural networks from different random initializations that has been shown to provide calibrated confidence~\cite{snoek2019can}.  However, Ensembles introduce additional overhead of training multiple models and significant memory complexity during test time. 
Approximate Bayesian inference methods for DNNs have been proposed as computing true posterior  is intractable, the methods include variational inference~\cite{graves2011practical,blundell2015weight,kingma2015variational}, stochastic gradient variants of MCMC~\cite{welling2011bayesian,chen2014stochastic}, Monte Carlo dropout~\cite{gal2016dropout} and SWAG~\cite{maddox2019simple}.  Approximate Bayesian inference methods are promising, but they may fail to provide calibrated uncertainty in between separated regions of observations as they tend to fit an approximation to a local mode and do not capture the complete true posterior~\cite{smith2018understanding,lakshminarayanan2017simple,heekwell,foong2019between}. This may cause the model to be overconfident under distributional shift.  Trainable calibration measures~\cite{kumar2018trainable} have been proposed that encourage confidence calibration during training by optimizing maximum mean calibration error.~\citet{snoek2019can} show the model calibration degrades with data shift for many of the existing methods that perform well under in-distribution conditions and provides a benchmark evaluating model calibration under data shift. Existing calibration methods do not explicitly account for the quality of predictive uncertainty estimates while training the model, or post-hoc calibration.



\paragraph{Uncertainty estimation}There are two types of uncertainties that constitute predictive uncertainty of models: \textit{aleatoric uncertainty} and \textit{epistemic uncertainty}~\cite{der2009aleatory,kendall2017uncertainties}. Aleatoric uncertainty captures noise inherent with the observation. Epistemic uncertainty captures the lack of knowledge in representing model parameters. 
Probabilistic DNNs can quantify both aleatoric and epistemic uncertainties, but deterministic DNNs can capture only aleatoric uncertainty. Various metrics have been proposed to quantify these uncertainties in classification tasks~\cite{gal2016uncertainty, smith2018understanding}, including predictive entropy~\cite{shannon1948mathematical}, variation ratio~\cite{freeman1965} and mutual information~\cite{shannon1948mathematical,houlsby2011bayesian}. These metrics are rooted with principled foundations in information theory and statistics. In this paper, we use predictive entropy as the uncertainty metric, which represents the predictive uncertainty of the model and captures combination of both epistemic and aleatoric uncertainties~\cite{mukhoti2018evaluating} in probabilistic models. We will use mean-field stochastic variational inference (SVI)~\cite{blundell2015weight, graves2011practical} in Bayesian neural networks to illustrate our proposed methods, we refer to Appendix~\ref{appdx:background} for background on SVI and uncertainty metrics.

\vspace{-3mm}
\paragraph{Loss-calibrated approximate inference} Bayesian decision theory~\cite{berger1985statistical} offers a theoretical framework for decision making under uncertainty about a parameter $\theta$. 
Loss-calibrated approximate inference~\cite{lacoste2011approximate,cobb2018loss} framework is built upon the basis of Bayesian decision theory to yield optimal predictions for a specific task incorporating a utility function $\mathrm{U(\theta,a)}$, which informs us the utility of taking action $\mathrm{a \in}\mathcal{A}$. The goal of accomplishing the specific task is defined by the utility function which guides the model learning. \citet{cobb2018loss} derived a loss-calibrated evidence lower bound comprising of standard evidence lower bound and an additional utility-dependent regularization term.
 
\vspace{-3mm}
\paragraph{Evaluation metrics} 
We use various metrics{\footnote[1]{Arrows next to each evaluation metric indicate which direction is better. Equations in Appendix~\ref{appdx:evalmetrics}}} to evaluate proposed methods and compare with high-performing Bayesian and non-Bayesian methods under distributional shift.  \textit{Expected calibration error} ({ECE})\contour{black}{$\downarrow$}~\cite{naeini2015obtaining} is popularly used for determining the calibration of DNNs, which represents the difference in expectation between model accuracy and its confidence. Recently, \textit{expected uncertainty calibration error} ({UCE})\contour{black}{$\downarrow$}~\cite{laves2019well}  has been proposed to measure miscalibration of uncertainty, which represents the difference in expectation between model error and its uncertainty. Model calibration is also measured using proper scoring rules~\cite{gneiting2007strictly} such as  \textit{negative log likelihood} ({NLL})\contour{black}{$\downarrow$} and  {\textit{Brier}} score\contour{black}{$\downarrow$}~\cite{brier1950verification}. The benefits and drawbacks of each of these metrics are described in~\cite{snoek2019can}. Conditional probabilities {\textit{p(\text{accurate }|\text{ certain})}}\contour{black}{$\uparrow$} and {\textit{p(\text{uncertain }|\text { inaccurate})}}\contour{black}{$\uparrow$}~\cite{mukhoti2018evaluating} have been proposed as model performance evaluation metrics for comparing the quality of uncertainty estimates obtained from different probabilistic methods. \textit{p(\text{accurate }|\text{ certain })} represents the probability that the model is accurate on its output given that it is certain about its predictions, and \textit{p(\text{uncertain }|\text { inaccurate })} represents the probability that the model is uncertain about its output given that it has made inaccurate prediction. We also use \textit{area under the receiver operating characteristic curve} (AUROC)\contour{black}{$\uparrow$}~\cite{davis2006relationship} and \textit{area under the precision-recall curve} (AUPR)\contour{black}{$\uparrow$}~\cite{saito2015precision} for measuring the distributional shift detection performance, which are typically used for evaluating out-of-distribution detection.

\vspace{-2mm}
\section{Obtaining well-calibrated uncertainties with AvUC loss}
\label{sec:avu_loss}
For evaluating uncertainty estimates from different methods,~\citet{mukhoti2018evaluating} had proposed patch accuracy versus patch uncertainty (PAvPU) metric that can be computed per image on semantic segmentation task. Their evaluation methodology was designed based on the assumptions that if a model is confident about its prediction, it
should be accurate on the same and if a model is inaccurate on an output, it should be uncertain about the same output.

\vspace{-1mm}
Extending on these ideas, we leverage the relationship between accuracy and uncertainty as an anchor for uncertainty calibration, since there is no ground-truth for uncertainty estimates. PAvPU metric is not differentiable to be used as a cost function while training the model. We propose differential approximations to the \textit{accuracy versus uncertainty} (AvU) defined in Equation~\ref{eqn:avu} to be used as utility function, which can be computed for a  mini-batch of data samples while training the model. We rely on the theoretically sound loss-calibrated approximate inference framework~\cite{lacoste2011approximate,cobb2018loss} rooted in Bayesian decision theory~\cite{berger1985statistical} by introducing AvUC loss as an additional utility-dependent penalty term to accomplish the task of improving uncertainty calibration. 
A task-specific utility function is employed in Bayesian decision theory to accomplish optimal predictions. In this work, AvU utility function is optimized for the task of obtaining well-calibrated uncertainties (model to provide lower uncertainty for accurate predictions and higher uncertainty towards inaccurate predictions). To estimate the AvU metric during each training step, outputs within a mini-batch can be grouped into four different categories: [i] accurate and certain (AC) [ii] accurate and uncertain (AU) [iii] inaccurate and certain (IC) [iv] inaccurate an uncertain (IU). $n_{AC}$, $n_{AU}$, $n_{IC}$ and $n_{IU}$ represent the number of samples in the categories AC, AU, IC and IU respectively.
 

\begin{minipage}{0.5\textwidth}
	\centering
		\includegraphics[scale=0.35]{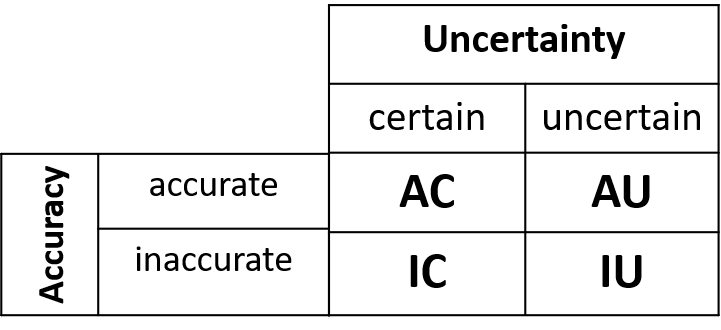}
\end{minipage}
\begin{minipage}{0.5\textwidth}	
	\begin{equation}
	\label{eqn:avu}
	\centering
	\mathrm{AvU}=\frac{\mathrm{n}_{\mathrm{AC}}+\mathrm{n}_{\mathrm{IU}}}{\mathrm{n}_{\mathrm{AC}}+\mathrm{n}_{\mathrm{AU}}+\mathrm{n}_{\mathrm{IC}}+\mathrm{n}_{\mathrm{IU}}}
	\end{equation}
\end{minipage}
\newline

A reliable and well-calibrated model will provide higher AvU measure ($\mathrm{AvU} \in[0,1]$). Ideally, we expect the model to be certain about its predictions when it is accurate and provide high uncertainty estimates when making inaccurate predictions. We propose differentiable approximations to the AvU utility and introduce a trainable uncertainty calibration loss ($\mathcal{L}_{\mathrm{AvUC}}$) in section~\ref{sec:avuloss}, which serves as the utility-dependent penalty term within the loss-calibrated approximate inference framework described in section~\ref{sec:avul}.

\subsection{Differentiable accuracy versus uncertainty calibration (AvUC) loss}
\label{sec:avuloss}
 \textbf{Notations} \quad  Consider a multi-class classification problem on a large labeled dataset with N examples and K classes denoted by $\mathcal{D}=\left\{\left(\mathrm{x_n}, \mathrm{y_n}\right)\right\}_{n=1}^{N}$. Dataset is partitioned into $\mathrm{M}$ mini-batches i.e.  $\mathcal{D}=\left\{\mathcal{D}_m\right\}_{m=1}^{M}$, each containing $\scriptstyle\mathrm{B}=\scriptstyle\mathrm{N/M}$ examples. During training, we process a group of randomly sampled examples (mini-batch) $\mathcal{D}_m=\left\{\left(\mathrm{x_i}, \mathrm{y_i}\right)\right\}_{\scriptstyle{i=1}}^{B}$ per iteration. For each example with input $\mathrm{x_i} \in \mathcal{X}$ and $\mathrm{y_{i}} \in \mathcal{Y}=\small{\{1,2, \cdots, k}\}$ representing the ground-truth class label, let $p_{i}\left(\mathrm{y}| \mathrm{x}_{i}, \mathrm{w}\right)$ be the output from the neural network $f_{\mathrm{\mathrm{w}}}\left(\mathrm{y} | \mathrm{x_i}\right)$. In case of probabilistic models, predictive distribution is obtained from $\mathrm{T}$ stochastic forward passes (Monte Carlo samples), $p_i\left(\mathrm{y}| \mathrm{x_i}, \mathrm{w}\right)=\frac{1}{T} \sum_{t=1}^T p_i^t\left(\mathrm{y} | \mathrm{x_i}, \mathrm{w}_{t}\right)$. Let us define $\widehat{\mathrm{y}_{i}}=\arg \max _{\mathrm{y} \in \mathcal{Y}} p_{i}\left(\mathrm{y} | \mathrm{x}_{i}, {\mathrm{w}}\right)$ as the predicted class label, $p_{i}=\max _{\mathrm{y} \in \mathcal{Y}} p_{i}\left(\mathrm{y} | \mathrm{x}_{i}, {\mathrm{w}}\right)$ and $u_{i}=-\sum_{\mathrm{y} \in \mathcal{Y}}p_i\left(\mathrm{y} | \mathrm{x_i}, \mathrm{w}\right) \log p_i\left(\mathrm{y} | \mathrm{x_i}, \mathrm{w}\right)$ as confidence (probability of predicted class) and predictive uncertainty estimate for the model prediction respectively. $u_{th}$ is the threshold above which prediction is considered to be uncertain, and $\mathbbm{1}$ is the indicator function. 
\begin{equation}
\label{eqn:indicator_fn}
\begin{split}
n_{A U}:=\sum_{i} \mathbbm{1}(\mathrm{\widehat{y_i}}=\mathrm{y_i} \; \small\textit { and } \; u_i > u_{\mathrm{th}}) \quad;\quad n_{I C}:=\sum_{i} \mathbbm{1}(\mathrm{\widehat{y_i}} \neq \mathrm{y_i} \; \small\textit { and } \; u_i \leq u_{\mathrm{th}}) \\
n_{A C}:=\sum_{i} \mathbbm{1}(\mathrm{\widehat{y_i}}=\mathrm{y_i} \; \small\textit { and } \; u_i \leq u_{\mathrm{th}}) \quad;\quad  n_{I U}:=\sum_{i} \mathbbm{1}(\mathrm{\widehat{y_i}} \neq \mathrm{y_i} \; \small\textit { and } \; u_i > u_{\mathrm{th}})
\end{split}
\end{equation}
We define the AvUC loss function representing negative log AvU in Equation ~\ref{eqn:avu_loss}. In order to make the loss function differentiable with respect to the neural network parameters, we define proxy functions to approximate $n_{AC}$, $n_{AU}$, $n_{IC}$ and $n_{IU}$ as given by Equations~\ref{eqn:avu_comp}. The hyperbolic tangent function is used to scale the uncertainty values between 0 and 1, $\tanh(u_i) \in [0,1]$. The intuition behind these approximations is that the probability of the predicted class $\{p_i \to 1\}$ when the predictions are accurate and $\{p_i \to 0\}$ when inaccurate. Also, the scaled uncertainty $\{\tanh(u_i) \to 0 \}$ when the predictions are certain and $\{\tanh(u_i) \to 1\}$  when uncertain. Under ideal conditions, these proxy functions in Equation~\ref{eqn:avu_comp} will be equivalent to indicator functions defined in Equations~\ref{eqn:indicator_fn}. This loss function can be used with standard gradient descent optimization and enables the model to learn to provide well-calibrated uncertainties, in addition to improved prediction accuracy. Minimizing the AvUC loss function is equivalent to maximizing AvU measure (Equation~\ref{eqn:avu}). 
The AvUC loss will be perfect 0 only when all the accurate predictions are certain and inaccurate predictions are uncertain. 

\begin{equation}
\label{eqn:avu_loss}
\begin{split}
\centering
\mathcal{L}_{\mathrm{AvUC}}:= -\log \left(\frac{\mathrm{n}_{\mathrm{AC}}+\mathrm{n}_{\mathrm{IU}}}{\mathrm{n}_{\mathrm{AC}}+\mathrm{n}_{\mathrm{IU}}+\mathrm{n}_{\mathrm{AU}}+\mathrm{n}_{\mathrm{IC}}}\right) =  \log \left(\mathrm{1}+\frac{\mathrm{n}_{\mathrm{AU}}+\mathrm{n}_{\mathrm{IC}}}{\mathrm{n}_{\mathrm{AC}}+\mathrm{n}_{\mathrm{IU}}}\right)
\end{split}
\end{equation}
where;
\begin{equation}
\begin{gathered}
\begin{aligned}
\label{eqn:avu_comp}
&\mathrm{n}_{A U} = \sum_{i \in \left\{\substack{\mathrm{\widehat{y_i}}=\mathrm{y_i}  \small\textit { and } \\[2pt]  u_{i} > u_{\mathrm{th}}}\right\}} p_{i} \odot \tanh \left(u_{i}\right) \quad\quad\quad;\quad
\mathrm{n}_{I C}=\sum_{i \in \left\{\substack{\mathrm{\widehat{y_i}} \neq \mathrm{y_i}    \small\textit {and}  \\[2pt] u_{i} \leq u_{\mathrm{th}}}\right\}}\left(1-p_{i}\right)\odot\left(1-\tanh \left(u_{i}\right)\right) \\
&\mathrm{n}_{AC}=\sum_{i \in \left\{\substack{\mathrm{\widehat{y_{i}}}=\mathrm{y_{i}} \small\textit { and } \\[2pt] u_{i} \leq u_{\mathrm{th}}}\right\}} p_{i}\odot\left(1-\tanh \left(u_{i}\right)\right) \;\; ;\quad
\mathrm{n}_{I U}=\sum_{i \in \left\{\substack{\mathrm{\widehat{y_i}} \neq \mathrm{y_i}  \small\textit { and } \\[2pt]  u_{i} > u_{\mathrm{th}}}\right\}}\left(1-p_{i}\right)\odot \tanh \left(u_{i}\right)
\end{aligned}
\end{gathered}
\end{equation}

AvUC loss is devised to improve uncertainty calibration that can be used as an additional penalty term and combined with existing losses without modifying the underlying principles (e.g. ELBO for Bayesian DNN, cross-entropy for non-Bayesian DNN classifier). AvUC enables uncertainty calibration by overcoming the challenge of unavailability of ground truth uncertainty estimates while accounting for the quality of principled aleatoric and epistemic uncertainties, which are important for many applications.


\subsection{Loss-calibrated approximate inference with AvUC loss}
\label{sec:avul}
 The loss-calibrated evidence lower bound (ELBO) is defined in Equation~\ref{eqn:total_loss} that incorporates AvUC loss as an additional utility-dependent penalty term and $\beta$ is the hyperparameter for relative weighting of AvUC loss with respect to ELBO. We illustrate our method with mean-field stochastic variational inference (SVI)~\cite{graves2011practical,blundell2015weight}. Our implementation is shown in Algorithm~\ref{alg:svi_avul} and we refer to this method as \textbf{SVI-AvUC}. The operations~\ref{op:unc}-\ref{op:loss} in Algorithm~\ref{alg:svi_avul} are the additional steps with respect to standard SVI.
\begin{equation}
\label{eqn:total_loss}
\mathcal{L}:=\underbrace{\underbrace{-\mathbb{E}_{q_{\theta}(\mathrm{w})}[\log p(\mathrm{y} | \mathrm{x,w})]}_\textrm{expected negative log likelihood} \; + \; \underbrace{\mathrm{KL}[q_{\theta}(\mathrm{w}) || p(\mathrm{w})]}_\textrm{Kullback-Leibler divergence}}_{\mathcal{L}_{\textrm{ELBO}} (\textrm{negative ELBO})} \;\; + \quad \mathrm{\beta} \; \underbrace{\; \log \left(\mathrm{1}+\frac{\mathrm{n}_{A U}+\mathrm{n}_{I C}}{\mathrm{n}_{A C}+\mathrm{n}_{I U}}\right)}_{\mathcal{L}_\textrm{AvUC}\textrm{(AvUC loss)}}
\end{equation}

\vspace{-3mm}
\begin{algorithm}
	\caption{\;SVI-AvUC optimization}
	\label{alg:svi_avul}
	\begin{algorithmic}[1]
		\State Given dataset $D=\{X,Y\}$
		\State let variational parameters $\theta=(\mu,\rho)$  \Comment{ approx variational posterior $q_{\theta}(\mathrm{w})=\mathcal{N}(\mu, \log (1+e^{\rho}))$ }
		\State set the weight priors, $p(\mathrm{w}):=\mathcal{N}(0,I)$ 
		\State initialize $\mu$ and $\rho$
		\State define learning rate schedule $\alpha$
		\Repeat
		\State Sample $B$ index set of training samples; $\mathcal{D}_m=\left\{\left(\mathrm{x_i}, \mathrm{y_i}\right)\right\}_{\scriptstyle{i=1}}^{B}$ \Comment{batch-size}
		\For{ $i \in B$ }
		\For{$t \gets 1$ to $T$}  \Comment{T Monte Carlo samples}
		\State Sample $\epsilon \sim \mathcal{N}(0, I)$
		\State $\mathrm{w_t}=\mu+\log (1+\exp (\rho)) \small\odot \epsilon$ \Comment{$ \small\odot$  represents pointwise multiplication}
		\State ${p_i^t}\left(\mathrm{y}| \mathrm{x_i}, \mathrm{w_t}\right)=f_{\mathrm{w} \sim \mathrm{q_\theta(w)}}\left(\mathrm{y} | \mathrm{x_i}\right)$ \Comment{ perform a stochastic forward pass with sampled weight}
		\EndFor  
		\State \begin{varwidth}[t]{\linewidth}
			Obtain predictive distribution from $T$ stochastic forward passes\par
			
			\hskip\algorithmicindent  \normalsize $p_i\left(\mathrm{y}| \mathrm{x_i}, \mathrm{w}\right)=\frac{1}{T} \sum_{t=1}^T p_i^t\left(\mathrm{y} | \mathrm{x_i}, \mathrm{w}_{t}\right)$ \par
		\end{varwidth}
	    \State \begin{varwidth}[t]{\linewidth}
	    	Obtain predicted label and probability of predicted class   \par
	    	\hskip\algorithmicindent $\widehat{\mathrm{y}_{i}}\leftarrow\underset{\mathrm{y} \in \mathrm{Y}}{\operatorname{argmax}} \; p_{i}\left(\mathrm{y} | \mathrm{x}_{i}, {\mathrm{w}}\right) \quad ;\quad p_{i}\leftarrow\underset{\mathrm{y} \in \mathrm{Y}}{\operatorname{max}} \; p_{i}\left(\mathrm{y} | \mathrm{x}_{i}, {\mathrm{w}}\right)$
	    \end{varwidth}
		\State \label{op:unc} \begin{varwidth}[t]{\linewidth}
			Calculate predictive uncertainty    \Comment{ predictive entropy}\par
			\hskip\algorithmicindent${\bold u_i}=-\sum_{k}\left(\frac{1}{T} \sum_{t} p_i^t\left(\mathrm{y}=k | \mathrm{x_i}, \mathrm{w}_{t}\right)\right) \log \left(\frac{1}{T} \sum_{t} p_i^t\left(\mathrm{y}=k | \mathrm{x_i}, \mathrm{w}_{t}\right)\right)$ \quad \Comment{where; $\mathrm{w}_{t}\sim q_{\theta}(\mathrm{w})$}\par
		\end{varwidth}
	    \EndFor
		\State \label{op:calc_nac} Compute $\mathbf{n_{A C}}, \mathbf{n_{A U}}, \mathbf{n_{I C}}, \mathbf{n_{I U}}$ \Comment{ Equations ~\ref{eqn:avu_comp}}
		\State \label{op:loss} Compute loss-calibrated ELBO (total loss),  $\mathcal{L} = \mathcal{L}_{\textrm{ELBO}} + \mathcal{L}_{\textbf{AvUC}}$  \Comment{ Equation ~\ref{eqn:total_loss}}
		\State Compute the gradients of loss function w.r.t $\mu$ and $\rho$, $\Delta\mathcal{L}_{\mu}$ and  $\Delta\mathcal{L}_{\rho}$ respectively
		\State \begin{varwidth}[t]{\linewidth}
			Update the variational parameters $\mu$ and $\rho$\par
			\hskip\algorithmicindent $\mu \leftarrow \mu-\alpha \Delta\mathcal{L}_{\mu}$ \par
			\hskip\algorithmicindent $\rho \leftarrow \rho-\alpha \Delta\mathcal{L}_{\rho}$
		\end{varwidth}		
		\Until{$\mu$ and $\rho$ has converged, or when stopped}		
	\end{algorithmic}
\end{algorithm}
 AvU is the utility function which guides optimal predictions in accomplishing the task of getting well-calibrated uncertainties and proposed AvUC loss serves as an utility-dependent penalty term within the loss-calibrated inference framework. For the initial few epochs, we train the model only with ELBO loss as this allows to learn the uncertainty threshold required for AvUC loss\footnote[2]{\scriptsize \label{note1}We also optimized area under the curve of AvU across various uncertainty thresholds towards a threshold free mechanism as presented in Appendix~\ref{appdx:auavuc}, but the results are similar except being more compute intensive during training.}.The threshold is obtained from the average of predictive uncertainty mean for accurate and inaccurate predictions on the training data from initial epochs.

Theoretically AvUC loss will be equal to 0 only when the model's uncertainty is perfectly calibrated (utility function is maximized, AvU=1). As noted in Equations~\ref{eqn:avu_loss} and~\ref{eqn:avu_comp}, AvUC loss attempts to maximize the utility function AvU, which will indirectly push the values of uncertainties up or down based on the accuracy of predictions. When uncertainty estimates are not accurate, $\mathrm{AvU}\to0$ and $\mathcal{L}_{\mathrm{AvUC}}\to\infty$ guiding the gradient computation exert AvUC loss towards 0, which will happen when AvU score is pushed higher ($\mathrm{AvU}\to 1$), enabling the model to maximize the utility to provide well-calibrated uncertainties. In Appendix~\ref{appdx:monitorloss}, we show how AvUC loss and ELBO loss vary during training and the impact of AvUC regularization term on loss-calibrated ELBO (total loss) and actual AvU score.

\vspace{-2mm}
\subsection{Post-hoc model calibration with AvU temperature scaling (AvUTS)}
\label{sec:avuts}
We propose post-hoc uncertainty calibration for pretrained models by extending the temperature scaling~\cite{guo2017calibration} methodology to optimize the AvUC loss instead of NLL. The optimal temperature $\mathrm{T} > 0$, a scalar value to rescale the logits of final layer is identified by minimizing the AvUC loss as defined in Equation~\ref{eqn:avu_loss} on held-out validation set. The uncertainty threshold \textsuperscript{\ref{note1}} required for calculating $n_{AC}$, $n_{AU}$, $n_{IC}$ and $n_{IU}$ is obtained by finding the average predictive uncertainty for accurate and inaccurate predictions from the uncalibrated model using the same held-out validation data $\mathcal{D}_\mathrm{v}=\left\{\left(\mathrm{x_v}, \mathrm{y_v}\right)\right\}_{{\mathrm{v}=1}}^{V}$,  $\mathrm{u}_{th}=\mathrm{\left(\frac{\overline{u}_{ (\widehat{y}_{v}={y_{v}})} + \overline{u}_{ (\widehat{y}_{v}\neq{y_{v}})}}{2} \right)}$. We refer this method applied to pretrained SVI model as \textbf{SVI-AvUTS}.

\vspace{-1.5mm}
\section{Experiments and Results}
\label{sec:results}
We perform a thorough empirical evaluation of our proposed methods SVI-AvUC and SVI-AvUTS on large-scale image classification task under distributional shift. We evaluate the model calibration; model performance with respect to confidence and uncertainty estimates; and the distributional shift detection performance. We use ResNet-50 and ResNet-20~\cite{He_2016} DNN architectures on ImageNet~\cite{deng2009imagenet} and CIFAR10~\cite{krizhevsky2009learning} datasets respectively. We compare the proposed methods with various high performing non-Bayesian and Bayesian methods including  vanilla DNN (Vanilla), Temperature scaling (Temp scaling)~\cite{guo2017calibration}, Deep-ensembles (Ensemble)~\cite{lakshminarayanan2017simple}, Monte Carlo dropout (Dropout)~\cite{gal2016dropout}, Mean-field stochastic variational inference (SVI)~\cite{blundell2015weight,graves2011practical}, Temperature scaling on SVI (SVI-TS) and Radial Bayesian neural network (Radial BNN)~\cite{farquhar_radial_2020}. In Appendix~\ref{appdx:results}, we compare with additional methods, Dropout and SVI on the last layer of neural network (LL-Dropout and LL-SVI)~\cite{riquelme2018deep,Subedar_2019_ICCV}. The work from~\citet{snoek2019can} suggests SVI is very promising on small-scale problems, but is difficult to scale to larger datasets. We choose SVI as a baseline to illustrate our methods AvUC and AvUTS. We were able to scale SVI to the large-scale ImageNet dataset with ResNet-50 by specifying the weight priors and initializing the variational parameters using Empirical Bayes method following~\cite{krishnanspecifying}. The results for the methods: Vanilla, Temp scaling, Ensemble, Dropout, LL Dropout and LL SVI are obtained from the model predictions provided in UQ benchmark~\cite{snoek2019can} and we follow the same methodology for model evaluation under distributional shift by utilizing 16 different types of image corruptions at 5 different levels of intensities for each datashift type proposed in~\cite{hendrycks2019robustness}, resulting in 80 variations of test data for datashift evaluation. We refer to Appendix~\ref{appdx:datasetshift} for details on datashift types used in experiments, along with visual examples. All the methods are compared both under in-distribution and distributional shift conditions with same evaluation criteria for fair comparison. For SVI-AvUC implementation, we use the same hyperparameters as SVI baseline. We provide details of our model implementations and hyperparameters for SVI, SVI-TS, SVI-AvUC, SVI-AvUTS and Radial BNN in Appendix~\ref{appdx:experiments}.

\begin{figure*}[t!]
	\centering
	\vspace*{-2mm}
	\begin{subfigure}[b]{\textwidth}
		\centering
		\includegraphics[scale=0.335]{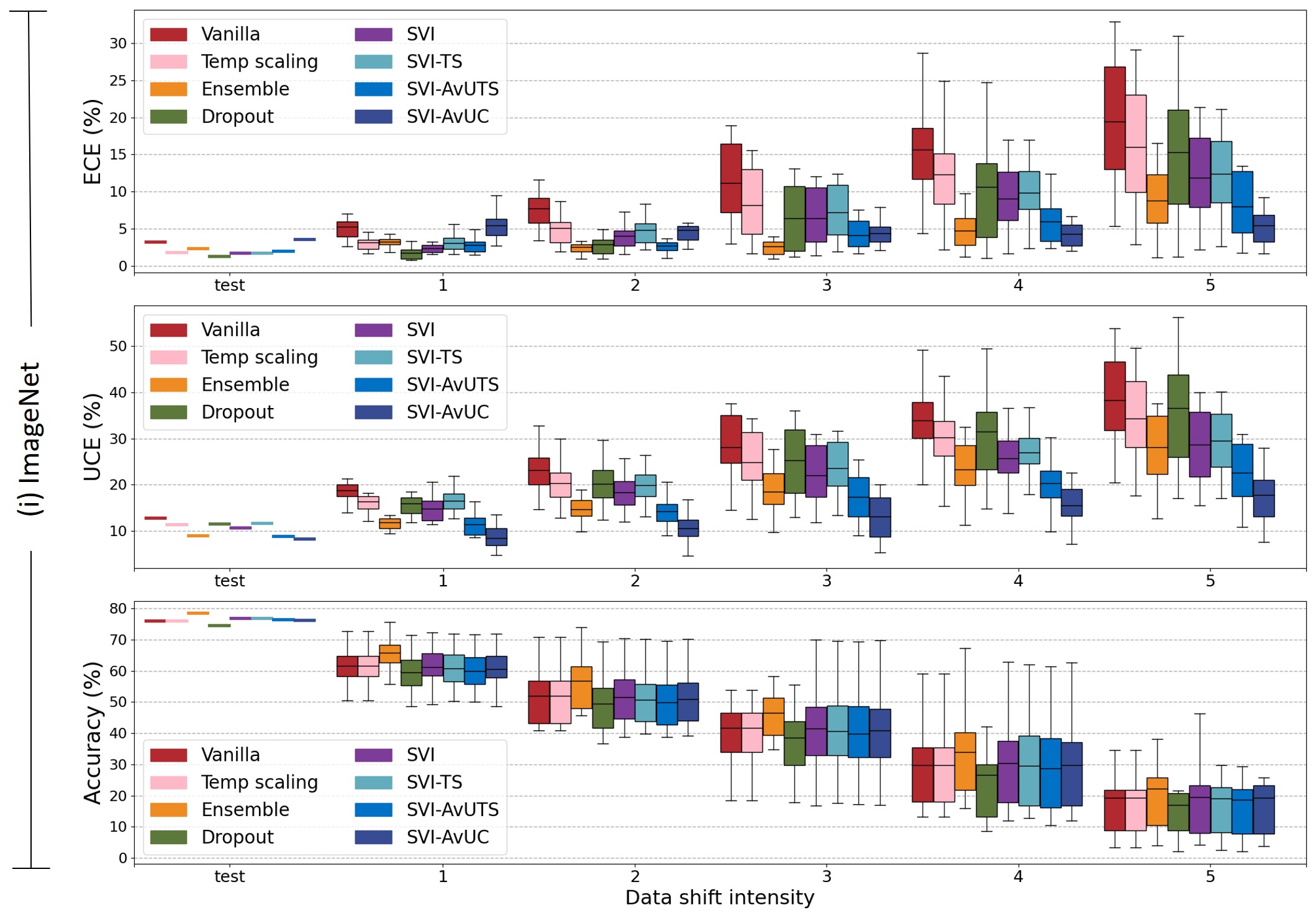}
	\end{subfigure}
	\begin{subfigure}[b]{\textwidth}
		\centering
		\includegraphics[scale=0.335]{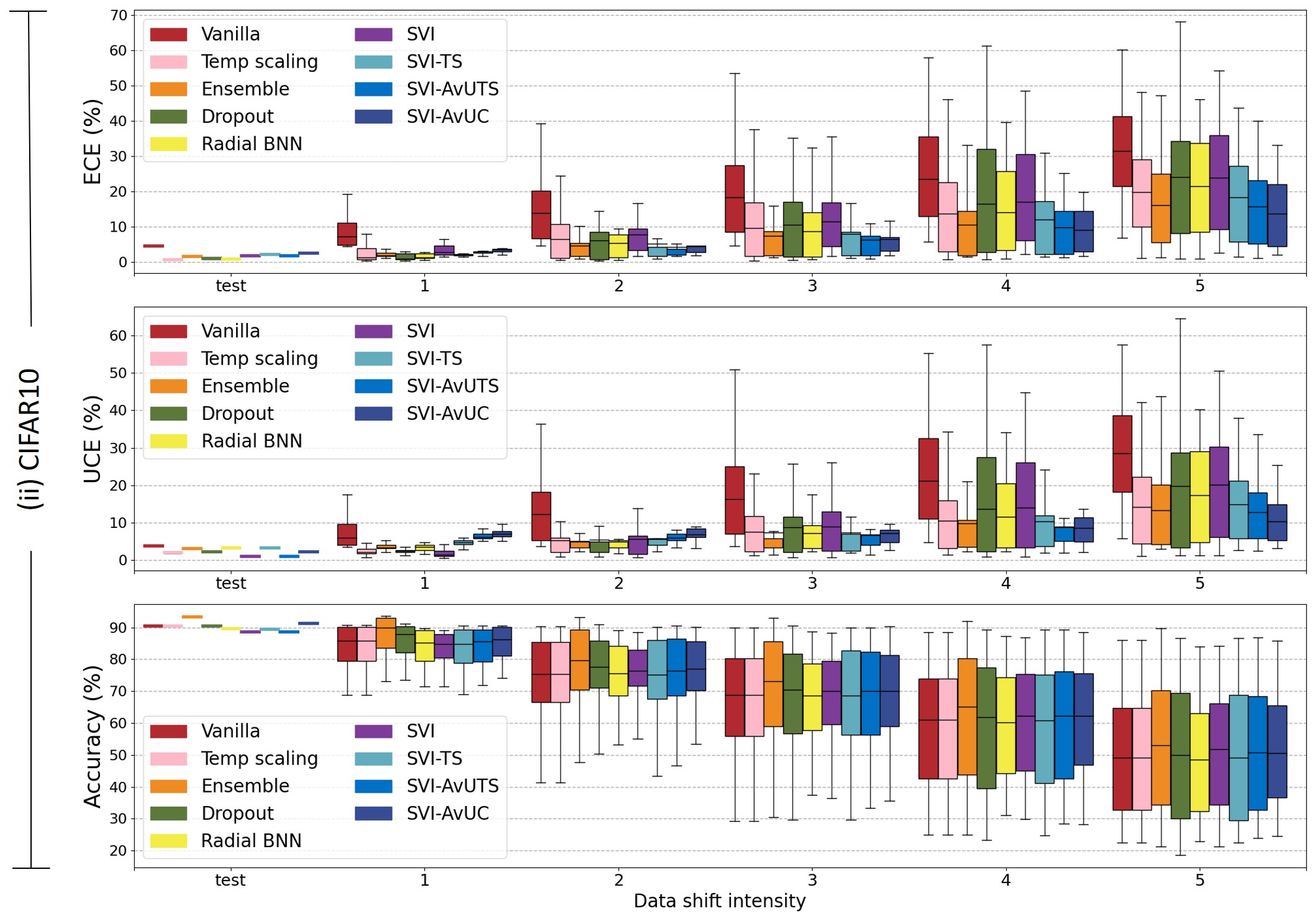}
	\end{subfigure}
	\caption{\small Model calibration comparison using ECE$\downarrow$ and UCE$\downarrow$ on (i) ImageNet and (ii) CIFAR10 under in-distribution (test) and dataset shift at different levels of shift intensities (1-5). A well-calibrated model should provide lower calibration errors even at increased datashift, though accuracy may degrade. At each shift intensity level, the boxplot summarizes the results across 16 different datashift types showing the min, max, mean and quartiles. \textbf{SVI-AvUC} provides lower ECE and UCE at increased dataset shift demonstrating it yields better model calibration compared to other methods. Spearman rank-order correlation coefficients~\cite{kokoska2000crc} assessing the relationship between calibration errors and dataset shift is provided in Appendix~\ref{appdx:addlmodelcalib}.  
	}
	\label{fig:boxplot}
	\vspace*{-3mm}
\end{figure*}

\paragraph{Model calibration evaluation}
\label{sec:cifar10}
We evaluate model calibration under in-distribution and dataset shift conditions following methodology in~\cite{snoek2019can}. 
Figure~\ref{fig:boxplot} shows the comparison of ECE, UCE and accuracy from different methods for test data (in-distribution) and dataset shift summarizing across 80 variations of shifted data on both ImageNet and CIFAR10. ECE represents the model calibration error with respect to confidence (probability of predicted class) and UCE represents the model calibration error with respect to predictive uncertainty representing entire predictive distribution of probabilities across the classes. A reliable and well-calibrated model should provide low calibration errors even with increased intensity of data shift, though accuracy may degrade with data shift. From Figure~\ref{fig:boxplot}, we can see the model accuracy reduces with increased data shift intensity and Ensembles method provides highest accuracy among existing methods. With model calibration, post-hoc calibration method SVI-AvUTS improves results over SVI baseline and SVI-AvUC outperforms all the methods by providing lower calibration errors (both ECE and UCE) at increased data shift levels while providing comparable model accuracy to Ensembles. We provide additional results (NLL, Brier score) and tables with numerical data comparison in Appendix~\ref{appdx:addlmodelcalib}.

\begin{table}[h!]
	\centering
	\small
	\vspace{-2mm}
	\caption{\small Additional results evaluating AvUC and AvUTS methods applied to Vanilla baseline on CIFAR10. Vanilla-AvUTS and Vanilla-AvUC provides lower ECE and UCE (mean across 16 different data shift types) compared to the baseline.}
	\begin{adjustbox}{width=1.0\textwidth}
			\begin{tabular}{@{}llccccccccccccc@{}}
				\toprule
				\multirow{2}{*}{Method} & \multirow{2}{*}{} & \multicolumn{6}{c}{ECE ($\%$)$\downarrow$ at various datashift intensities}                                                                     & \multicolumn{1}{l}{\multirow{2}{*}{}} & \multicolumn{6}{c}{UCE ($\%$)$\downarrow$ at various datashift intensities}                                                                   \\ \cmidrule(lr){3-8} \cmidrule(l){10-15} 
					&                   & \begin{tabular}[c]{@{}c@{}}0\end{tabular} & 1            & 2            & 3             & 4             & 5             & \multicolumn{1}{l}{}                  & \begin{tabular}[c]{@{}c@{}}0\end{tabular} & 1            & 2            & 3            & 4            & 5             \\ \midrule
					Vanilla                 &                   & 4.6                                                & 9.8          & 13.9         & 18.3          & 23.6          & 31.5          &                                       & 3.8                                                & 8.5          & 12.2         & 16.2         & 21.2         & 28.5          \\
					Vanilla-AvuTS           &                   & \textbf{2.1}                                       & \textbf{4.3} & \textbf{7.3} & 11.8          & 15.0          & 27.7          &                                       & \textbf{1.1}                                       & \textbf{2.6} & 4.7          & 8.6          & 11.7         & 23.1          \\
					Vanilla-AvUC            &                   & 3.4                                                & 5.7          & 8.4          & \textbf{11.4} & \textbf{14.3} & \textbf{23.2} & \textbf{}                             & 1.7                                                & 2.8          & \textbf{4.6} & \textbf{6.9} & \textbf{9.3} & \textbf{16.8} \\ \bottomrule
				\end{tabular}
	\end{adjustbox}
	\vspace{-10mm}
	\label{tab:deterministicavuc}
\end{table}

In addition to SVI-AvUC and SVI-AvUTS, we evaluate AvUC and AvUTS methods applied to vanilla baseline with entropy of softmax used as the predictive uncertainty in computing AvUC loss, which is combined with the cross-entropy loss. Table~\ref{tab:deterministicavuc} shows AvUTS and AvUC improves the model calibration errors (ECE and UCE) on the vanilla baseline as well.

\begin{figure}[hb!]
	\small
	\begin{multicols}{3}
	\begin{subfigure}{0.32\textwidth}
		\centering
		\captionsetup{
			justification=centering}
		\includegraphics[scale=0.137]{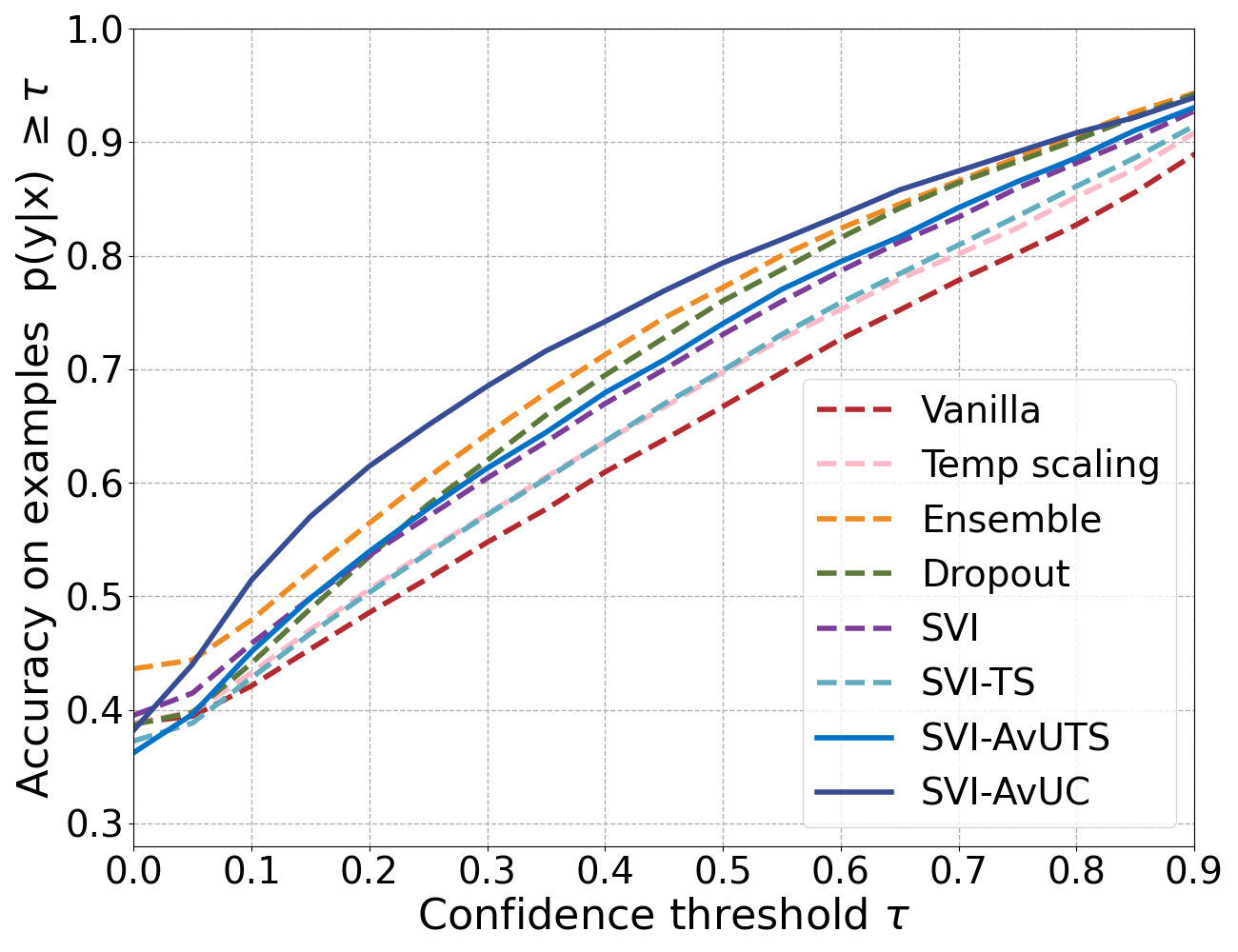}
		\caption{\scriptsize ImageNet: Confidence vs Accuracy ($\uparrow$)}
		\label{fig:imagenetaccvsconf}
	\end{subfigure}
    \begin{subfigure}{0.32\textwidth}
    	\centering
    	\captionsetup{
    		justification=centering}
    	\includegraphics[scale=0.137]{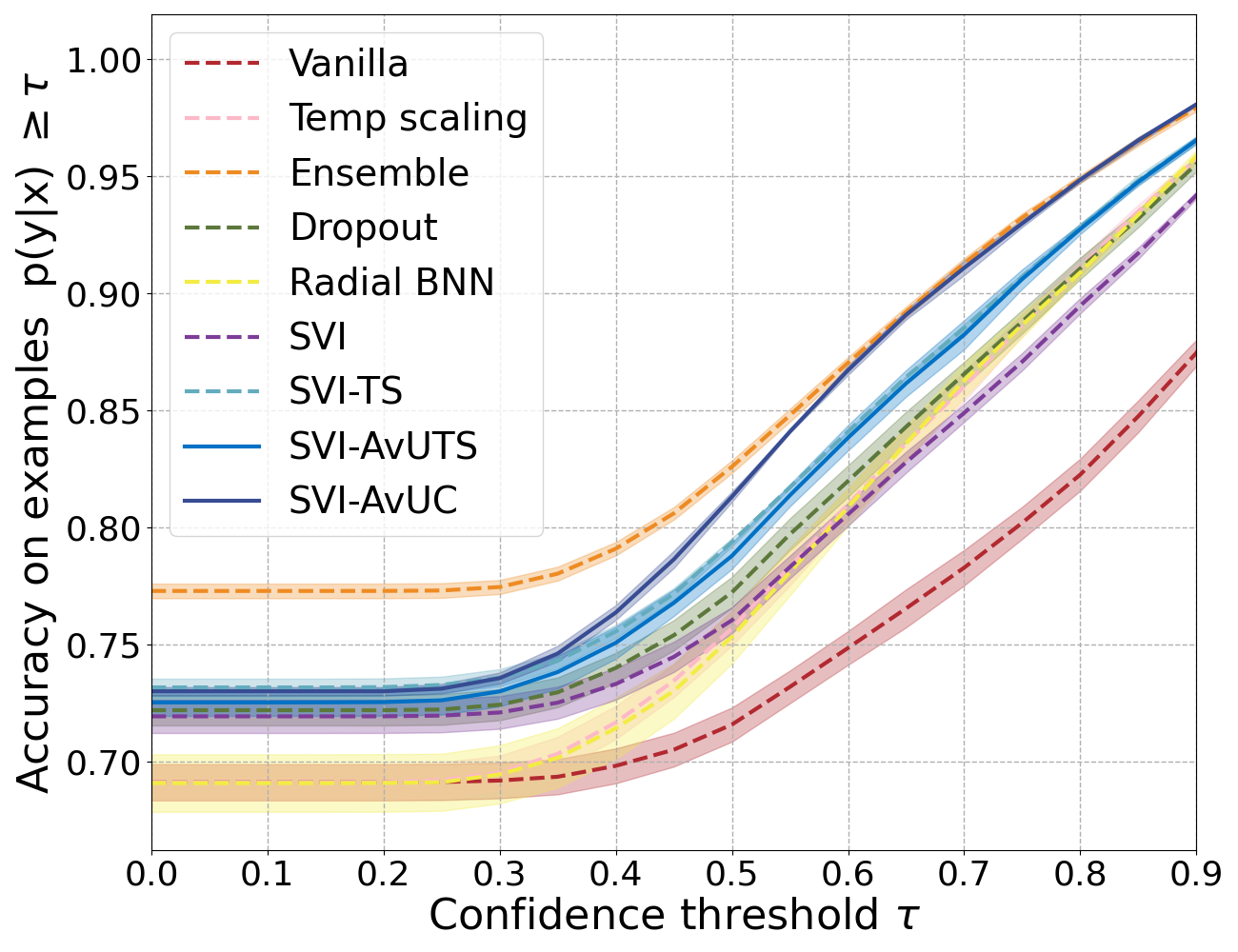}
    	\caption{\scriptsize CIFAR: Confidence vs Accuracy ($\uparrow$)}
    	\label{fig:cifaraccvsconf}
    \end{subfigure}
    \begin{subfigure}{0.32\textwidth}
    	\centering
    	\captionsetup{
    		justification=centering}
    	\includegraphics[scale=0.137]{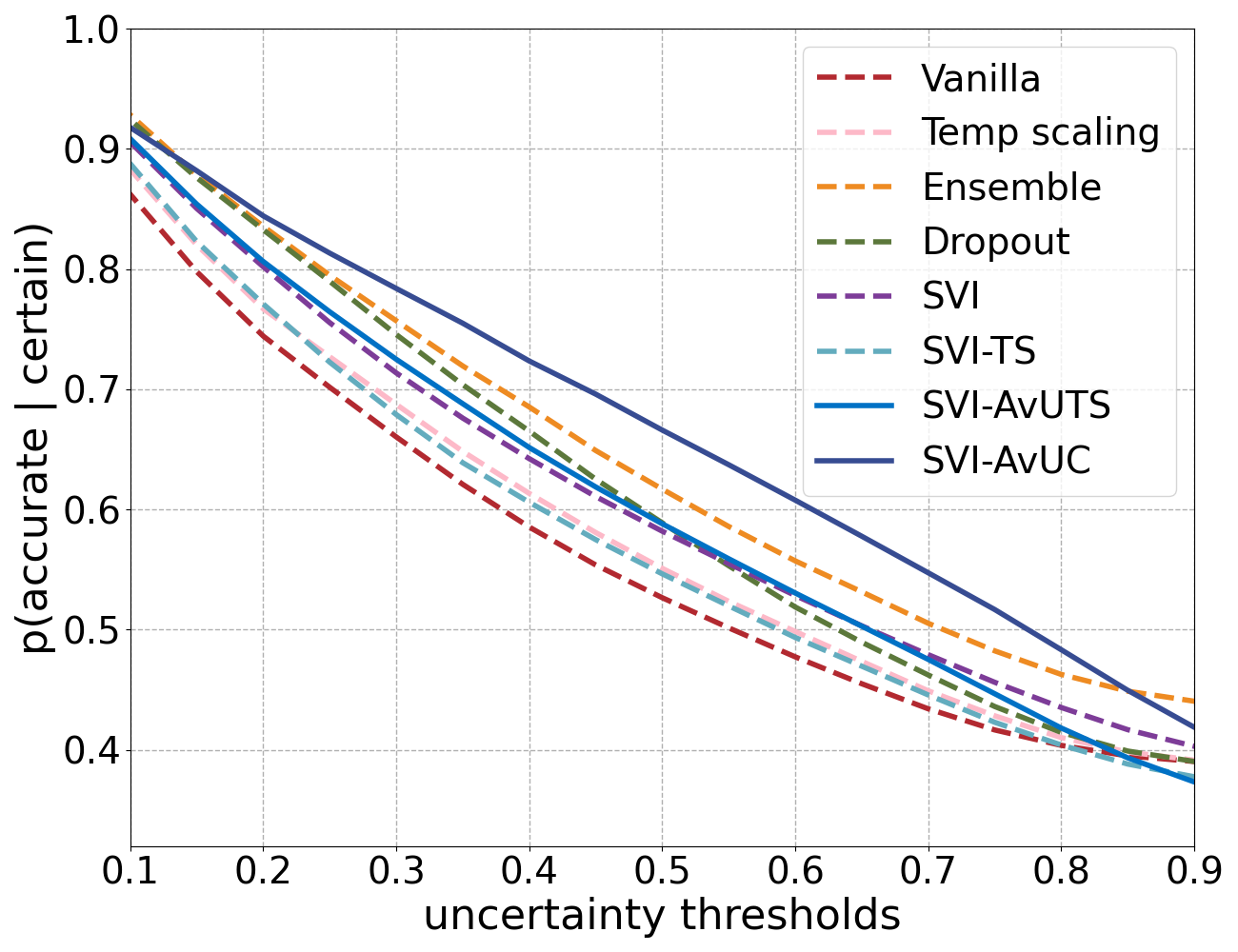}
    	\caption{\scriptsize ImageNet: Accurate when certain ($\uparrow$)  }
    	\label{fig:imagenetpac}
    \end{subfigure}
	\begin{subfigure}{0.32\textwidth}
		\centering
		\captionsetup{
			justification=centering}
		\includegraphics[scale=0.137]{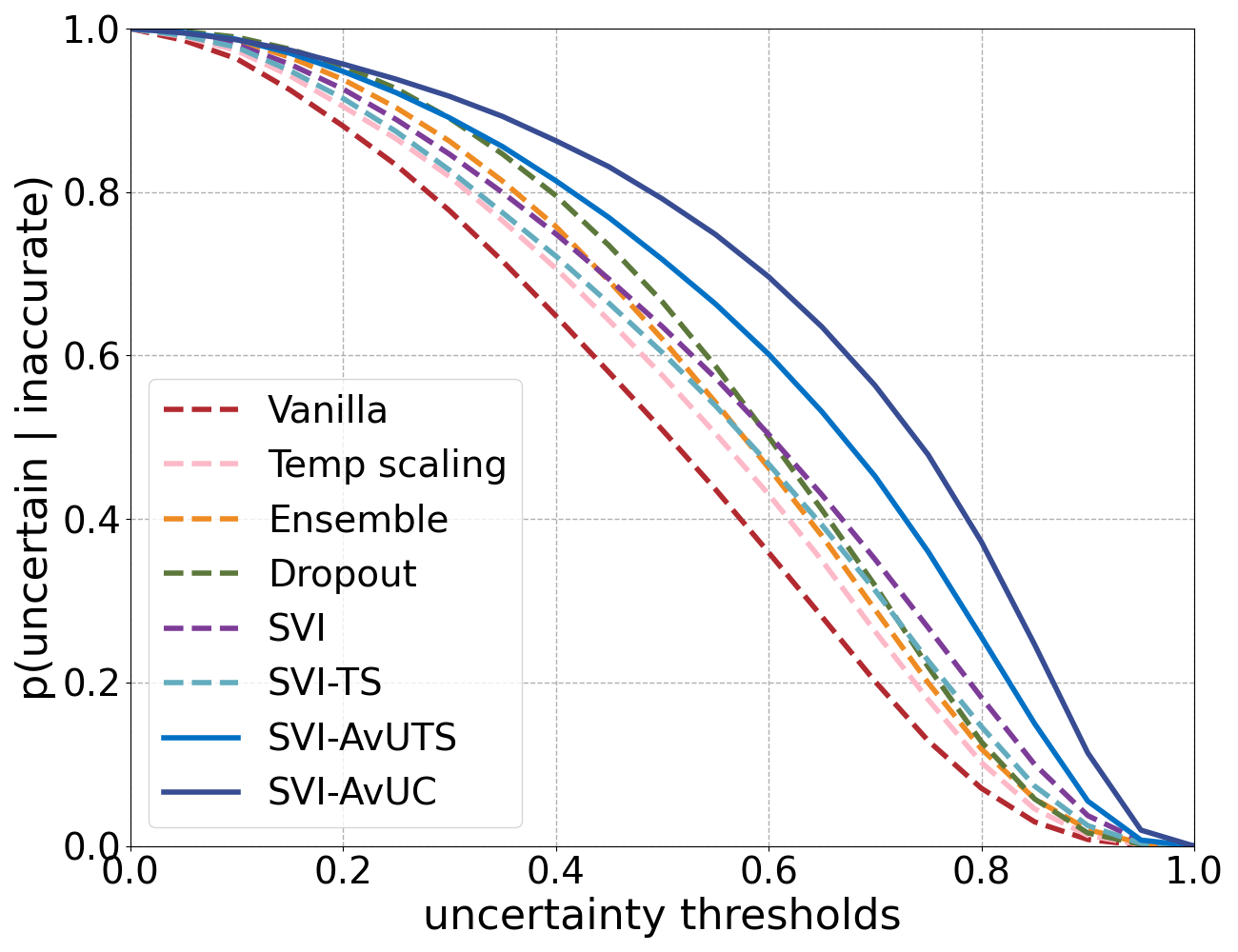}
		\caption{\scriptsize ImageNet: Uncertain when inaccurate($\uparrow$)}
		\label{fig:imagenetpui}
	\end{subfigure}	
    \begin{subfigure}{0.32\textwidth}
    	\centering
    	\captionsetup{
    		justification=centering}
    	\includegraphics[scale=0.137]{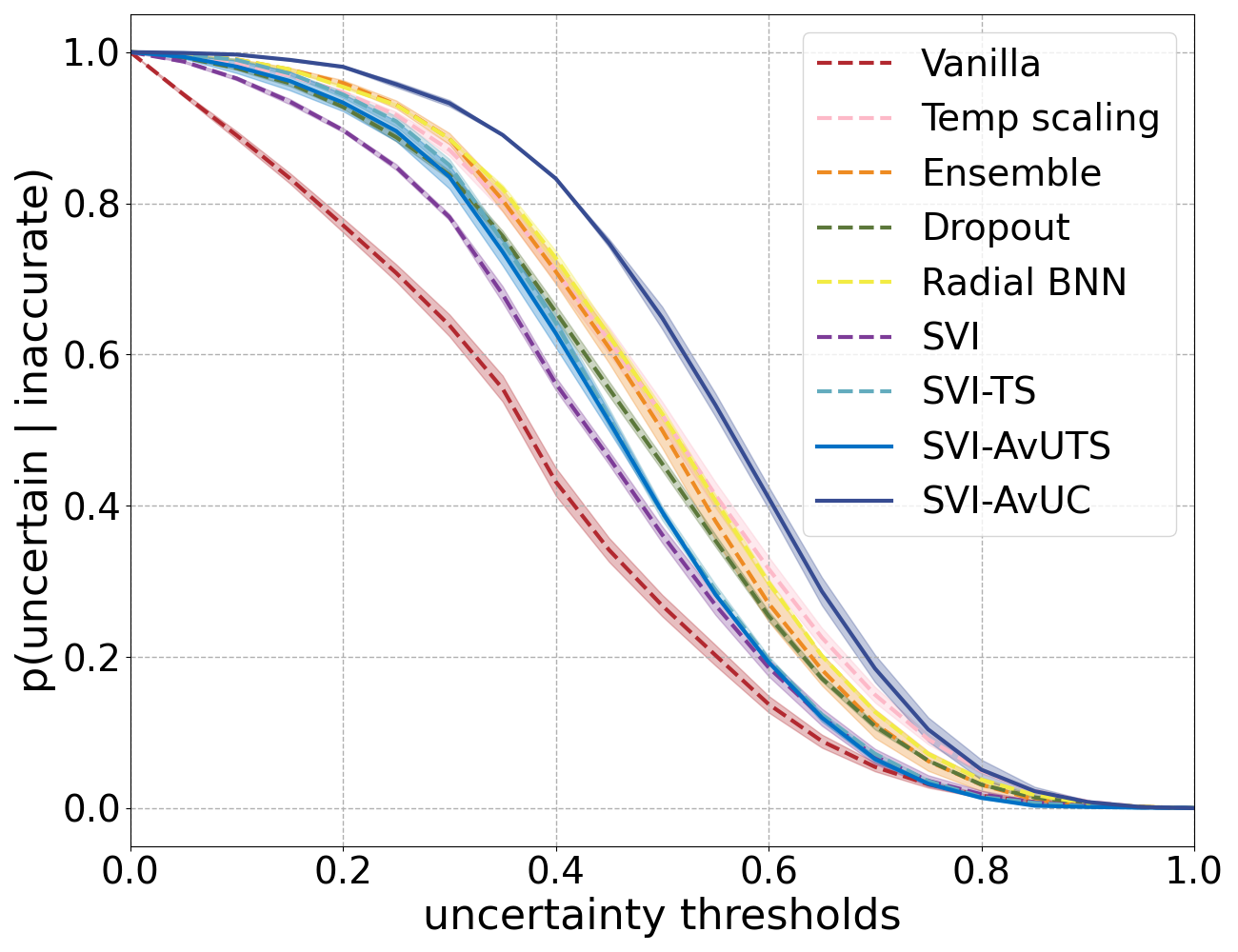}
    	\caption{\scriptsize CIFAR: Uncertain when inaccurate ($\uparrow$)}
    	\label{fig:cifarpui}
    \end{subfigure}	
    \begin{subfigure}{0.32\textwidth}
    	\centering
    	\captionsetup{
    		justification=centering}
    	\includegraphics[scale=0.137]{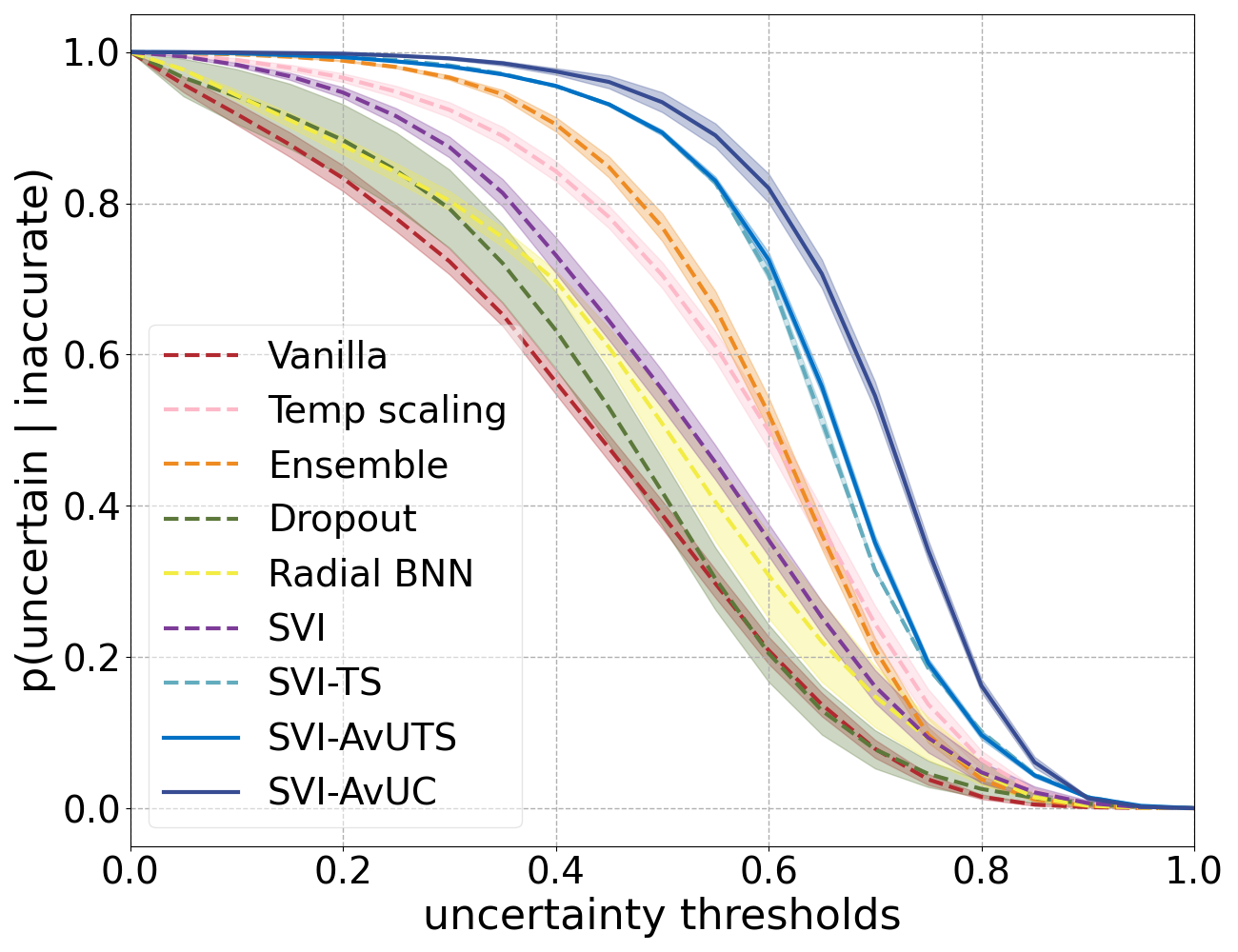}
    	\caption{\scriptsize OOD: Uncertain given inaccurate ($\uparrow$) }
    	\label{fig:oodpui}
    \end{subfigure}
    \begin{subfigure}{0.32\textwidth}
    	\centering
    	\captionsetup{
    		justification=centering}
    	\includegraphics[scale=0.137]{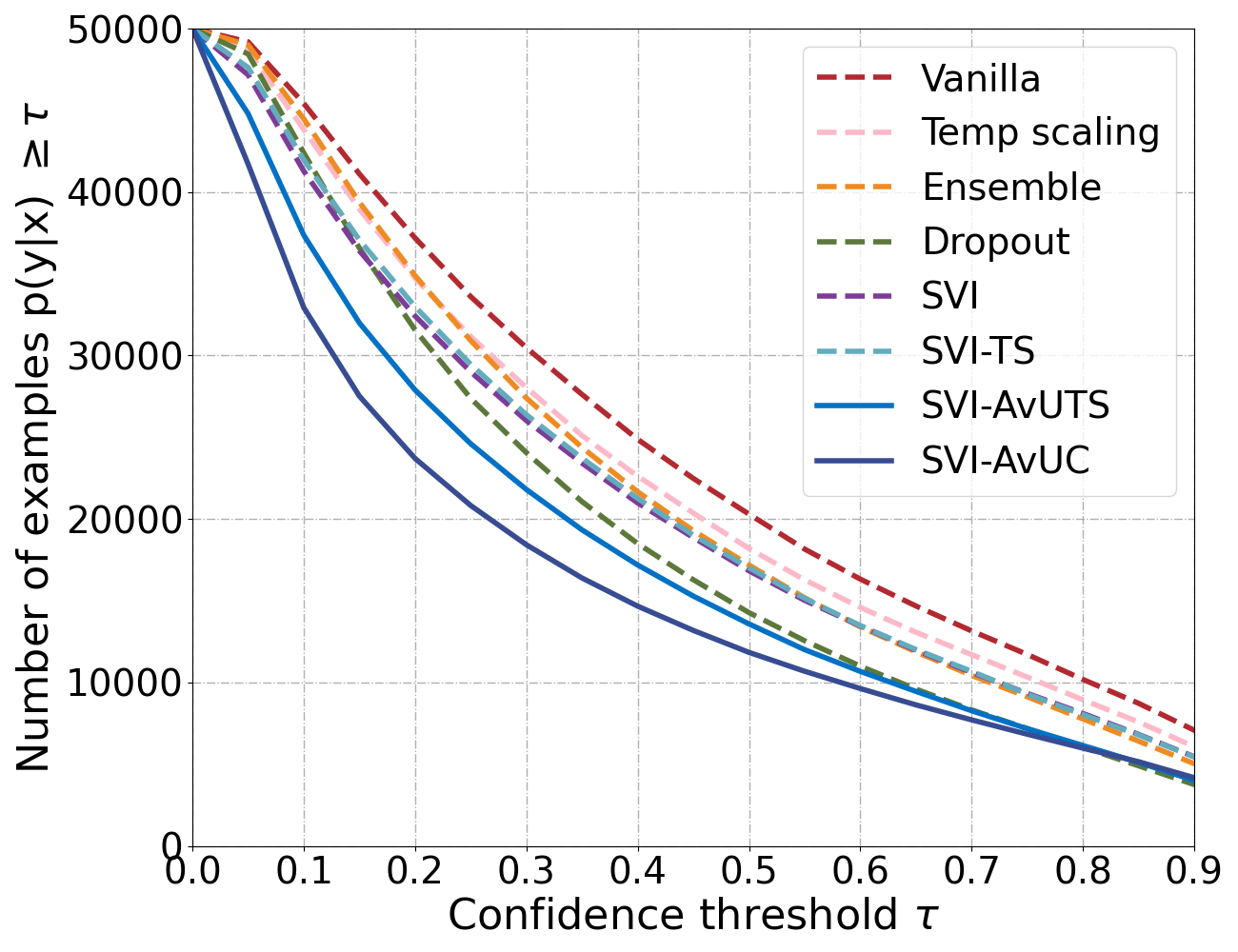}
    	\caption{\scriptsize ImageNet: Count vs Confidence ($\downarrow$)}
    	\label{fig:imagenetcount}
    \end{subfigure}	
	\begin{subfigure}{0.32\textwidth}
		\centering
		\captionsetup{
			justification=centering}
		\includegraphics[scale=0.137]{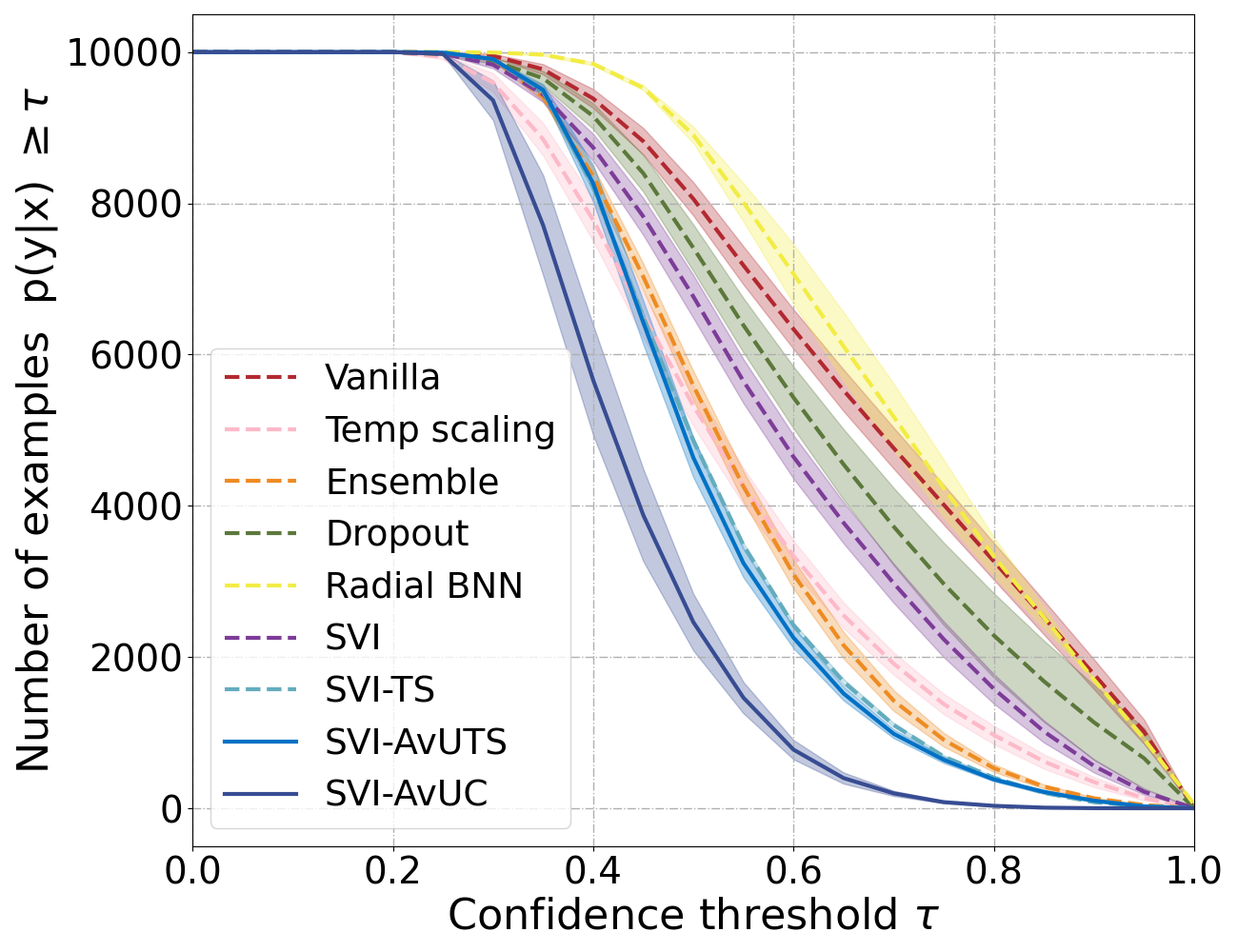}
		\caption{\scriptsize OOD: Count vs Confidence ($\downarrow$)}
		\label{fig:oodconf}
	\end{subfigure}
	\begin{subfigure}{0.32\textwidth}
		\centering
		\captionsetup{
			justification=centering}
		\includegraphics[scale=0.137]{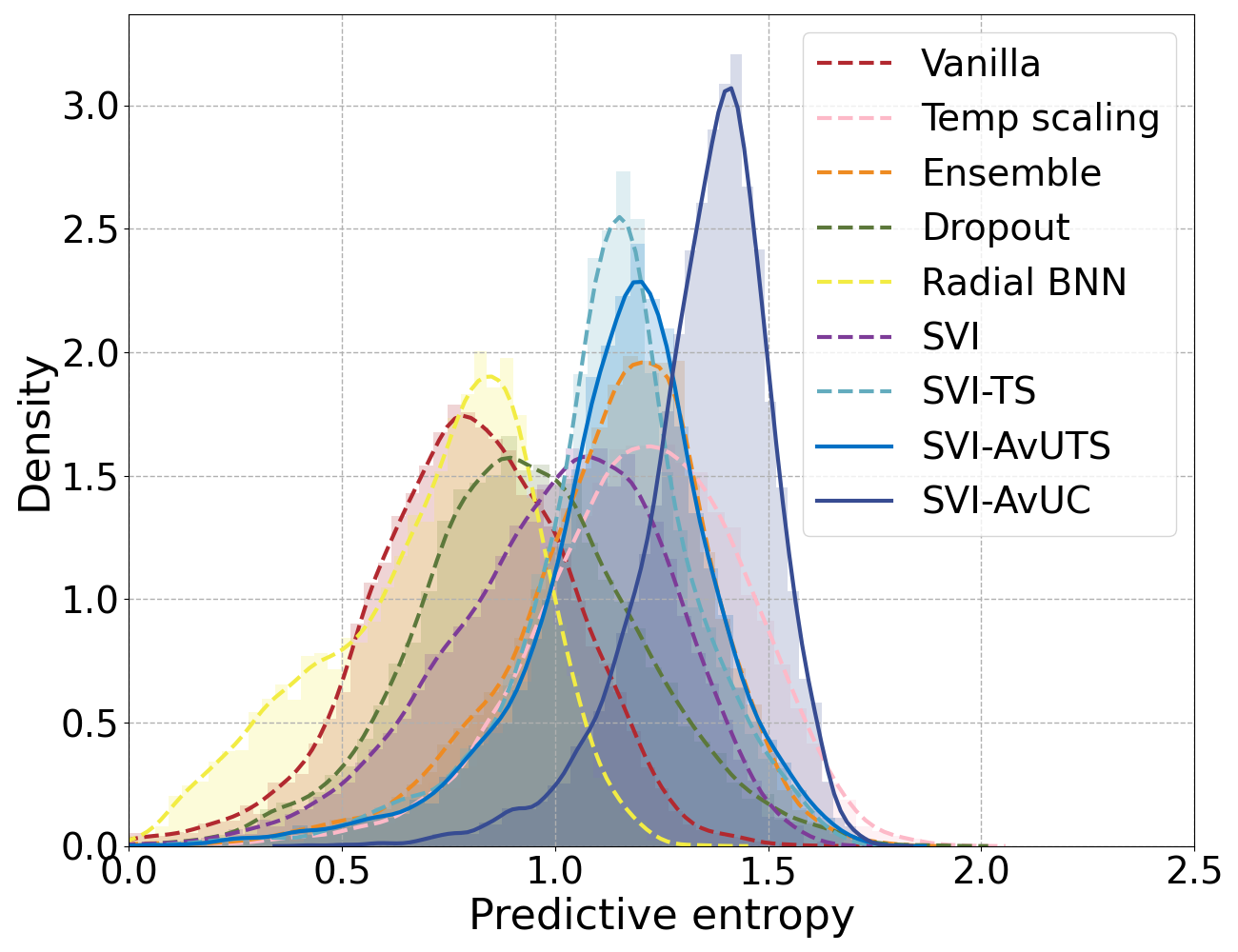}
		\caption{\scriptsize OOD: uncertainty density histogram ($\rightarrow$)}
		\label{fig:oodunc}
	\end{subfigure}
    \end{multicols}
	\caption{\small Model confidence and uncertainty evaluation under distributional shift (dataset shift on ImageNet and CIFAR10 with Gaussian blur of intensity 3, SVHN~\cite{netzer2011reading} is used as out-of-distribution (OOD) on model trained with CIFAR10). Column 1: ~\ref{fig:imagenetaccvsconf} and~\subref{fig:cifaraccvsconf} show accuracy as a function of confidence, ~\ref{fig:imagenetpac} show probability of model being accurate when certain about its predictions. Column 2: ~\ref{fig:imagenetpui},~\subref{fig:cifarpui} and ~\subref{fig:oodpui} show probability of model being uncertain when making inaccurate predictions. Normalized uncertainty thresholds $\mathrm{t\in[0,1]}$ are shown in plots as the uncertainty range varies for different methods. Column 3: ~\ref{fig:imagenetcount} and~\subref{fig:oodconf} show number of examples above given confidence value and ~\ref{fig:oodunc} shows density histogram of entropy on OOD data.}
	\label{fig:oodeval}
	\vspace*{-3mm}
\end{figure}

\vspace{-1mm} 
\paragraph{Model confidence and uncertainty evaluation}
\label{sec:confunceval}
We evaluate the quality of confidence measures using \textit{accuracy vs confidence} plots following the methodology from~\cite{lakshminarayanan2017simple,snoek2019can}. We evaluate the quality of predictive uncertainty estimates with \textit{p(uncertain | inaccurate)} and \textit{p(accurate | certain)} metrics across various uncertainty thresholds as proposed in~\cite{mukhoti2018evaluating}. A reliable model should be accurate when it is certain about its prediction and indicate high uncertainty when it is likely to be inaccurate. Figures~\ref{fig:imagenetaccvsconf} and~\subref{fig:cifaraccvsconf} show SVI-AvUC is more accurate at higher confidence, Figure~\ref{fig:imagenetpac} show SVI-AVUC is more accurate at lower uncertainty (being certain). Figures~\ref{fig:imagenetpui},~\subref{fig:cifarpui},~\subref{fig:oodpui} shows SVI-AvUC is more uncertain when making inaccurate predictions under distributional shift, compared to other methods. Figures~\ref{fig:imagenetcount} and~\subref{fig:oodconf} show SVI-AvUC has lesser number of examples with higher confidence when model accuracy is low under distributional shift. Figure~\ref{fig:oodunc} show SVI-AvUC provides higher predictive entropy on out-of-distribution data. We provide additional results in Appendix~\ref{appdx:addlconfunc}. In summary, SVI-AvUTS improves the quality of confidence and uncertainty measures over the SVI baseline, while preserving or improving accuracy. SVI-AvUC outperforms other methods in providing calibrated confidence and uncertainty measures under distributional shift.

\paragraph{Distributional shift detection}
\label{sec:shiftdetection}
We evaluate the performance of detecting distributional shift in neural networks using uncertainty estimates. This is a binary classification problem of identifying if an input sample is from in-distribution or shifted data. We evaluate using \textit{AUROC}, \textit{AUPR} and \textit{detection accuracy} metrics following the methodology in~\cite{lee2018simple}. We expect higher uncertainty under distributional shift as model tends to make inaccurate predictions and lower uncertainty for in-distribution data. In Figure~\ref{fig:shiftentropy}, we see better separation of predictive uncertainty densities for SVI-AvUC as compared to other methods, which is also quantified with Wasserstein distance~\cite{ramdas2017wasserstein}. In Table~\ref{tab:shiftdetection}, we present the dataset shift detection performance for ImageNet and CIFAR10 shifted with Gaussian blur at intensity 5. We also provide the out-of-distribution detection performance when the model trained with CIFAR10 is introduced with SVHN data during test time. Results in Table~\ref{tab:shiftdetection} show SVI-AvUC outperforms other methods in distributional shift detection.

\begin{table}[ht]
	\centering
	\small
	\begin{adjustbox}{width=1\textwidth}
		\begin{tabular}{llcccccccc}
			\hline \\[-1.5ex]
			Method                                                                              &  & Vanilla                    & Temp scaling               & Ensemble                   & Dropout                    & SVI                        & SVI-TS                     & SVI-AvUTS                  & SVI-AvUC                            \\ \hline \\[-1.5ex]
			\multicolumn{1}{c}{\begin{tabular}[c]{@{}c@{}}Wasserstein\\ distance\end{tabular}} &  & \multicolumn{1}{c}{2.7319} & \multicolumn{1}{c}{2.9098} & \multicolumn{1}{c}{3.0219} & \multicolumn{1}{c}{3.2837} & \multicolumn{1}{c}{3.7311} & \multicolumn{1}{c}{3.6961} & \multicolumn{1}{c}{3.9443} & \multicolumn{1}{c}{\textbf{4.2887}} \\ \hline
		\end{tabular}
	\end{adjustbox}
	\label{tab:wassimagenetgauss}
	\vspace{1mm}
	\includegraphics[width=1\linewidth]{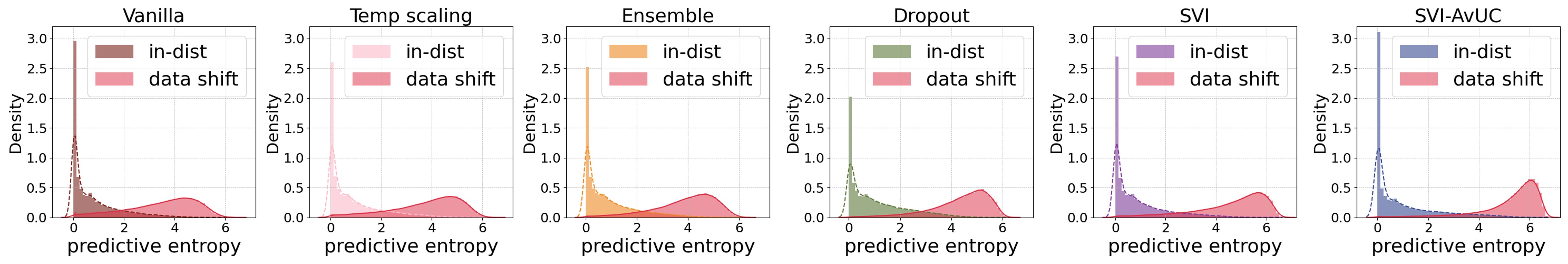}
	\vspace{0.2mm}
	\captionof{figure}{\small Density histograms of predictive uncertainty estimates on ImageNet in-distribution test set and data shifted with Gaussian blur of intensity 5. SVI-AvUC shows best separation of densities between in-distribution and data-shift as quantified by Wasserstein distance.} 
	\label{fig:shiftentropy}
\end{table}

\vspace{-6mm}
\begin{table}[h]
	\centering
	\caption{\small Distributional shift detection using predictive uncertainty. For dataset shift detection on ImageNet and CIFAR10, we use test data shifted with Gaussian blur of intensity 5. SVHN is used as out-of-distribution(OOD) data for OOD detection on model trained with CIFAR10.  All values are in percentages and best results are indicated in bold. SVI-AvUC outperforms across all the metrics.} 
	\begin{adjustbox}{width=1\textwidth}
				\centering
				\begin{tabular}{@{}lccccccccclcccc@{}}
					\toprule
					\multirow{2}{*}{ \quad Method} & \multicolumn{4}{c}{ImageNet (Dataset shift detection)}                                                                                                                                        &                      & \multicolumn{4}{c}{CIFAR10 (Dataset shift detection)}                                                                                                                                     &  & \multicolumn{4}{c}{CIFAR10 (OOD detection)}                                                                                                                                             \\ \cmidrule(lr){2-5} \cmidrule(lr){7-10} \cmidrule(l){12-15} 
					& \begin{tabular}[c]{@{}c@{}}AUROC\\ $\uparrow$\end{tabular}         & \begin{tabular}[c]{@{}c@{}}Detection\\ accuracy$\uparrow$\end{tabular} & \begin{tabular}[c]{@{}c@{}}AUPR\\   in$\uparrow$\end{tabular} & \begin{tabular}[c]{@{}c@{}}AUPR\\   out$\uparrow$\end{tabular} &                      & \begin{tabular}[c]{@{}c@{}}AUROC\\ $\uparrow$\end{tabular}          & \begin{tabular}[c]{@{}c@{}}Detection\\ accuracy$\uparrow$\end{tabular} & \begin{tabular}[c]{@{}c@{}}AUPR\\ in$\uparrow$\end{tabular} & \begin{tabular}[c]{@{}c@{}}AUPR\\ out$\uparrow$\end{tabular} &  & \begin{tabular}[c]{@{}c@{}}AUROC\\ $\uparrow$\end{tabular}          & \begin{tabular}[c]{@{}c@{}}Detection \\ accuracy$\uparrow$\end{tabular} & \begin{tabular}[c]{@{}c@{}}AUPR\\ in$\uparrow$\end{tabular} & \begin{tabular}[c]{@{}c@{}}AUPR\\ out$\uparrow$\end{tabular} \\ \cmidrule(r){1-5} \cmidrule(lr){7-10} \cmidrule(l){12-15} 
					Vanilla DNN~\cite{He_2016}                & 93.36          & 86.08                                                        & 92.82                                               & 93.71                                                &                      & 92.36          & 85.78                                                        & 93.81                                             & 89.87                                              &  & 96.53          & 91.60                                                         & 97.23                                             & 95.23                                              \\ \cmidrule(r){1-5} \cmidrule(lr){7-10} \cmidrule(l){12-15} 
					Temp scaling~\cite{guo2017calibration}            & 93.71          & 86.47                                                        & 93.21                                               & 94.01                                                &                      & 92.71          & 86.72                                                        & 94.21                                             & 90.11                                              &  & 96.65          & 92.14                                                         & 97.39                                             & 95.29                                              \\ \cmidrule(r){1-5} \cmidrule(lr){7-10} \cmidrule(l){12-15} 
					Ensemble~\cite{lakshminarayanan2017simple}                & 95.49          & 88.82                                                        & 95.31                                               & 95.64                                                &                      & 90.71          & 83.94                                                        & 92.55                                             & 87.68                                              &  & 95.78          & 91.47                                                         & 96.95                                             & 92.65                                              \\ \cmidrule(r){1-5} \cmidrule(lr){7-10} \cmidrule(l){12-15} 
					Dropout~\cite{gal2016dropout}              & 96.38          & 89.98                                                        & 96.16                                               & 96.67                                                &                      & 87.64          & 81.20                                                        & 89.83                                             & 83.13                                              &  & 91.48          & 86.84                                                         & 93.99                                             & 86.37                                              \\ \cmidrule(r){1-5} \cmidrule(lr){7-10} \cmidrule(l){12-15} 
					SVI~\cite{blundell2015weight}                     & 96.40          & 90.03                                                        & 95.97                                               & 96.83                                                &                      & 85.89          & 79.31                                                        & 88.34                                             & 81.48                                              &  & 93.94          & 87.87                                                         & 95.30                                             & 91.61                                              \\ \cmidrule(r){1-5} \cmidrule(lr){7-10} \cmidrule(l){12-15} 
					SVI-TS~\cite{blundell2015weight,guo2017calibration}                     & 96.61          & 90.45                                                        & 96.24                                               & 96.98                                                &                      & 81.08          & 75.43                                                        & 84.85                                             & 74.16                                              &  & 90.81          & 87.59                                                         & 93.84                                             & 82.18                                              \\ \cmidrule(r){1-5} \cmidrule(lr){7-10} \cmidrule(l){12-15}
					SVI-AvUTS                & 96.89          & 90.93                                                        & 96.58                                               & 97.19                                                &                      & 81.19          & 75.82                                                        & 85.09                                             & 74.17                                              &  & 93.79          & 89.39                                                         & 95.49                                             & 87.99                                              \\ \cmidrule(r){1-5} \cmidrule(lr){7-10} \cmidrule(l){12-15} 
					\textbf{SVI-AvUC}               & \textbf{97.60} & \textbf{92.07}                                               & \textbf{97.39}                                      & \textbf{97.85}                                       & \textbf{}            & \textbf{95.54} & \textbf{88.43}                                               & \textbf{96.32}                                    & \textbf{94.66}                                     &  & \textbf{99.35} & \textbf{97.16}                                                & \textbf{99.50}                                    & \textbf{98.91}                                     \\ \bottomrule 
				\end{tabular}
			\label{tab:shiftdetection}		
    \end{adjustbox} 
\end{table}

\vspace{-2mm}
\section{Conclusion}
\label{sec:related}
We introduced the \textit{accuracy versus uncertainty calibration} (AvUC) loss and proposed novel optimization methods AvUC and AvUTS for improving uncertainty calibration in deep neural networks. 
Uncertainty calibration is important for reliable and informed decision making in safety critical applications, we envision AvUC as a step towards advancing probabilistic deep neural networks in providing well-calibrated uncertainties along with improved accuracy. Our work shows that accounting for uncertainty estimation during training can improve model calibration significantly. We demonstrated our method SVI-AvUC provides better model calibration than existing state-of-the-art methods under distributional shift. We showed our simple post-hoc calibration method AvUTS can improve the uncertainty calibration over the baseline. We also demonstrated the effectiveness of proposed methods in detecting distributional shift while outperforming the other methods. We have illustrated AvUC and AvUTS on stochastic variational inference (Bayesian) and vanilla (non-Bayesian) methods. 
We have made the code {\footnote[3]{\url{https://github.com/IntelLabs/AVUC}}} available to facilitate probabilistic deep learning community to evaluate and improve model calibration for various other baselines.

\clearpage
\section*{Broader Impact}

As AI systems backed by deep learning are used in safety-critical applications like autonomous vehicles, medical diagnosis, robotics etc., it is important for these systems to be explainable and trustworthy for successful deployment in real-world. Having the ability to derive uncertainty estimates provides a big step towards explainability of AI systems based on Deep Learning. Having calibrated uncertainty quantification provides grounded means for uncertainty measurement in such models. A principled way to measure reliable uncertainty is the basis on which trustworthy AI systems can be built. Research results and multiple resulting frameworks have been released for AI Fairness measurement that base components of fairness quantification on uncertainty measurements of classified output of deep learning models. We believe that our work can be a big step towards measuring such uncertainties in a reliable fashion. The resulting, well calibrated, uncertainty measures can then be used as an input for building fair and trustworthy AI models that implement explainable behavior. This explanation is also critical for building AI systems that are robust to adversarial blackbox and whitebox attacks. These well calibrated uncertainties can guide AI practitioners to better understand the predictions for reliable decision making, i.e. to know “when to trust” and “when not to trust” the model predictions (especially in high-risk domains like healthcare, financial, legal etc). In addition, calibrated uncertainty opens the doors for wider adoption of deep network architectures in interesting applications like multimodal fusion, anomaly detection and active learning. Using calibrated uncertainty as a measure for distributional shift (out-of-distribution and dataset shift) detection is also a key enabler for self-learning systems that form a critical component of realizing the dream of Artificial General Intelligence (AGI).
 
\section*{Acknowledgement}
We would like to thank Mahesh Subedar (Intel Labs), Willem M Beltman (Intel Labs) and the reviewers for their comments and feedback that helped to improve the manuscript.

{
\small
\bibliographystyle{unsrtnat}
\bibliography{neurips_2020}

\begin{thebibliography}{55}
\providecommand{\natexlab}[1]{#1}
\providecommand{\url}[1]{\texttt{#1}}
\expandafter\ifx\csname urlstyle\endcsname\relax
  \providecommand{\doi}[1]{doi: #1}\else
  \providecommand{\doi}{doi: \begingroup \urlstyle{rm}\Url}\fi

\bibitem[Ghahramani(2015)]{ghahramani2015probabilistic}
Zoubin Ghahramani.
\newblock Probabilistic machine learning and artificial intelligence.
\newblock \emph{Nature}, 521\penalty0 (7553):\penalty0 452--459, 2015.

\bibitem[Graves(2011)]{graves2011practical}
Alex Graves.
\newblock Practical variational inference for neural networks.
\newblock In \emph{Advances in neural information processing systems}, pages
  2348--2356, 2011.

\bibitem[Blundell et~al.(2015)Blundell, Cornebise, Kavukcuoglu, and
  Wierstra]{blundell2015weight}
Charles Blundell, Julien Cornebise, Koray Kavukcuoglu, and Daan Wierstra.
\newblock Weight uncertainty in neural network.
\newblock In \emph{International Conference on Machine Learning}, pages
  1613--1622, 2015.

\bibitem[Kingma et~al.(2015)Kingma, Salimans, and
  Welling]{kingma2015variational}
Durk~P Kingma, Tim Salimans, and Max Welling.
\newblock Variational dropout and the local reparameterization trick.
\newblock In \emph{Advances in Neural Information Processing Systems}, pages
  2575--2583, 2015.

\bibitem[Gal and Ghahramani(2016)]{gal2016dropout}
Yarin Gal and Zoubin Ghahramani.
\newblock Dropout as a bayesian approximation: Representing model uncertainty
  in deep learning.
\newblock In \emph{international conference on machine learning}, pages
  1050--1059, 2016.

\bibitem[Maddox et~al.(2019)Maddox, Izmailov, Garipov, Vetrov, and
  Wilson]{maddox2019simple}
Wesley~J Maddox, Pavel Izmailov, Timur Garipov, Dmitry~P Vetrov, and
  Andrew~Gordon Wilson.
\newblock A simple baseline for bayesian uncertainty in deep learning.
\newblock In \emph{Advances in Neural Information Processing Systems}, pages
  13132--13143, 2019.

\bibitem[Rohekar et~al.(2019)Rohekar, Gurwicz, Nisimov, and
  Novik]{rohekar2019modeling}
Raanan~Yehezkel Rohekar, Yaniv Gurwicz, Shami Nisimov, and Gal Novik.
\newblock Modeling uncertainty by learning a hierarchy of deep neural
  connections.
\newblock In \emph{Advances in Neural Information Processing Systems}, pages
  4246--4256, 2019.

\bibitem[Farquhar et~al.(2020)Farquhar, Osborne, and Gal]{farquhar_radial_2020}
Sebastian Farquhar, Michael Osborne, and Yarin Gal.
\newblock Radial bayesian neural networks: Beyond discrete support in
  large-scale bayesian deep learning.
\newblock \emph{Proceedings of the 23rtd International Conference on Artificial
  Intelligence and Statistics}, 2020.

\bibitem[Lakshminarayanan et~al.(2017)Lakshminarayanan, Pritzel, and
  Blundell]{lakshminarayanan2017simple}
Balaji Lakshminarayanan, Alexander Pritzel, and Charles Blundell.
\newblock Simple and scalable predictive uncertainty estimation using deep
  ensembles.
\newblock In \emph{Advances in neural information processing systems}, pages
  6402--6413, 2017.

\bibitem[Lee et~al.(2018)Lee, Lee, Lee, and Shin]{lee2018simple}
Kimin Lee, Kibok Lee, Honglak Lee, and Jinwoo Shin.
\newblock A simple unified framework for detecting out-of-distribution samples
  and adversarial attacks.
\newblock In \emph{Advances in Neural Information Processing Systems}, pages
  7167--7177, 2018.

\bibitem[Guo et~al.(2017)Guo, Pleiss, Sun, and Weinberger]{guo2017calibration}
Chuan Guo, Geoff Pleiss, Yu~Sun, and Kilian~Q Weinberger.
\newblock On calibration of modern neural networks.
\newblock In \emph{Proceedings of the 34th International Conference on Machine
  Learning-Volume 70}, pages 1321--1330. JMLR. org, 2017.

\bibitem[Kumar et~al.(2018)Kumar, Sarawagi, and Jain]{kumar2018trainable}
Aviral Kumar, Sunita Sarawagi, and Ujjwal Jain.
\newblock Trainable calibration measures for neural networks from kernel mean
  embeddings.
\newblock In \emph{International Conference on Machine Learning}, pages
  2805--2814, 2018.

\bibitem[Mukhoti et~al.(2020)Mukhoti, Kulharia, Sanyal, Golodetz, Torr, and
  Dokania]{mukhoti2020calibrating}
Jishnu Mukhoti, Viveka Kulharia, Amartya Sanyal, Stuart Golodetz, Philip~HS
  Torr, and Puneet~K Dokania.
\newblock Calibrating deep neural networks using focal loss.
\newblock \emph{arXiv preprint arXiv:2002.09437}, 2020.

\bibitem[Kuleshov et~al.(2018)Kuleshov, Fenner, and Ermon]{kuleshov18accurate}
Volodymyr Kuleshov, Nathan Fenner, and Stefano Ermon.
\newblock Accurate uncertainties for deep learning using calibrated regression.
\newblock In \emph{Proceedings of the 35th International Conference on Machine
  Learning}, volume~80 of \emph{Proceedings of Machine Learning Research},
  pages 2796--2804. PMLR, 2018.

\bibitem[Foong et~al.(2019)Foong, Li, Hern{\'a}ndez-Lobato, and
  Turner]{foong2019between}
Andrew~YK Foong, Yingzhen Li, Jos{\'e}~Miguel Hern{\'a}ndez-Lobato, and
  Richard~E Turner.
\newblock 'in-between'uncertainty in bayesian neural networks.
\newblock \emph{arXiv preprint arXiv:1906.11537}, 2019.

\bibitem[Heek(2018)]{heekwell}
Jonathan Heek.
\newblock Well-calibrated bayesian neural networks.
\newblock \emph{University of Cambridge}, 2018.

\bibitem[Kumar et~al.(2019)Kumar, Liang, and Ma]{kumar2019verified}
Ananya Kumar, Percy~S Liang, and Tengyu Ma.
\newblock Verified uncertainty calibration.
\newblock In \emph{Advances in Neural Information Processing Systems}, pages
  3787--3798, 2019.

\bibitem[Kull et~al.(2019)Kull, Nieto, K{\"a}ngsepp, Silva~Filho, Song, and
  Flach]{kull2019beyond}
Meelis Kull, Miquel~Perello Nieto, Markus K{\"a}ngsepp, Telmo Silva~Filho, Hao
  Song, and Peter Flach.
\newblock Beyond temperature scaling: Obtaining well-calibrated multi-class
  probabilities with dirichlet calibration.
\newblock In \emph{Advances in Neural Information Processing Systems}, pages
  12295--12305, 2019.

\bibitem[Thulasidasan et~al.(2019)Thulasidasan, Chennupati, Bilmes,
  Bhattacharya, and Michalak]{thulasidasan2019mixup}
Sunil Thulasidasan, Gopinath Chennupati, Jeff~A Bilmes, Tanmoy Bhattacharya,
  and Sarah Michalak.
\newblock On mixup training: Improved calibration and predictive uncertainty
  for deep neural networks.
\newblock In \emph{Advances in Neural Information Processing Systems}, pages
  13888--13899, 2019.

\bibitem[Hendrycks and Dietterich(2019)]{hendrycks2019robustness}
Dan Hendrycks and Thomas Dietterich.
\newblock Benchmarking neural network robustness to common corruptions and
  perturbations.
\newblock \emph{Proceedings of the International Conference on Learning
  Representations}, 2019.

\bibitem[Hendrycks et~al.(2020)Hendrycks, Mu, Cubuk, Zoph, Gilmer, and
  Lakshminarayanan]{hendrycks*2020augmix}
Dan Hendrycks, Norman Mu, Ekin~Dogus Cubuk, Barret Zoph, Justin Gilmer, and
  Balaji Lakshminarayanan.
\newblock Augmix: A simple method to improve robustness and uncertainty under
  data shift.
\newblock In \emph{International Conference on Learning Representations}, 2020.

\bibitem[Moreno-Torres et~al.(2012)Moreno-Torres, Raeder, Alaiz-Rodr{\'\i}Guez,
  Chawla, and Herrera]{moreno2012unifying}
Jose~G Moreno-Torres, Troy Raeder, Roc{\'\i}O Alaiz-Rodr{\'\i}Guez, Nitesh~V
  Chawla, and Francisco Herrera.
\newblock A unifying view on dataset shift in classification.
\newblock \emph{Pattern recognition}, 45\penalty0 (1):\penalty0 521--530, 2012.

\bibitem[Alcorn et~al.(2019)Alcorn, Li, Gong, Wang, Mai, Ku, and
  Nguyen]{alcorn2019strike}
Michael~A Alcorn, Qi~Li, Zhitao Gong, Chengfei Wang, Long Mai, Wei-Shinn Ku,
  and Anh Nguyen.
\newblock Strike (with) a pose: Neural networks are easily fooled by strange
  poses of familiar objects.
\newblock In \emph{Proceedings of the IEEE Conference on Computer Vision and
  Pattern Recognition}, pages 4845--4854, 2019.

\bibitem[Blum et~al.(2019)Blum, Sarlin, Nieto, Siegwart, and
  Cadena]{blum2019fishyscapes}
Hermann Blum, Paul-Edouard Sarlin, Juan Nieto, Roland Siegwart, and Cesar
  Cadena.
\newblock The fishyscapes benchmark: measuring blind spots in semantic
  segmentation.
\newblock \emph{arXiv preprint arXiv:1904.03215}, 2019.

\bibitem[Amodei et~al.(2016)Amodei, Olah, Steinhardt, Christiano, Schulman, and
  Man{\'e}]{amodei2016concrete}
Dario Amodei, Chris Olah, Jacob Steinhardt, Paul Christiano, John Schulman, and
  Dan Man{\'e}.
\newblock Concrete problems in ai safety.
\newblock \emph{arXiv preprint arXiv:1606.06565}, 2016.

\bibitem[Snoek et~al.(2019)Snoek, Ovadia, Fertig, Lakshminarayanan, Nowozin,
  Sculley, Dillon, Ren, and Nado]{snoek2019can}
Jasper Snoek, Yaniv Ovadia, Emily Fertig, Balaji Lakshminarayanan, Sebastian
  Nowozin, D~Sculley, Joshua Dillon, Jie Ren, and Zachary Nado.
\newblock Can you trust your model's uncertainty? evaluating predictive
  uncertainty under dataset shift.
\newblock In \emph{Advances in Neural Information Processing Systems}, pages
  13969--13980, 2019.

\bibitem[Lacoste-Julien et~al.(2011)Lacoste-Julien, Husz{\'a}r, and
  Ghahramani]{lacoste2011approximate}
Simon Lacoste-Julien, Ferenc Husz{\'a}r, and Zoubin Ghahramani.
\newblock Approximate inference for the loss-calibrated bayesian.
\newblock In \emph{Proceedings of the Fourteenth International Conference on
  Artificial Intelligence and Statistics}, pages 416--424, 2011.

\bibitem[Cobb et~al.(2018)Cobb, Roberts, and Gal]{cobb2018loss}
Adam~D Cobb, Stephen~J Roberts, and Yarin Gal.
\newblock Loss-calibrated approximate inference in bayesian neural networks.
\newblock \emph{arXiv preprint arXiv:1805.03901}, 2018.

\bibitem[Berger(1985)]{berger1985statistical}
James~O Berger.
\newblock \emph{Statistical Decision Theory and Bayesian Analysis}.
\newblock Springer, 1985.

\bibitem[Welling and Teh(2011)]{welling2011bayesian}
Max Welling and Yee~W Teh.
\newblock Bayesian learning via stochastic gradient langevin dynamics.
\newblock In \emph{Proceedings of the 28th international conference on machine
  learning (ICML-11)}, pages 681--688, 2011.

\bibitem[Chen et~al.(2014)Chen, Fox, and Guestrin]{chen2014stochastic}
Tianqi Chen, Emily Fox, and Carlos Guestrin.
\newblock Stochastic gradient hamiltonian monte carlo.
\newblock In \emph{International conference on machine learning}, pages
  1683--1691, 2014.

\bibitem[Smith and Gal(2018)]{smith2018understanding}
Lewis Smith and Yarin Gal.
\newblock Understanding measures of uncertainty for adversarial example
  detection.
\newblock \emph{arXiv preprint arXiv:1803.08533}, 2018.

\bibitem[Der~Kiureghian and Ditlevsen(2009)]{der2009aleatory}
Armen Der~Kiureghian and Ove Ditlevsen.
\newblock Aleatory or epistemic? does it matter?
\newblock \emph{Structural safety}, 31\penalty0 (2):\penalty0 105--112, 2009.

\bibitem[Kendall and Gal(2017)]{kendall2017uncertainties}
Alex Kendall and Yarin Gal.
\newblock What uncertainties do we need in bayesian deep learning for computer
  vision?
\newblock In \emph{Advances in neural information processing systems}, pages
  5574--5584, 2017.

\bibitem[Gal(2016)]{gal2016uncertainty}
Yarin Gal.
\newblock Uncertainty in deep learning.
\newblock \emph{PhD thesis, University of Cambridge}, 2016.

\bibitem[Shannon(1948)]{shannon1948mathematical}
Claude~E Shannon.
\newblock A mathematical theory of communication.
\newblock \emph{Bell system technical journal}, 27\penalty0 (3):\penalty0
  379--423, 1948.

\bibitem[Freeman(1965)]{freeman1965}
Linton~G Freeman.
\newblock \emph{Elementary applied statistics}.
\newblock John Wiley and Sons, 1965.

\bibitem[Houlsby et~al.(2011)Houlsby, Husz{\'a}r, Ghahramani, and
  Lengyel]{houlsby2011bayesian}
Neil Houlsby, Ferenc Husz{\'a}r, Zoubin Ghahramani, and M{\'a}t{\'e} Lengyel.
\newblock Bayesian active learning for classification and preference learning.
\newblock \emph{arXiv preprint arXiv:1112.5745}, 2011.

\bibitem[Mukhoti and Gal(2018)]{mukhoti2018evaluating}
Jishnu Mukhoti and Yarin Gal.
\newblock Evaluating bayesian deep learning methods for semantic segmentation.
\newblock \emph{arXiv preprint arXiv:1811.12709}, 2018.

\bibitem[Naeini et~al.(2015)Naeini, Cooper, and
  Hauskrecht]{naeini2015obtaining}
Mahdi~Pakdaman Naeini, Gregory Cooper, and Milos Hauskrecht.
\newblock Obtaining well calibrated probabilities using bayesian binning.
\newblock In \emph{Twenty-Ninth AAAI Conference on Artificial Intelligence},
  2015.

\bibitem[Laves et~al.(2019)Laves, Ihler, Kortmann, and Ortmaier]{laves2019well}
Max-Heinrich Laves, Sontje Ihler, Karl-Philipp Kortmann, and Tobias Ortmaier.
\newblock Well-calibrated model uncertainty with temperature scaling for
  dropout variational inference.
\newblock \emph{arXiv preprint arXiv:1909.13550}, 2019.

\bibitem[Gneiting and Raftery(2007)]{gneiting2007strictly}
Tilmann Gneiting and Adrian~E Raftery.
\newblock Strictly proper scoring rules, prediction, and estimation.
\newblock \emph{Journal of the American statistical Association}, 102\penalty0
  (477):\penalty0 359--378, 2007.

\bibitem[Brier(1950)]{brier1950verification}
Glenn~W Brier.
\newblock Verification of forecasts expressed in terms of probability.
\newblock \emph{Monthly weather review}, 78\penalty0 (1):\penalty0 1--3, 1950.

\bibitem[Davis and Goadrich(2006)]{davis2006relationship}
Jesse Davis and Mark Goadrich.
\newblock The relationship between precision-recall and roc curves.
\newblock In \emph{Proceedings of the 23rd international conference on Machine
  learning}, pages 233--240, 2006.

\bibitem[Saito and Rehmsmeier(2015)]{saito2015precision}
Takaya Saito and Marc Rehmsmeier.
\newblock The precision-recall plot is more informative than the roc plot when
  evaluating binary classifiers on imbalanced datasets.
\newblock \emph{PloS one}, 10\penalty0 (3), 2015.

\bibitem[He et~al.(2016)He, Zhang, Ren, and Sun]{He_2016}
Kaiming He, Xiangyu Zhang, Shaoqing Ren, and Jian Sun.
\newblock Deep residual learning for image recognition.
\newblock \emph{IEEE Conference on Computer Vision and Pattern Recognition
  (CVPR)}, 2016.

\bibitem[Deng et~al.(2009)Deng, Dong, Socher, Li, Li, and
  Fei-Fei]{deng2009imagenet}
Jia Deng, Wei Dong, Richard Socher, Li-Jia Li, Kai Li, and Li~Fei-Fei.
\newblock Imagenet: A large-scale hierarchical image database.
\newblock In \emph{2009 IEEE conference on computer vision and pattern
  recognition}, pages 248--255. Ieee, 2009.

\bibitem[Krizhevsky(2009)]{krizhevsky2009learning}
Alex Krizhevsky.
\newblock Learning multiple layers of features from tiny images.
\newblock 2009.

\bibitem[Riquelme et~al.(2018)Riquelme, Tucker, and Snoek]{riquelme2018deep}
Carlos Riquelme, George Tucker, and Jasper Snoek.
\newblock Deep bayesian bandits showdown: An empirical comparison of bayesian
  deep networks for thompson sampling.
\newblock In \emph{International Conference on Learning Representations}, 2018.

\bibitem[Subedar et~al.(2019)Subedar, Krishnan, Meyer, Tickoo, and
  Huang]{Subedar_2019_ICCV}
Mahesh Subedar, Ranganath Krishnan, Paulo~Lopez Meyer, Omesh Tickoo, and
  Jonathan Huang.
\newblock Uncertainty-aware audiovisual activity recognition using deep
  bayesian variational inference.
\newblock In \emph{Proceedings of the IEEE/CVF International Conference on
  Computer Vision (ICCV)}, 2019.

\bibitem[Krishnan et~al.(2020)Krishnan, Subedar, and
  Tickoo]{krishnanspecifying}
Ranganath Krishnan, Mahesh Subedar, and Omesh Tickoo.
\newblock Specifying weight priors in bayesian deep neural networks with
  empirical bayes.
\newblock \emph{Thirty-Fourth AAAI Conference on Artificial Intelligence},
  2020.

\bibitem[Netzer et~al.(2011)Netzer, Wang, Coates, Bissacco, Wu, and
  Ng]{netzer2011reading}
Yuval Netzer, Tao Wang, Adam Coates, Alessandro Bissacco, Bo~Wu, and Andrew~Y
  Ng.
\newblock Reading digits in natural images with unsupervised feature learning.
\newblock 2011.

\bibitem[Ramdas et~al.(2017)Ramdas, Trillos, and Cuturi]{ramdas2017wasserstein}
Aaditya Ramdas, Nicol{\'a}s~Garc{\'\i}a Trillos, and Marco Cuturi.
\newblock On wasserstein two-sample testing and related families of
  nonparametric tests.
\newblock \emph{Entropy}, 19\penalty0 (2):\penalty0 47, 2017.

\bibitem[Paszke et~al.(2019)Paszke, Gross, Massa, Lerer, Bradbury, Chanan,
  Killeen, Lin, Gimelshein, Antiga, et~al.]{paszke2019pytorch}
Adam Paszke, Sam Gross, Francisco Massa, Adam Lerer, James Bradbury, Gregory
  Chanan, Trevor Killeen, Zeming Lin, Natalia Gimelshein, Luca Antiga, et~al.
\newblock Pytorch: An imperative style, high-performance deep learning library.
\newblock In \emph{Advances in neural information processing systems}, pages
  8026--8037, 2019.

\bibitem[Kokoska and Zwillinger(2000)]{kokoska2000crc}
Stephen Kokoska and Daniel Zwillinger.
\newblock \emph{CRC standard probability and statistics tables and formulae}.
\newblock Crc Press, 2000.

\end{thebibliography}
}

\clearpage
\appendix

\section*{\centering \Large Appendix: Improving model calibration with accuracy versus uncertainty optimization}

\renewcommand{\thetable}{T\arabic{table}}  
\renewcommand{\thefigure}{F\arabic{figure}}
\setcounter{figure}{0}
\setcounter{table}{0}


\section{Dataset shift}
\label{appdx:datasetshift}

\begin{figure}[h]
	\small
	\begin{subfigure}{\textwidth}
		\centering
		\captionsetup{
			justification=centering}
		\includegraphics[width=0.95\linewidth]{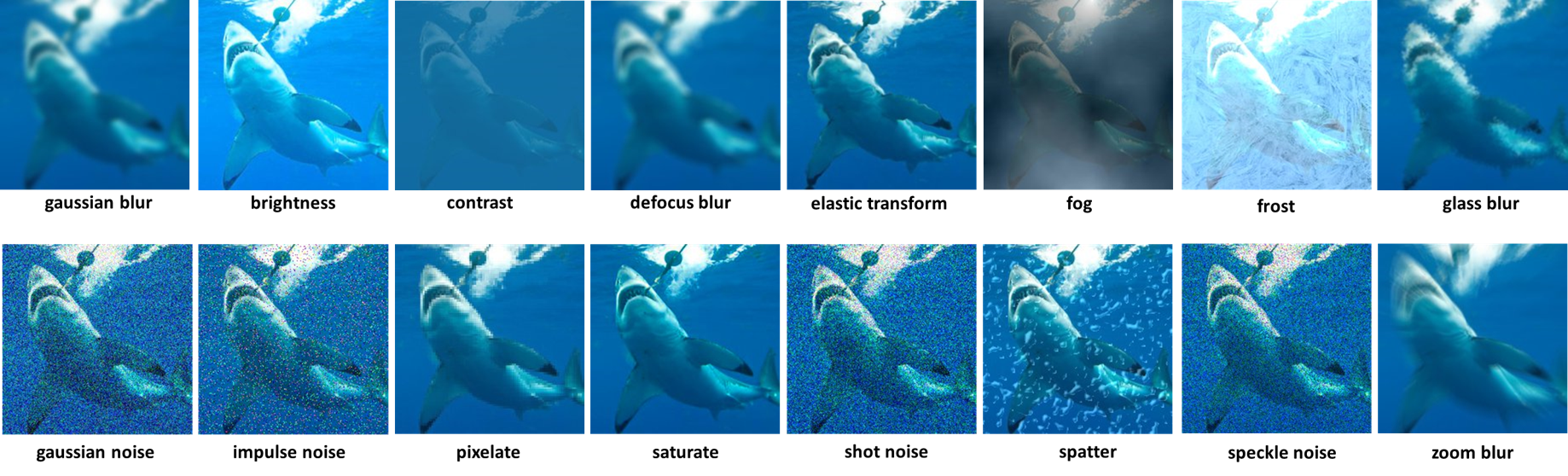}
	\end{subfigure}
	\caption{Example of sixteen different image corruptions~\cite{hendrycks2019robustness} used during test time (dataset shift) }
	\label{fig:image_corruptions}
	
\end{figure}

\begin{figure}[h]
	\small
	\begin{subfigure}{\textwidth}
		\centering
		\captionsetup{
			justification=centering}
		\includegraphics[scale=0.32]{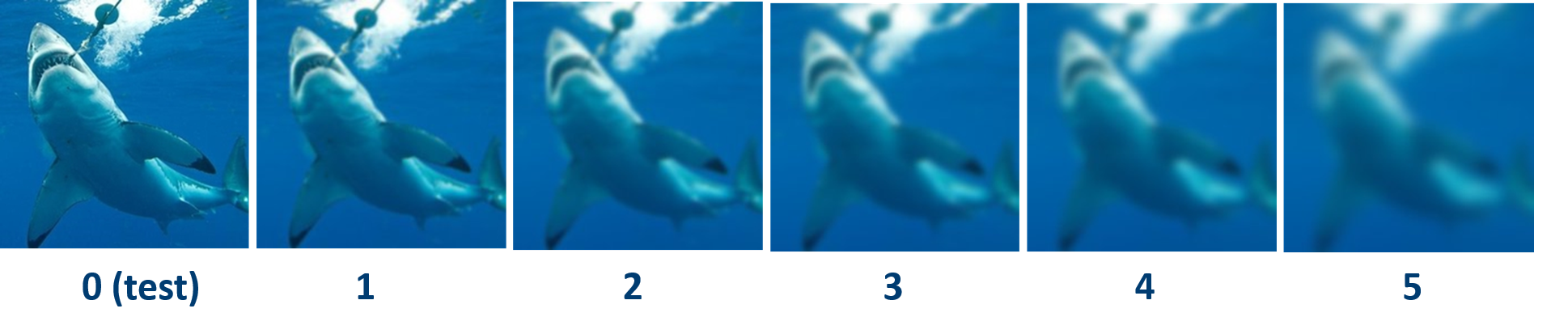}
	\end{subfigure}
	\caption{Example of Gaussian blur at different levels of shift intensity (1-5) }
	\label{fig:corruption_intensity}
	
\end{figure}

We use various image corruptions and perturbations proposed by \citet{hendrycks2019robustness} for evaluating model calibration under dataset shift, following the methodology in uncertainty quantification (UQ) benchmark~\cite{snoek2019can}. We evaluate our proposed methods with the high performing baselines provided in the UQ benchmark. For dataset shift evaluation, 16 different types of image corruptions at 5 different levels of intensities are utilized, resulting in 80 variants of datashift. Figure~\ref{fig:image_corruptions} shows an example of 16 different datashift types on ImageNet used in our experiments during test time. Figure~\ref{fig:corruption_intensity} shows an example of different shift intensities (from level 1 to 5) for Gaussian blur. The same datashifts are applied to CIFAR10 as well. These dataset shifts are encountered during test time only, the models are trained with clean data.
 
\section{Experimental details and Parameters}
\label{appdx:experiments}
\paragraph{Codebase} We have made our code available open-source at \url{https://github.com/intelLabs/AVUC}. We have implemented the code necessary for our experiments of SVI (\textit{mean-field stochastic variational inference}), SVI-AvUC (\textit{accuracy vs uncertainty calibration}) and SVI-AvUTS (\textit{accuracy vs uncertainty temperature scaling}) in PyTorch~\cite{paszke2019pytorch}, including AvUC loss and variational layers support required for stochastic variational inference. 

\subsection{Model details}
In this section we describe all hyper-parameters used for training the models and evaluation we performed in Section~\ref{sec:results}. On CIFAR10 and ImageNet image classification tasks under distributional shift, we use ResNet-20 and ResNet-50~\cite{He_2016} architectures respectively. 
The results for the methods: Vanilla, Temp scaling, Ensemble, Dropout, LL Dropout and LL SVI are computed from the model predictions provided in UQ benchmark~\cite{snoek2019can}.

\subsubsection{CIFAR10/ResNet-20}

\paragraph{SVI-AvUC} We use the same hyper-parameters as ~\citet{snoek2019can} 
used for SVI on CIFAR10 for fair comparison. The models were trained with Adam optimizer for $\mathrm{200}$ epochs with initial learning rate of $\mathrm{1.189e^{-3}}$ and batch size of $\mathrm{107}$. As part of the learning rate schedule, initial learning rate was multiplied by 0.1, 0.01, 0.001 and 0.0005 at epochs 80, 120, 160 and 180 respectively. The training samples were distorted with random horizontal flips and random crops with 4-pixel padding as mentioned in~\cite{He_2016}. We used $\beta=3$ in Equation~\ref{eqn:total_loss} for relative weighting of AvUC loss with respect to ELBO loss. We used $128$ Monte Carlo samples from weight posterior for evaluation.

\paragraph{SVI-AvUTS} We find the optimal temperature for pretrained SVI model by minimizing the \textit{accuracy versus uncertainty calibration} (AvUC) loss on hold-out validation data. We adapted the code from~\cite{guo2017calibration} and replaced negative log-likelihood loss with our AvUC loss implementation for optimization at learning rate of 0.005.
The CIFAR10 training data was split into 9:1 ratio (45k train set and 5k hold-out validation set images). The SVI baseline model was trained with same hyper-parameters as in UQ benchmark~\cite{snoek2019can}, described above.  

\paragraph{Radial BNN} To compare our methods SVI-AvUC and SVI-AvUTS with Radial BNN, we implemented ResNet-20 for Radial BNN adapting the code from~\cite{farquhar_radial_2020}. The models were trained with Adam optimizer for $\mathrm{200}$ epochs with initial learning rate of $\mathrm{1e^{-3}}$ and batch size of $\mathrm{256}$. As part of the learning rate schedule, initial learning rate was multiplied by 0.1, 0.01, 0.001 and 0.0005 at epochs 80, 120, 160 and 180 respectively. The training samples were distorted with random horizontal flips and random crops with 4-pixel padding as mentioned in~\cite{He_2016} 

We evaluate with 10k test images, along with 80 variants of dataset shift (each with 10k images) that includes 16 different types of datashift at 5 different intensities as described in Section~\ref{appdx:datasetshift}.

For out-of-distribution (OOD) evaluation, we use SVHN dataset as OOD data on models trained with CIFAR10.

\subsubsection{ImageNet/ResNet-50}

\paragraph{SVI} In order to scale SVI to large-scale ImageNet dataset and ResNet-50 model, we specify the weight priors and initialize the variational parameters using Empirical Bayes method as proposed in~\cite{krishnanspecifying}. The weights are modeled with fully factorized Gaussian distributions represented by $\mu$ and $\sigma$. In order to ensure non-negative variance, $\sigma$ is expressed in terms of softplus function with unconstrained parameter $\rho$, i.e. $\sigma=\log(1+exp(\rho))$. The weight prior is set to $\mathcal{N}(\mathrm{w_{\scalebox{.7}{$\scriptscriptstyle MLE$}}}, I)$ and the variational parameters $\mu$ and $\rho$ are initialized with $\mathrm{w}_{\scalebox{.7}{$\scriptscriptstyle MLE$}}$ and $\log (e^{\delta \mid \mathrm{w}_{\scalebox{.7}{$\scriptscriptstyle MLE$}} \mid} -1)$ repectively. The initial maximum likelihood estimate (MLE) for weights $\mathrm{w_{\scalebox{.7}{$\scriptscriptstyle MLE$}}}$ are obtained from pretained ResNet-50 model available in the torchvision package{\footnote{\tiny \url{ https://github.com/pytorch/vision/blob/master/torchvision/models/resnet.py}}} and $\delta$ is set to 0.5. The model was trained for 50 epochs using SGD optimizer with initial learning rate of $0.001$, momentum of 0.9, weight decay of $1e^{-4}$ and batch size of $96$. We used learning rate schedule that multiplies the learning rate by 0.1 every 30 epochs.The training samples were distorted with random horizontal flips and random crops as mentioned in~\cite{He_2016}. We used $128$ Monte Carlo (MC) samples from weight posterior for fair comparison with other stochastic methods in UQ benchmark~\cite{snoek2019can}, but we were able to get similar results with reduced number of MC samples.

\paragraph{SVI-AvUC} The model is trained with the same hyper-parameters and initializations with Empirical Bayes as described for SVI above, except that the model is trained with AvUC loss in addition to the ELBO loss. We used $\beta=3$ in Equation~\ref{eqn:total_loss} for relative weighting of AvUC loss with respect to ELBO loss.

\paragraph{SVI-AvUTS} We find the optimal temperature for pretrained SVI model by minimizing the \textit{accuracy versus uncertainty calibration} (AvUC) loss on hold-out validation data. We adapted the code from~\cite{guo2017calibration} and replaced negative log-likelihood loss with our AvUC loss implementation. We used 50k images (randomly sampled from 1281.1k training images) for finding the optimal temperature to modify the logits of pretrained SVI. We used $128$ Monte Carlo samples from weight posterior for evaluation.

\paragraph{AvUTS} We applied AvUTS (AvU Temperature Scaling) method on pretrained vanilla ResNet-50 model with AvUC loss in order to compare with conventional  temperature scaling~\cite{guo2017calibration} that optimizes negative log-likelihood loss. Results are provided in Appendix~\ref{appdx:avutssec}. We used the pretrained model available in the torchvision package\textsuperscript. We used entropy of softmax as uncertainty for AvUC loss computation. We followed the same procedure as SVI-AvUTS described above, except that the method is applied to deterministic model.

We evaluate with 50k test images, along with 80 variants of dataset shift (each with 50k images) that includes 16 different types of datashift at 5 different intensities as described in Section~\ref{appdx:datasetshift}.

\section{Additional background}
\label{appdx:background}
In this section, we follow the same notations described in Section~\ref{sec:avuloss} of the main paper.
\subsection{SVI in Bayesian deep neural networks}

Bayesian deep neural networks provide a probabilistic interpretation of deep learning models by learning probability distributions over the neural network weights. In Bayesian setting, we would like to infer a distribution over weights $\mathrm{w}$. A prior distribution is assumed over the weights $p(\mathrm{w})$ that captures our prior belief as to which parameters would have likely generated the outputs before observing any data. Given the evidence data $p(\mathrm{y}|\mathrm{x})$, prior distribution $p(\mathrm{w})$ and model likelihood $p(\mathrm{y}\,|\,\mathrm{x}, \mathrm{w})$, the goal is to infer the posterior distribution over the weights $p(\mathrm{w}|\mathrm{D})$:
\begin{equation}
p(\mathrm{w}|\mathrm{D}) = \frac{p(\mathrm{y}\,|\,\mathrm{x},\mathrm{w})\,p(\mathrm{w})}{\int{p(\mathrm{y}\,|\,\mathrm{x},\mathrm{w})\,p(\mathrm{w})}\,\mathrm{dw}}
\label{eq:bayes}
\end{equation}

Computing the posterior distribution $p(\mathrm{w}|\mathrm{D})$ is analytically intractable, stochastic variational inference (SVI)~\cite{graves2011practical,blundell2015weight,kingma2015variational} is an approximate method that has been proposed to achieve tractable inference. SVI approximates a complex probability distribution $p(\mathrm{w}|\mathrm{D})$ with a simpler distribution $q_\theta(\mathrm{w})$, parameterized by variational parameters $\theta$ while minimizing the Kullback-Leibler (KL) divergence. Minimizing the KL divergence is equivalent to maximizing the log evidence lower bound~(ELBO)~\cite{gal2016dropout}, as given by Equation~\ref{eqn:ELBO}. Conventionally ELBO loss (negative ELBO) as given by Equation~\ref{eqnsupp:ELBO_loss} is mizimized while training Bayesian deep neural networks with stochastic gradient descent optimization.

\begin{equation}
\label{eqn:ELBO}
\begin{aligned}
\mathbb{L} := \mathbb{E}_{q_\theta(\mathrm{w})}\left[\log\,p(\mathrm{y}|\mathrm{x},\mathrm{w})\right] - KL[q_\theta(\mathrm{w})||p(\mathrm{w})]
\end{aligned}
\end{equation}

\begin{equation}
\label{eqnsupp:ELBO_loss}
\begin{aligned}
\mathcal{L}_{\textrm{ELBO}} := -\mathbb{E}_{q_\theta(\mathrm{w})}\left[\log\,p(\mathrm{y}|\mathrm{x},\mathrm{w})\right] + KL[q_\theta(\mathrm{w})||p(\mathrm{w})]
\end{aligned}
\end{equation}

In mean-field stochastic variation inference, weights are modeled with fully factorized Gaussian distribution parameterized by variational parameters $\mu$ and $\sigma$.
\begin{equation}
q_\theta(\mathrm{w})=\mathcal{N}(\mathrm{w}\,|\,\mu,\sigma)
\label{eq:posterior}
\end{equation}
The variational distribution $q_\theta(\mathrm{w})$ and its parameters $\mu$ and $\sigma$  are learned while optimizing the cost function ELBO with the stochastic gradient steps.

\subsection{Uncertainty metrics}

Predictive distribution is obtained through multiple stochastic forward passes on the network while sampling from the weight posteriors using Monte Carlo estimators.  Equation~\ref{eq:pred_dist} shows the predictive distribution of the output $\mathrm{y}$ given input $\mathrm{x}$: 
\begin{equation}
\begin{gathered}
p(\mathrm{y}|\mathrm{x},\mathrm{D}) \approx \frac{1}{T} \sum_{t=1}^{T}p(\mathrm{y}|\mathrm{x},\mathrm{w_t})~,~~\mathrm{w_t}\sim p(\mathrm{w}\,|\,\mathrm{D})
\end{gathered}
\label{eq:pred_dist}
\end{equation}

\paragraph{Predictive entropy} The entropy~\cite{shannon1948mathematical} of the predictive distribution captures a combination of aleatoric and epistemic uncertainties~\cite{mukhoti2018evaluating} given by Equation~\ref{eqn:pe}~\cite{gal2016uncertainty}.

\begin{equation}
\label{eqn:pe}
\mathbb{H}(\mathrm{y}|\mathrm{x},\mathrm{D}):=-\sum_{k}\left(\frac{1}{T} \sum_{t=1}^{T} p\left(\mathrm{y}=k | \mathrm{x}, \mathrm{w}_{t}\right)\right) \log \left(\frac{1}{T} \sum_{t=1}^{T} p\left(\mathrm{y}=k | \mathrm{x}, \mathrm{w}_{t}\right)\right)
\end{equation}

For deterministic models (Vanilla, Temp scaling), predictive entropy is computed with Equation~\ref{eqn:pe_det}.

\begin{equation}
\label{eqn:pe_det}
\mathbb{H}(\mathrm{y}|\mathrm{x},\mathrm{D}):=-\sum_{k}\left( p\left(\mathrm{y}=k | \mathrm{x}, \mathrm{w}\right)\right) \log \left( p\left(\mathrm{y}=k | \mathrm{x}, \mathrm{w}\right)\right)
\end{equation}

\paragraph{Mutual information} The mutual information~\cite{shannon1948mathematical} between weight posterior and predictive distribution captures the epistemic uncertainty~\cite{houlsby2011bayesian,gal2016uncertainty} given by Equation~\ref{eq:mutualinformation}.

\begin{equation}
MI(\mathrm{y},\mathrm{w|x}, D) := \mathbb{H}(\mathrm{y}|\mathrm{x},\mathrm{D}) -\mathbb{E}_{p(\mathrm{w}|D)}\left[\mathbb{H}(\mathrm{y|x}, \mathrm{w})\right]\\
\label{eq:mutualinformation}
\end{equation}

\subsection{Evaluation metrics}
\label{appdx:evalmetrics}
\subsubsection{Model calibration evaluation metrics}

Expected calibration error (ECE)~\cite{naeini2015obtaining} measures the difference in expectation between model accuracy and its confidence as defined in Equation~\ref{eq:ece}. ECE quantifies the model miscalibration with respect to confidence (probability of predicted class).  The predictions of the neural network is partitioned into L bins of equal width, where $l^{th}$ bin is the interval $\left(\frac{l-1}{L}, \frac{l}{L}\right]$. ECE is computed using the equation below, where N is the total number of samples and $B_l$ is the set of indices of samples whose prediction confidence falls into the $l^{th}$ bin.

\begin{equation}
\label{eq:ece}
\mathrm{ECE}=\sum_{l=1}^{L} \frac{\left|B_{l}\right|}{N}\left|\operatorname{acc}\left(B_{l}\right)-\operatorname{conf}\left(B_{l}\right)\right|
\end{equation}

where the model accuracy and confidence per bin are defined as below. 

\begin{equation}
\operatorname{acc}\left(B_{l}\right)=\frac{1}{\left|B_{l}\right|} \sum_{i \in B_{l}} \mathbbm{1}\left(\widehat{\mathrm{y}}_{i}=\mathrm{y}_{i}\right) \quad\quad;\quad\quad \operatorname{conf}\left(B_{l}\right)=\frac{1}{\left|B_{l}\right|} \sum_{i \in B_{l}} {p}_{i}
\end{equation}

Expected uncertainty calibration error (UCE)~\cite{laves2019well} measures the difference in expectation between model error and its uncertainty as defined in Equation~\ref{eq:uce}. UCE quantifies the model miscalibration with respect to predictive uncertainty representing entire predictive distribution of probabilities across the classes.

\begin{equation}
\label{eq:uce}
\mathrm{UCE}=\sum_{l=1}^{L} \frac{\left|B_{l}\right|}{N}\left|\operatorname{err}\left(B_{l}\right)-\operatorname{uncert}\left(B_{l}\right)\right|
\end{equation}

where the model error and uncertainty per bin are defined as below. $\tilde{\mathrm{u}}_{i}\in[0,1]$ represents normalized uncertainty.
\begin{equation}
\operatorname{err}\left(B_{l}\right)=\frac{1}{\left|B_{l}\right|} \sum_{i \in B_{l}} \mathbbm{1}\left(\widehat{\mathrm{y}}_{i} \neq \mathrm{y}_{i}\right) \quad\quad;\quad\quad \operatorname{uncert}\left(B_{l}\right)=\frac{1}{\left|B_{l}\right|} \sum_{i \in B_{l}} \tilde{\mathrm{u}}_{i}
\end{equation}

\subsubsection{Uncertainty evaluation metrics}
Conditional probabilities {\textit{p(\text{accurate }|\text{ certain})}} and {\textit{p(\text{uncertain }|\text { inaccurate})}} have been proposed in~\cite{mukhoti2018evaluating} as model performance evaluation metrics for comparing the quality of uncertainty estimates obtained from different probabilistic methods. {\textit{p(\text{accurate }|\text{ certain})}} is given by Equation~\ref{eq:p_ac}, measures the probability that the model is accurate on its output given that it is confident on the same. {\textit{p(\text{uncertain }|\text { inaccurate})}} is given by Equation~\ref{eq:p_ui}, measures the probability that the model is uncertain about its output given that it has made inaccurate prediction.

\begin{equation}
\label{eq:p_ac}
p(\mathrm{accurate | certain}) = \frac{\mathrm{n}_{AC}}{\mathrm{n}_{AC}+\mathrm{n}_{IC}}
\end{equation}

\begin{equation}
\label{eq:p_ui}
p(\mathrm{uncertain | inaccurate}) = \frac{\mathrm{n}_{IU}}{\mathrm{n}_{IC}+\mathrm{n}_{IU}}
\end{equation}

\clearpage
\section{Additional Results}
\label{appdx:results}
\subsection{Monitoring metrics and loss functions while training with SVI-AvUC}
\label{appdx:monitorloss}
\begin{figure*}[ht]
	\small
	\centering
	\begin{subfigure}{\textwidth}
		\includegraphics[width=0.9\linewidth]{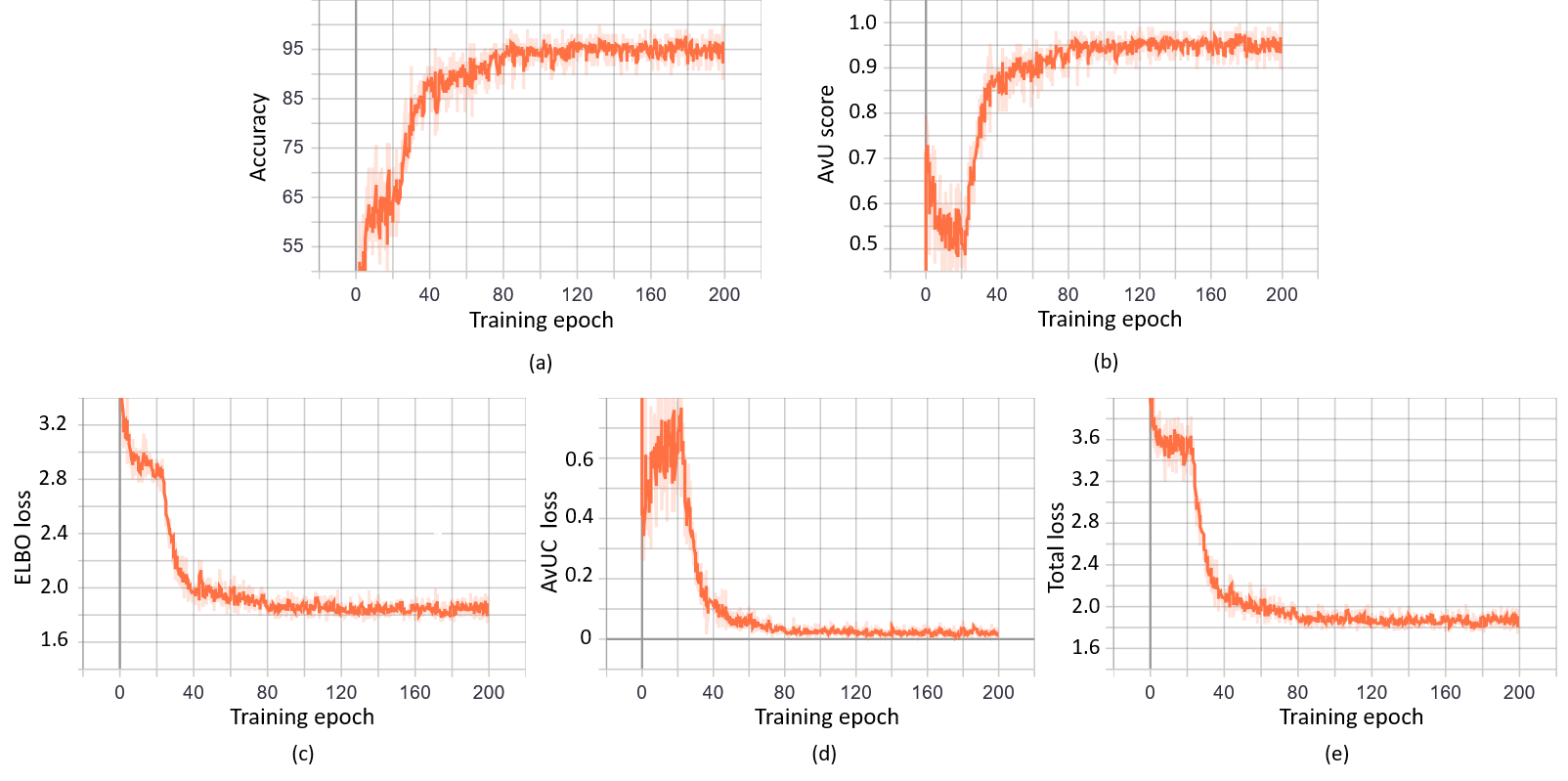}
		\label{fig:avu_imagenet}
	\end{subfigure}
	\caption{SVI-AvUC ResNet-20/CIFAR: Training. Monitoring accuracy, AvU metric, ELBO loss, AvUC loss and total loss at each training epoch.}
	\label{fig:supp_training1}
\end{figure*}
\begin{figure*}[h]
	\small
	\centering
	\begin{subfigure}{\textwidth}
		\centering
		\includegraphics[width=0.6\linewidth]{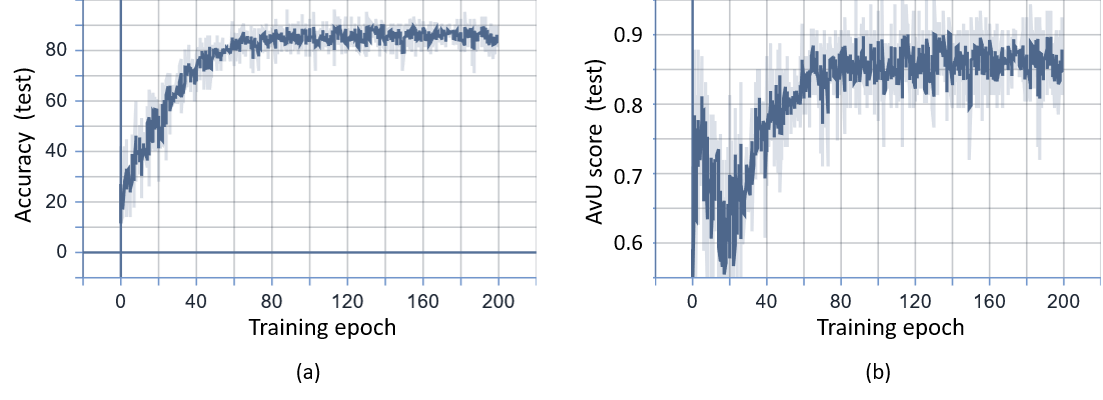}
		\label{fig:avu_imagenet}
	\end{subfigure}
	\caption{SVI-AvUC ResNet-20/CIFAR: Validation accuracy and AvU score. Monitoring accuracy and AvU metric on test data at after each training epoch.}
	\label{fig:supp_training2}
\end{figure*}

Figure~\ref{fig:supp_training1} shows ELBO loss, AvUC (\textit{acuuracy vs uncertainty calibration}) loss and total loss (combination of ELBO and AvUC losses) along with accuracy and AvU metrics at each training epoch. ELBO loss consist of two components including negative expected log-likelihood and Kullback-Leibler divergence as given by Equation~\ref{eqnsupp:ELBO_loss}. We can observe that the ELBO loss decreases as accuracy is increasing indicating the inverse correlation between them. We can also see that ELBO loss is decreasing even if the AvU score is not increasing. AvU provides relationship between accuracy and uncertainty that hints model calibration as described in Section~\ref{sec:avu_loss}. Figure~\ref{fig:supp_training1}(b) and (d) show that the proposed differentiable AvUC loss and actual AvU metric is inversely correlated, guiding the gradient optimization of total loss with respect to improving both accuracy and uncertainty calibration. Figure~\ref{fig:supp_training2} shows accuracy and AvU score on test data obtained from 1 Monte Carlo sample at the end of each training epoch (for monitoring). The model accuracy and AvU score during evaluation phase will be higher as we use larger number of Monte Carlo samples to marginalize over the weight posterior. 

\subsection{Additional results for model calibration evaluation}
\label{appdx:addlmodelcalib}
\begin{figure*}[ht]
\small
\centering
\begin{subfigure}{\textwidth}
\includegraphics[width=1\linewidth]{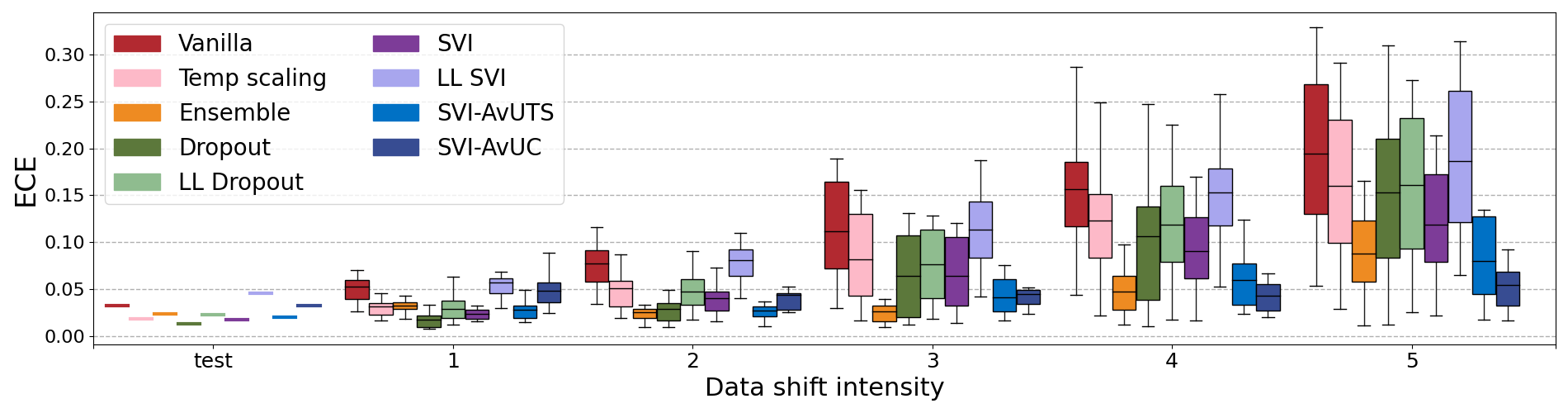}
\label{fig:ece_imagenet} 
\end{subfigure}
\begin{subfigure}{\textwidth}
\includegraphics[width=1\linewidth]{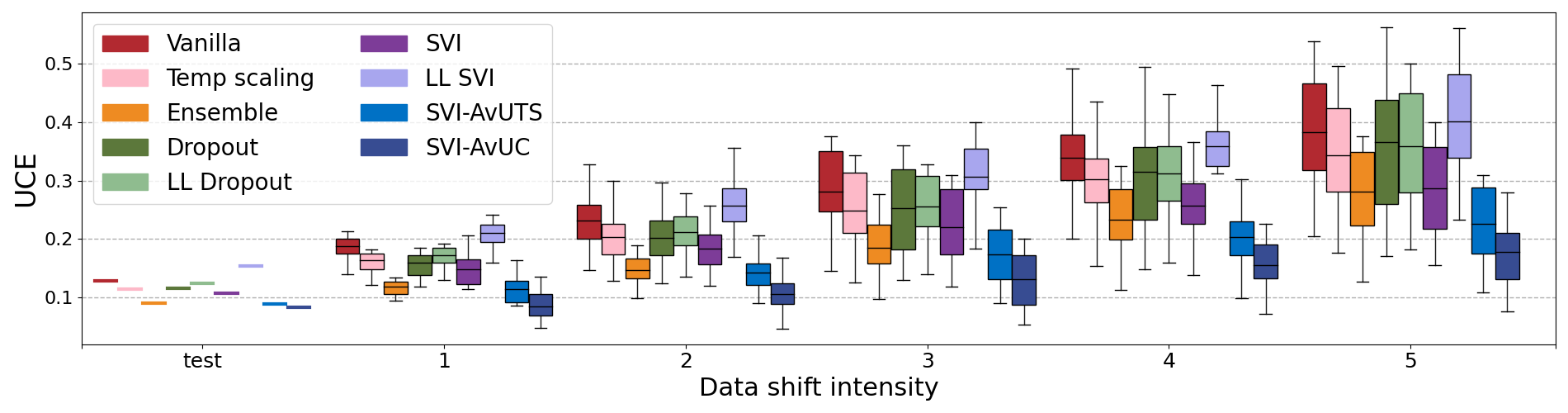}
\label{fig:uce_imagenet} 
\end{subfigure}
\begin{subfigure}{\textwidth}
\includegraphics[width=1\linewidth]{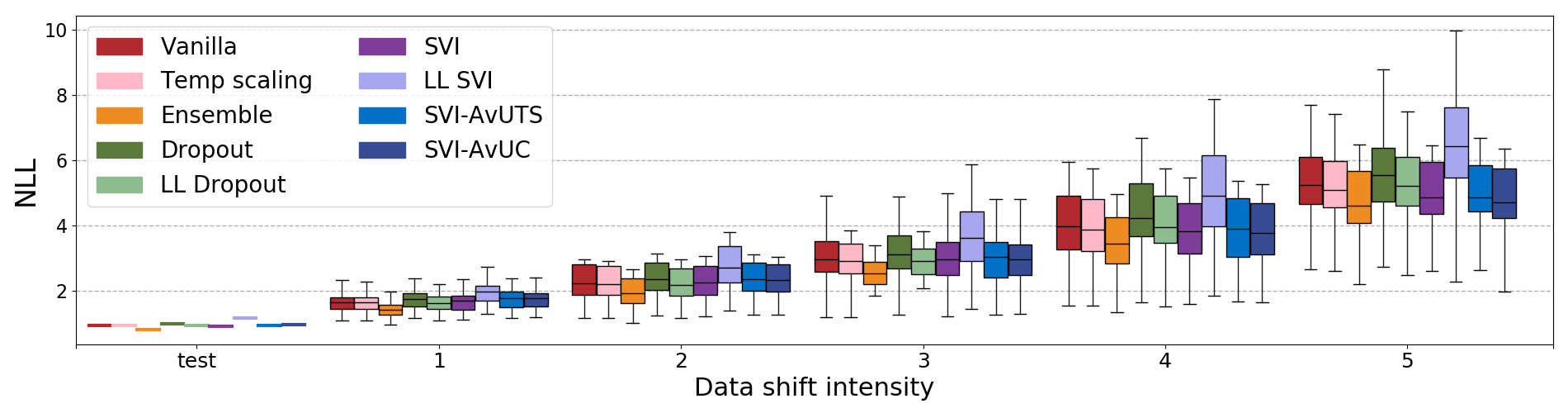}
\label{fig:nll_imagenet} 
\end{subfigure}
\begin{subfigure}{\textwidth}
\includegraphics[width=1\linewidth]{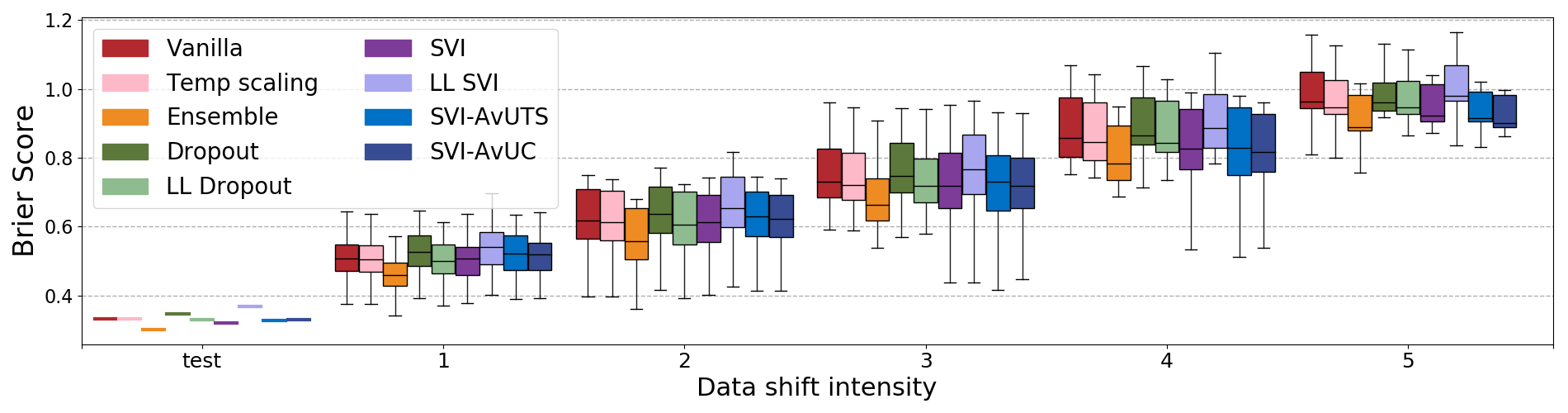}
\label{fig:brier_imagenet} 
\end{subfigure}
\begin{subfigure}{\textwidth}
\includegraphics[width=1\linewidth]{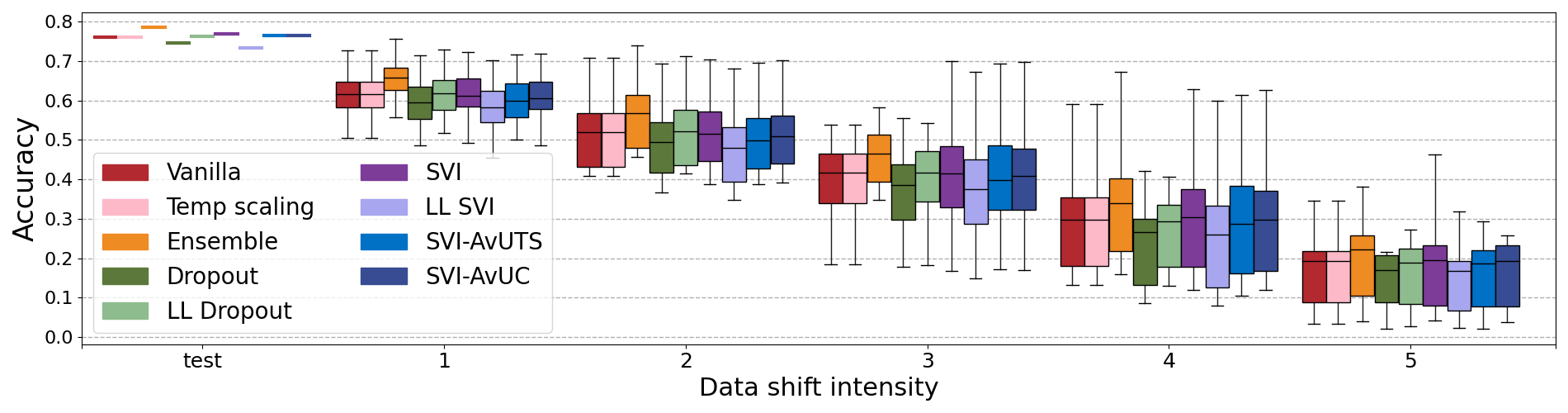}
\label{fig:acc_imagenet}
\end{subfigure}
\caption{\small  ResNet-50/ImageNet: Model calibration comparison using ECE$\downarrow$, UCE$\downarrow$, NLL$\downarrow$ and Brier score$\downarrow$ on ImageNet under in-distribution (test) and dataset shift at different levels of shift intensities (1-5). A well-calibrated model should consistently provide lower ECE, UCE, NLL and Brier score even at increased levels of datashift, as accuracy may degrade with increased datashift. At each shift intensity level, the boxplot summarizes the results across 16 different datashift types showing the min, max, mean and quartiles.}
\label{fig:supp_boxplot_imagenet}
\end{figure*}

\begin{figure*}
	\small
	\centering
	\begin{subfigure}{\textwidth}
		\includegraphics[width=1\linewidth]{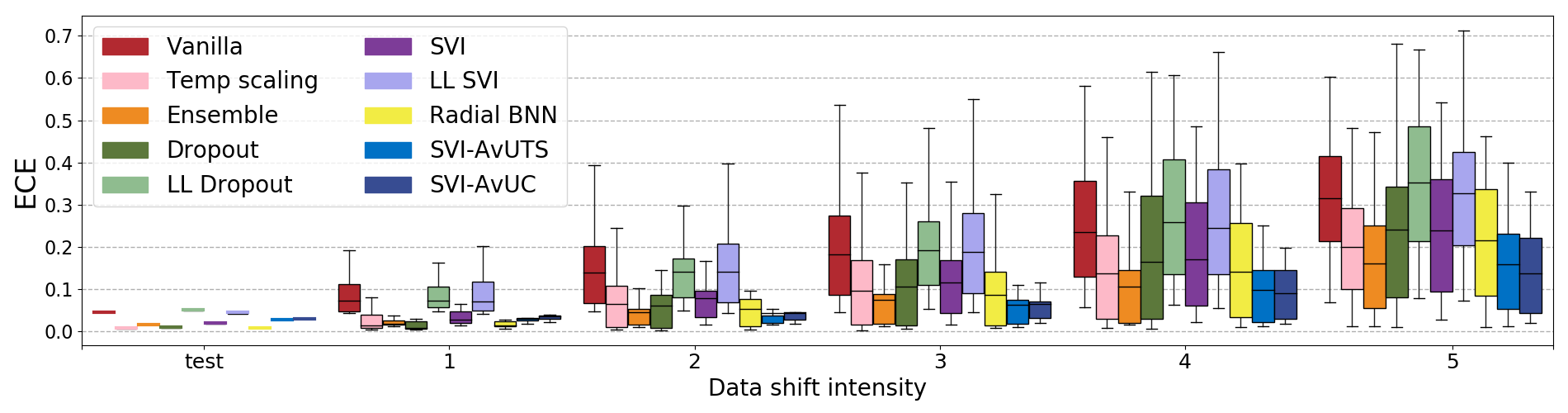}
		\label{fig:ece_imagenet} 
	\end{subfigure}
	\begin{subfigure}{\textwidth}
		\includegraphics[width=1\linewidth]{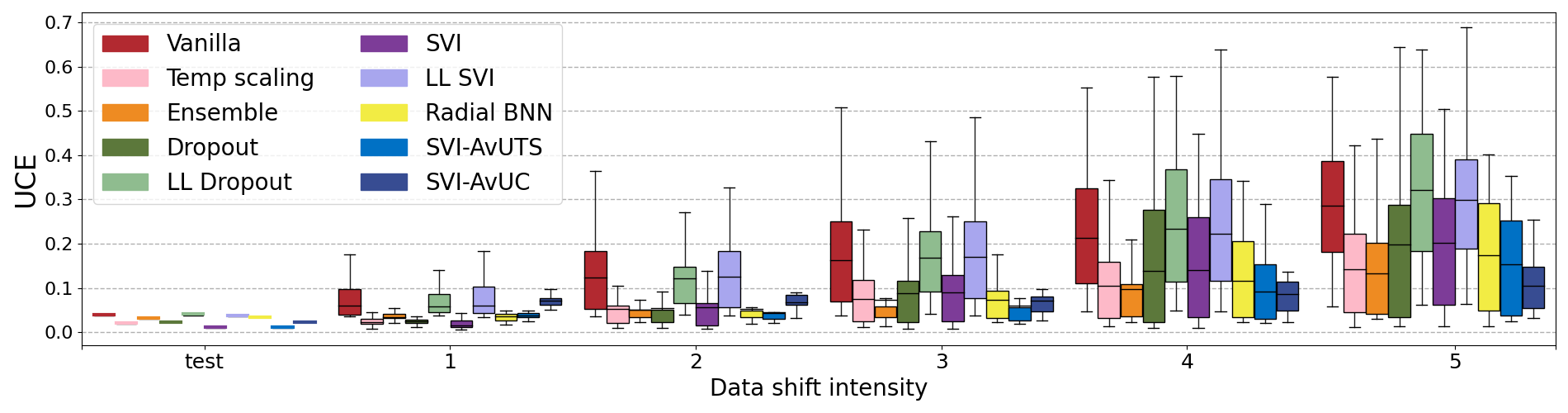}
		\label{fig:uce_imagenet} 
	\end{subfigure}
	\begin{subfigure}{\textwidth}
		\includegraphics[width=1\linewidth]{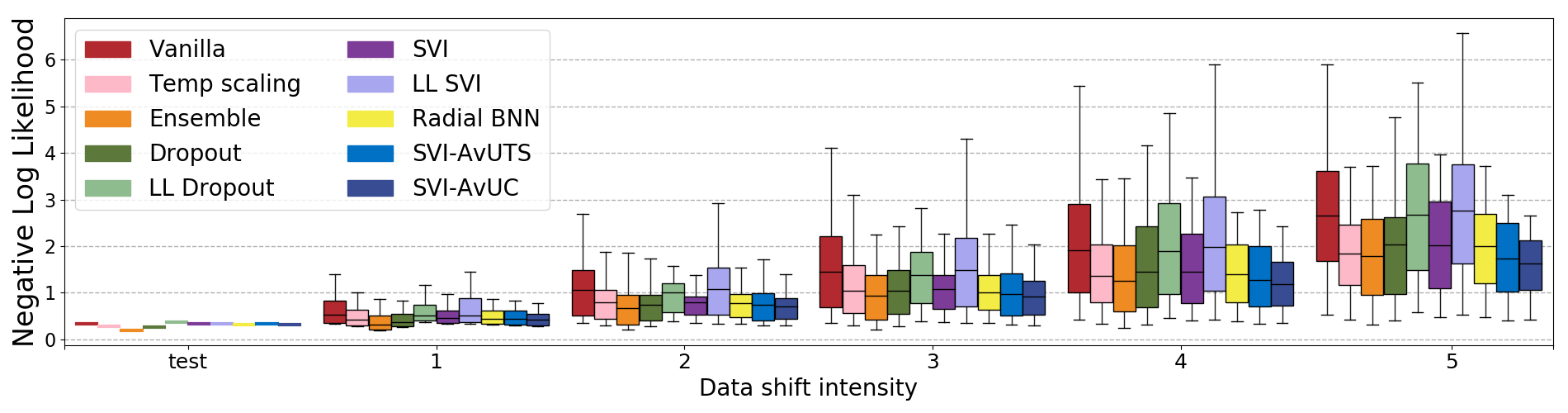}
		\label{fig:nll_imagenet} 
	\end{subfigure}
	\begin{subfigure}{\textwidth}
		\includegraphics[width=1\linewidth]{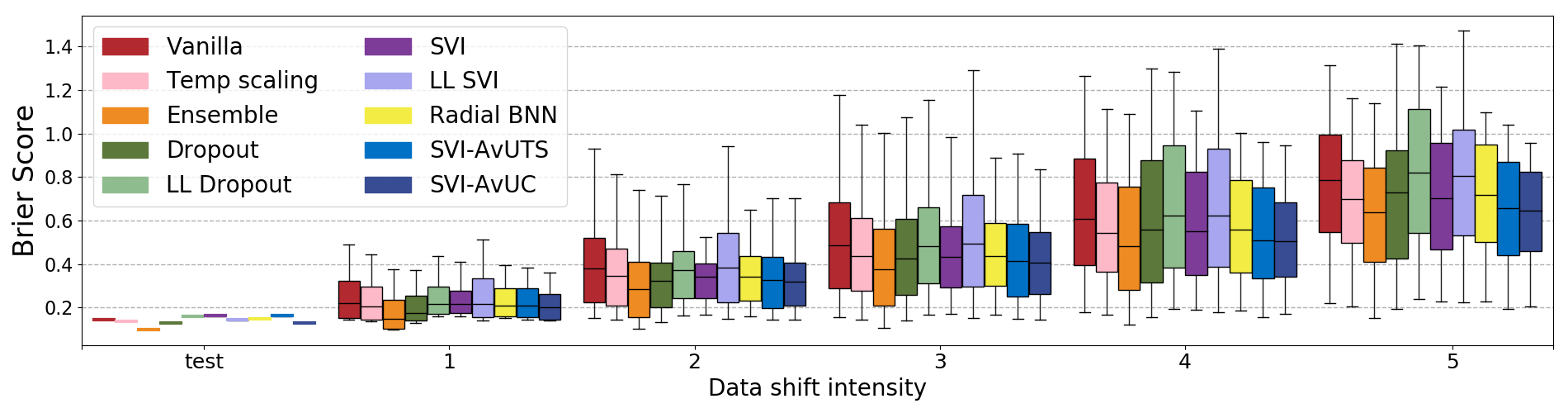}
		\label{fig:brier_imagenet} 
	\end{subfigure}
	\begin{subfigure}{\textwidth}
		\includegraphics[width=1\linewidth]{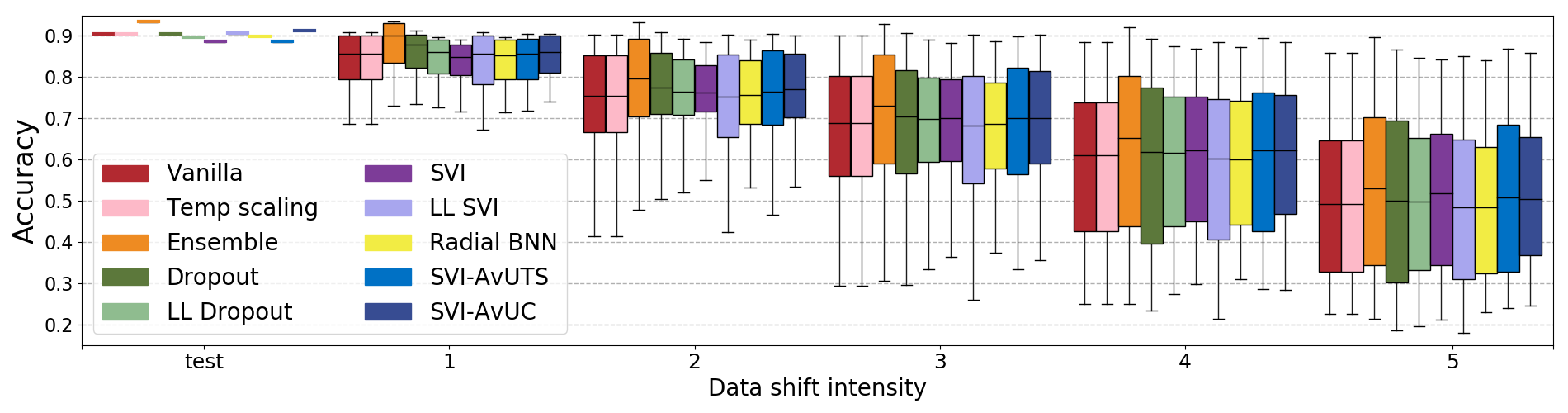}
		\label{fig:acc_imagenet}
	\end{subfigure}
	\caption{\small  ResNet-20/CIFAR10: Model calibration comparison using ECE$\downarrow$, UCE$\downarrow$, NLL$\downarrow$ and Brier score$\downarrow$ on CIFAR10 under in-distribution (test) and dataset shift at different levels of shift intensities (1-5). A well-calibrated model should consistently provide lower ECE, UCE, NLL and Brier score even at increased levels of datashift, as accuracy may degrade with increased datashift. At each shift intensity level, the boxplot summarizes the results across 16 different datashift types showing the min, max, mean and quartiles.}
	\label{fig:supp_boxplot_cifar}
\end{figure*}

In addition to model calibration evaluation with \textit{expected calibration error} (ECE) $\downarrow$ and \textit{expected uncertainty calibration error} (UCE) $\downarrow$ metrics in Figure~\ref{fig:boxplot} of Section~\ref{sec:results}, we also compare negative log-likelood (NLL) $\downarrow$ and Brier score metrics $\downarrow$ obtained from different methods on ImageNet (ResNet-50) and CIFAR10 (ResNet-20) across 80 combinations of datashift including 16 different types of shift at 5 different levels of shift intensities. The results are shown in Figure~\ref{fig:supp_boxplot_imagenet} for ImageNet and in Figure~\ref{fig:supp_boxplot_cifar} for CIFAR10.
 
\clearpage
The Spearman rank-order correlation coefficient ($\rho$)~\cite{kokoska2000crc} is a nonparametric measure of rank correlation, which asseses the monotonic relationships between two variables. Spearman's $\rho\in[-1, 1]$, with -1 or +1 implies exact monotonic relationship (negative and positive correlations respectively) and 0 implies no correlation between two variables. We assess the effect of increasing data shift intensities on the model calibration errors with Spearman rank-order correlation coefficient as shown in Table~\ref{tab:spearman}. A perfectly calibrated and robust model will have Spearman's $\rho$ equal to 0 indicating the model calibration errors are not correlated to data shift. The results in Table~\ref{tab:spearman} shows that ECE and UCE increases with data shift for all the methods, with comparatively lower $\rho$ values for SVI-AvUC indicating the proposed method is robust to data shift.   

\vspace{-2mm}
\begin{table}[h]
	\centering
	\caption{\small Spearman rank-order correlation coefficient assessing the monotonic relationship between model calibration errors (ECE and UCE) and the data shift intensity for the results in the Figures~\ref{fig:supp_boxplot_imagenet} and \ref{fig:supp_boxplot_cifar}. Spearman's $\rho$ indicates SVI-AvUC is robust as model calibration errors are less correlated to data shift compared to other methods. $\rho$ value near to 0 is better.}
	\begin{adjustbox}{width=1\linewidth}
		\begin{tabular}{@{}cccccccccccccc@{}}
			\toprule
			\multirow{2}{*}{Dataset/Model}                                                 & \multirow{2}{*}{} & \multirow{2}{*}{\begin{tabular}[c]{@{}c@{}}Spearman's $\rho$\\ rank-order correlation co-eff\\wrt dataset shift intensity\end{tabular}} & \multirow{2}{*}{} & \multicolumn{10}{c}{Method}                                                                                                                             \\ \cmidrule(l){5-14} 
			&                   &                                                                                                         &                   & Vanilla & \begin{tabular}[c]{@{}c@{}}Temp\\  scaling\end{tabular} & Ensemble & Dropout & LL Dropout & SVI & LL-SVI & SVI-TS & SVI-AvUTS & SVI-AvUC      \\ \midrule
			\multirow{2}{*}{\begin{tabular}[c]{@{}c@{}}ImageNet/\\ ResNet-50\end{tabular}} & \multirow{2}{*}{} & $\rho$$_{ECE}$                             &                   & 1.0     & 1.0                                                     & 0.6      & 1.0     & 1.0        & 1.0 & 1.0    & 1.0    & 0.94      & \textbf{0.31} \\
			&                   & $\rho$$_{UCE}$                              &                   & 1.0     & 1.0                                                     & 1.0      & 1.0     & 1.0        & 1.0 & 1.0    & 1.0    & 1.0       & \textbf{0.94} \\ \midrule
			\multirow{2}{*}{\begin{tabular}[c]{@{}c@{}}CIFAR10/\\ ResNet-20\end{tabular}}  & \multirow{2}{*}{} & $\rho$$_{ECE}$                              &                   & 1.0     & 1.0                                                     & 1.0      & 0.94    & 1.0        & 1.0 & 1.0    & 0.94   & 0.82      & \textbf{0.71} \\
			&                   & $\rho$$_{UCE}$                              &                   & 1.0     & 1.0                                                     & 1.0      & 1.0     & 1.0        & 1.0 & 1.0    & 0.77   & 0.82      & \textbf{0.71} \\ \bottomrule
		\end{tabular}
		\label{tab:spearman}
	\end{adjustbox}
\end{table}

\vspace{-4mm}
\begin{table}[h]
	\centering
	\caption{ImageNet: calibration under distributional shift. The lower quartile(25th percentile), median (50th percentile), mean and upper quartile (75th percentile) of ECE\;$\downarrow$, UCE\;$\downarrow$, NLL\;$\downarrow$ and Brier score\;$\downarrow$ computed across 16 different types of datashift at intensity 5 are presented below.}
	\begin{adjustbox}{width=1\textwidth}
	\begin{tabular}{@{}llccccccccc@{}}
		\toprule
		\multicolumn{2}{l}{\multirow{2}{*}{Metric}}                                                      & \multicolumn{9}{c}{Methods}                                                                       \\ \cmidrule(l){3-11} 
		\multicolumn{2}{l}{}                                                                             & Vanilla & Temp scaling & Ensemble & Dropout & LL Dropout & SVI    & LL SVI & SVI-AvUTS & SVI-AvUC \\ \midrule
		\multicolumn{1}{c}{\multirow{4}{*}{\textbf{ECE}\;$\downarrow$}}                               & lower quartile & 0.1244  & 0.0959       & 0.0503   & 0.0783  & 0.0925     & 0.0722 & 0.1212 & 0.0420    & 0.0319   \\
		\multicolumn{1}{c}{}                                                            & median         & 0.1737  & 0.1392       & 0.0900   & 0.1339  & 0.1450     & 0.1144 & 0.1684 & 0.0807    & 0.0447   \\
		\multicolumn{1}{c}{}                                                            & mean           & 0.1942  & 0.1600       & 0.0880   & 0.1530  & 0.1612     & 0.1188 & 0.1868 & 0.0800    & 0.0542   \\
		\multicolumn{1}{c}{}                                                            & upper quartile & 0.2744  & 0.2364       & 0.1264   & 0.2186  & 0.2364     & 0.1723 & 0.2676 & 0.1275    & 0.0696   \\ \midrule
		\multirow{4}{*}{\textbf{UCE}\;$\downarrow$}                                                   & lower quartile & 0.3068  & 0.2701       & 0.2179   & 0.2552  & 0.2727     & 0.2125 & 0.3356 & 0.1725    & 0.1310   \\  
		& median         & 0.3664  & 0.3251       & 0.2848   & 0.3506  & 0.3427     & 0.2872 & 0.3817 & 0.2323    & 0.1853   \\
		& mean           & 0.3826  & 0.3428       & 0.2813   & 0.3651  & 0.3593     & 0.2865 & 0.4007 & 0.2263    & 0.1774   \\
		& upper quartile & 0.4752  & 0.4335       & 0.3506   & 0.4511  & 0.4572     & 0.3587 & 0.4917 & 0.2901    & 0.2113   \\ \midrule
		\multirow{4}{*}{\textbf{NLL}\;$\downarrow$}                                                   & lower quartile & 4.635   & 4.530        & 4.035    & 4.699   & 4.563      & 4.322  & 5.417  & 4.278     & 4.164    \\ 
		& median         & 5.115   & 4.993        & 4.624    & 5.093   & 5.034      & 4.853  & 6.076  & 4.912     & 4.823    \\
		& mean           & 5.234   & 5.091        & 4.604    & 5.553   & 5.201      & 4.865  & 6.422  & 4.860     & 4.707    \\
		& upper quartile & 6.292   & 6.165        & 5.893    & 6.522   & 6.342      & 6.034 & 7.755  & 5.941     & 5.778    \\ \midrule
		\multirow{4}{*}{\textbf{\begin{tabular}[c]{@{}l@{}}Brier\\ score\;$\downarrow$\end{tabular}}} & lower quartile & 0.941   & 0.926        & 0.877    & 0.933   & 0.923      & 0.906  & 0.963  & 0.893     & 0.883    \\  
		& median         & 0.987   & 0.970        & 0.922    & 0.967   & 0.969      & 0.943  & 0.998  & 0.948     & 0.935    \\
		& mean           & 0.964   & 0.945        & 0.888    & 0.961   & 0.947      & 0.922  & 0.979  & 0.914     & 0.900    \\
		& upper quartile & 1.052   & 1.027        & 0.989    & 1.025   & 1.025      & 1.013  & 1.072  & 0.996     & 0.985    \\ \bottomrule
	\end{tabular}
	\label{tab:imagenet_metrics}
	\end{adjustbox}
\end{table}

\begin{table}[ht]
	\centering
	\caption{CIFAR10: calibration under distributional shift. The lower quartile(25th percentile), median (50th percentile), mean and upper quartile (75th percentile) of ECE\;$\downarrow$, UCE\;$\downarrow$, NLL\;$\downarrow$ and Brier score\;$\downarrow$ computed across 16 different types of datashift at intensity 5 are presented below.}
	\begin{adjustbox}{width=1\linewidth}
	\begin{tabular}{@{}llccccccclcc@{}}
		\toprule
		\multicolumn{2}{l}{\multirow{2}{*}{Metric}}                                                      & \multicolumn{10}{c}{Methods}                                                                                                                                                                                            \\ \cmidrule(l){3-12} 
		\multicolumn{2}{l}{}                                                                             & Vanilla & \begin{tabular}[c]{@{}c@{}}Temp \\ scaling\end{tabular} & Ensemble & Dropout & LL Dropout & SVI    & LL SVI & \multicolumn{1}{c}{\begin{tabular}[c]{@{}c@{}}Radial\\ BNN\end{tabular}} & SVI-AvUTS & SVI-AvUC \\ \midrule \multirow{4}{*}{\textbf{ECE\;$\downarrow$}}                                                   & lower quartile & 0.2121  & 0.0997                                                  & 0.0549   & 0.0794  & 0.2022     & 0.0925 & 0.2027 & 0.0797                                                                   & 0.0466    & 0.0398   \\
		& median         & 0.3022  & 0.1834                                                  & 0.1045   & 0.1889  & 0.3643     & 0.2146 & 0.3077 & 0.1950                                                                   & 0.1516    & 0.1107   \\
		& mean           & 0.3151  & 0.1993                                                  & 0.1611   & 0.2405  & 0.3518     & 0.2389 & 0.3267 & 0.2150                                                                   & 0.1585    & 0.1374   \\
		& upper quartile & 0.4148  & 0.2915                                                  & 0.2551   & 0.3518  & 0.4854     & 0.3636 & 0.4246 & 0.3410                                                                   & 0.2345    & 0.2303   \\ \midrule
		\multirow{4}{*}{\textbf{UCE\;$\downarrow$}}                                                   & lower quartile & 0.1813  & 0.0419                                                  & 0.0417   & 0.0328  & 0.1728     & 0.0594 & 0.1875 & 0.0473                                                                   & 0.0575    & 0.0495   \\  
		& median         & 0.2773  & 0.1147                                                  & 0.0653   & 0.1382  & 0.3336     & 0.1723 & 0.2747 & 0.1449                                                                   & 0.11486   & 0.0740   \\
		& mean           & 0.2853  & 0.1429                                                  & 0.1333   & 0.1974  & 0.3204     & 0.2008 & 0.2983 & 0.1741                                                                   & 0.1272    & 0.1038   \\
		& upper quartile & 0.3871  & 0.2232                                                  & 0.2103   & 0.2903  & 0.4486     & 0.3034 & 0.3902 & 0.2941                                                                   & 0.1827    & 0.1512   \\ \midrule 
		\multirow{4}{*}{\textbf{NLL\;$\downarrow$}}                                                   & lower quartile & 1.634   & 1.166                                                   & 0.955    & 0.971   & 1.419      & 1.052  & 1.629  & 1.179                                                                    & 0.984     & 1.035    \\ 
		& median         & 2.666   & 1.957                                                   & 1.753    & 1.952   & 2.767      & 2.001  & 2.752  & 2.038                                                                    & 1.747     & 1.742    \\
		& mean           & 2.653   & 1.846                                                   & 1.779    & 2.036   & 2.682      & 2.017  & 2.764  & 1.995                                                                    & 1.728     & 1.633    \\
		& upper quartile & 3.617   & 2.467                                                   & 2.587    & 2.652   & 3.780      & 2.952  & 3.762  & 2.706                                                                    & 2.507     & 2.158    \\ \midrule 
		\multirow{4}{*}{\textbf{\begin{tabular}[c]{@{}l@{}}Brier\\ score\;$\downarrow$\end{tabular}}} & lower quartile & 0.546   & 0.496                                                   & 0.407    & 0.421   & 0.526      & 0.449  & 0.529  & 0.488                                                                    & 0.434     & 0.454    \\  
		& median         & 0.871   & 0.765                                                   & 0.651    & 0.727   & 0.848      & 0.702  & 0.850  & 0.738                                                                    & 0.675     & 0.692    \\
		& mean           & 0.785   & 0.697                                                   & 0.639    & 0.728   & 0.820      & 0.702  & 0.803  & 0.719                                                                    & 0.657     & 0.646    \\
		& upper quartile & 0.995   & 0.876                                                   & 0.844    & 0.943   & 1.111      & 0.957  & 1.017  & 0.960                                                                    & 0.876     & 0.837    \\ \bottomrule
	\end{tabular}
	\label{tab:my-table}
	\end{adjustbox}
\end{table}

\clearpage
\subsection{Additional results for confidence and uncertainty evaluation under distributional shift}
\label{appdx:addlconfunc}
\begin{figure}[h]
   \centering
    \begin{subfigure}[b]{\textwidth}
	    \centering
		\includegraphics[width=0.325\linewidth]{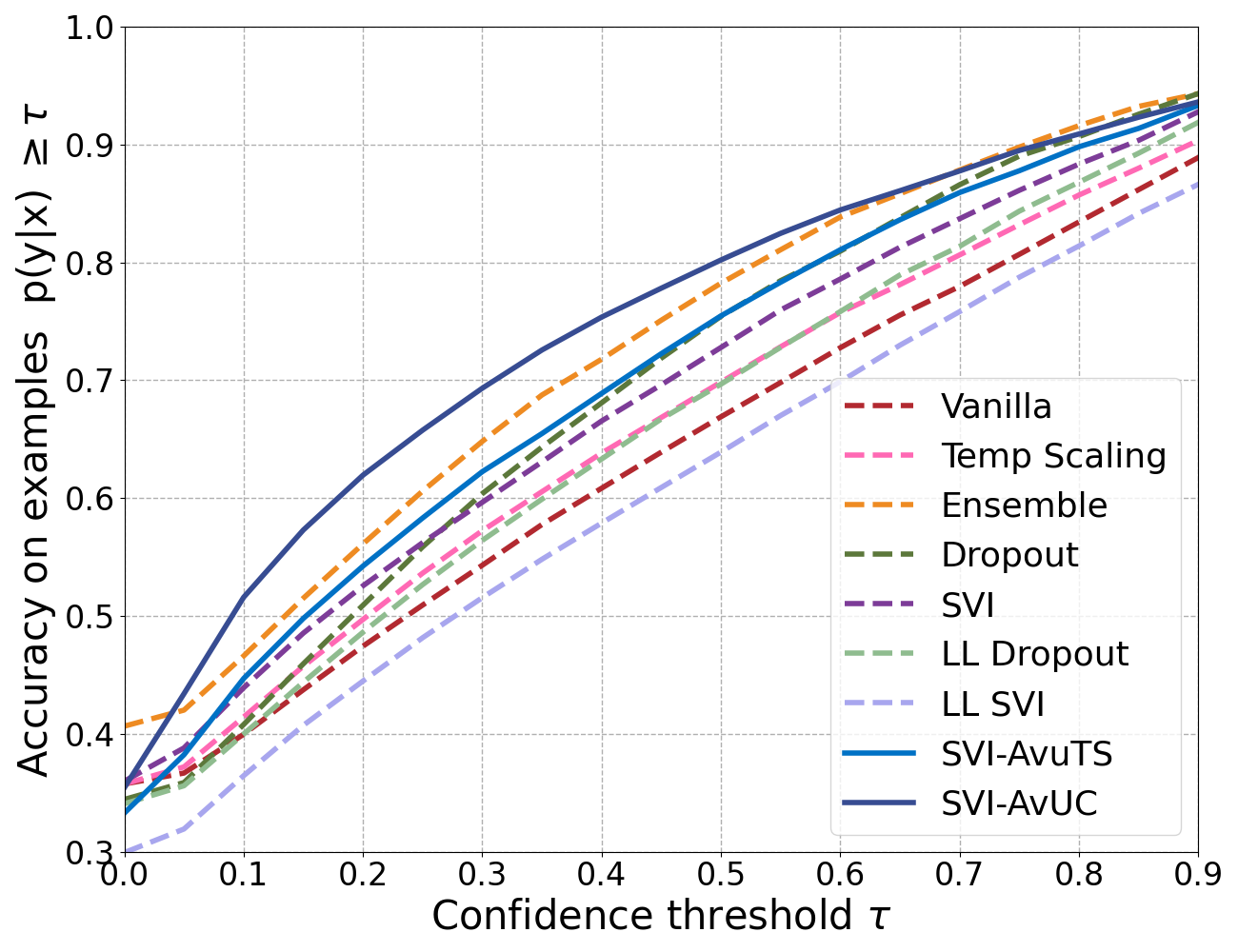}
		\hfill
		\includegraphics[width=0.325\linewidth]{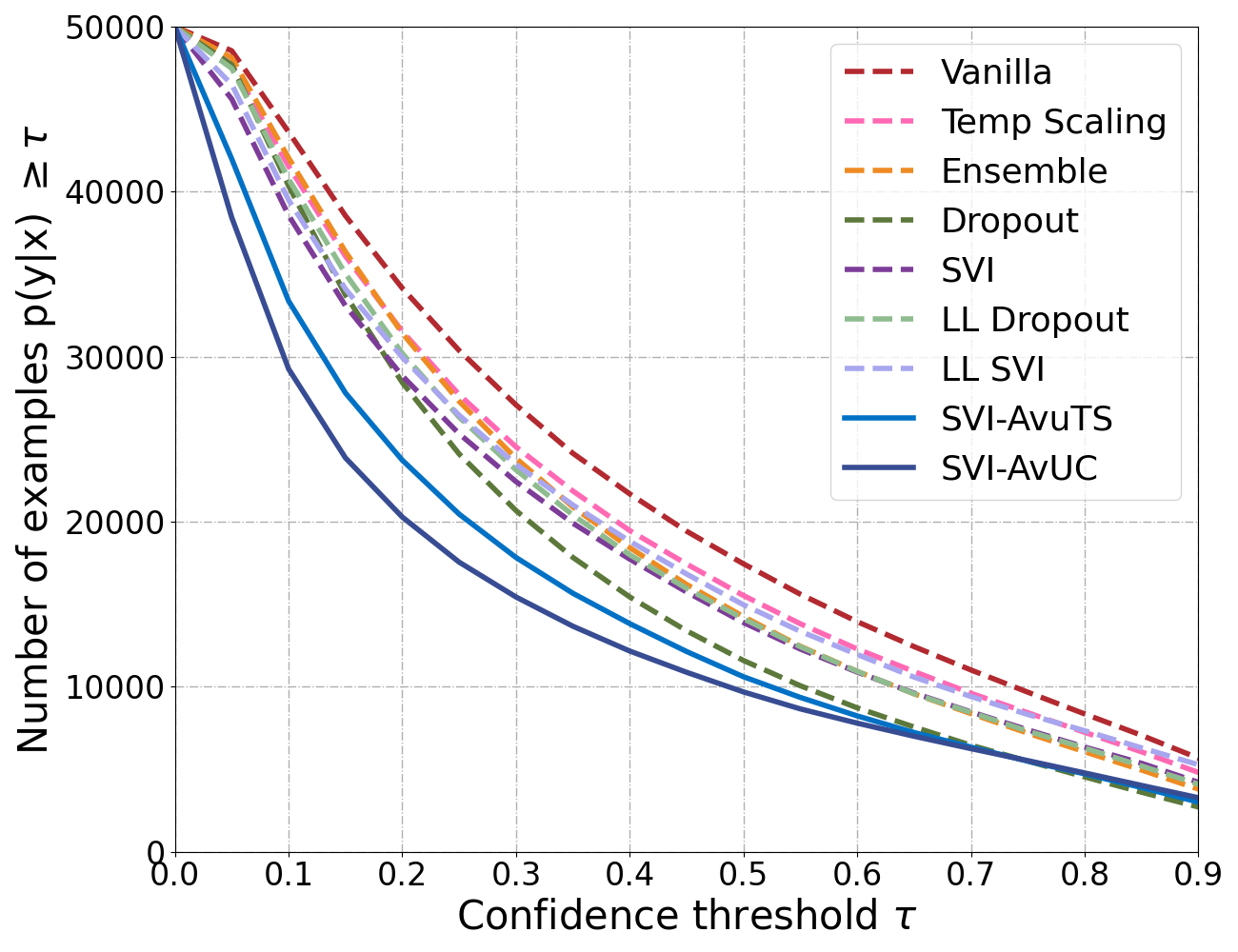}
		\hfill
		\includegraphics[width=0.325\linewidth]{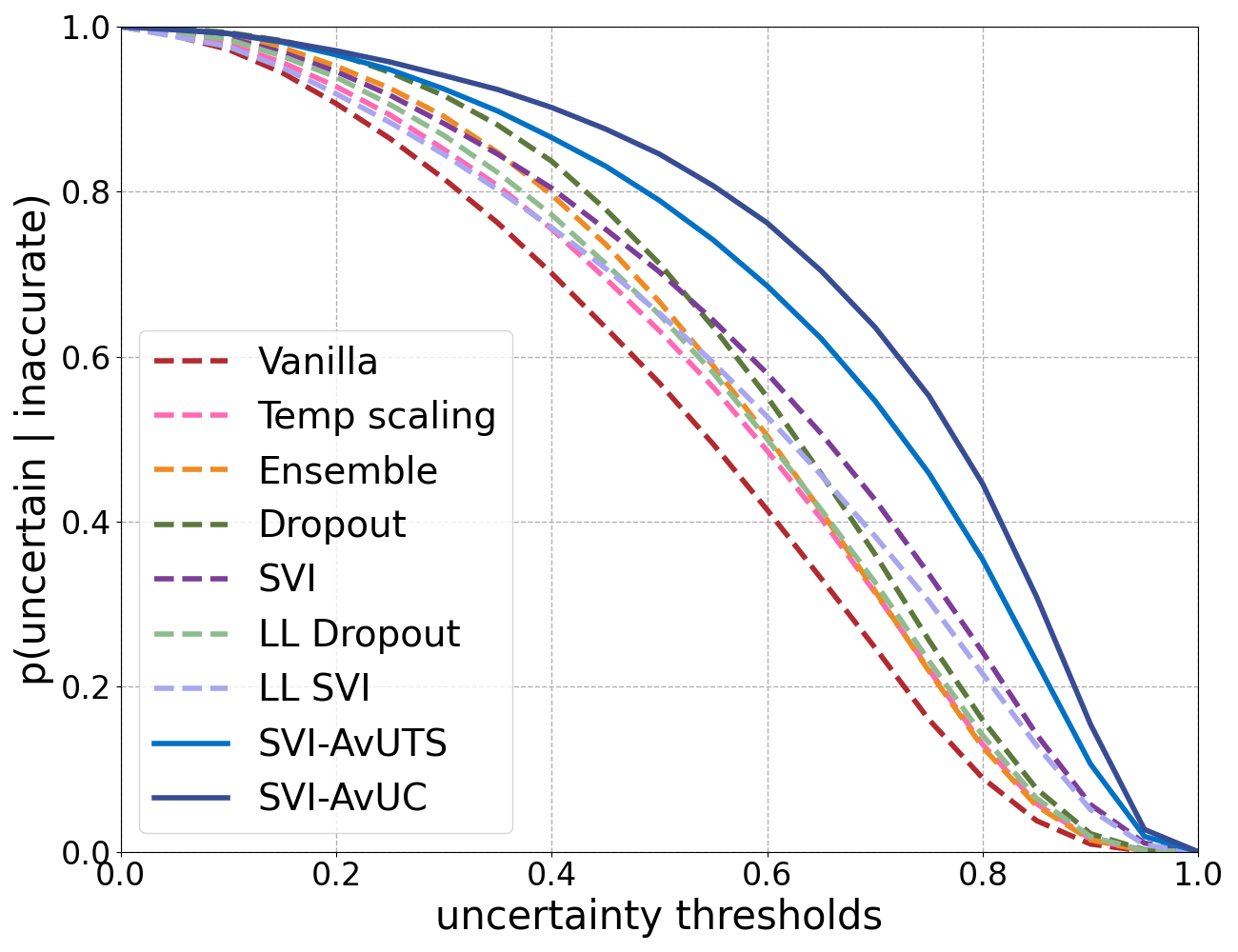}
		\hfill
		\caption{\small Defocus blur}
	\end{subfigure}
    \begin{subfigure}[b]{\textwidth}
    	\centering
    	\includegraphics[width=0.325\linewidth]{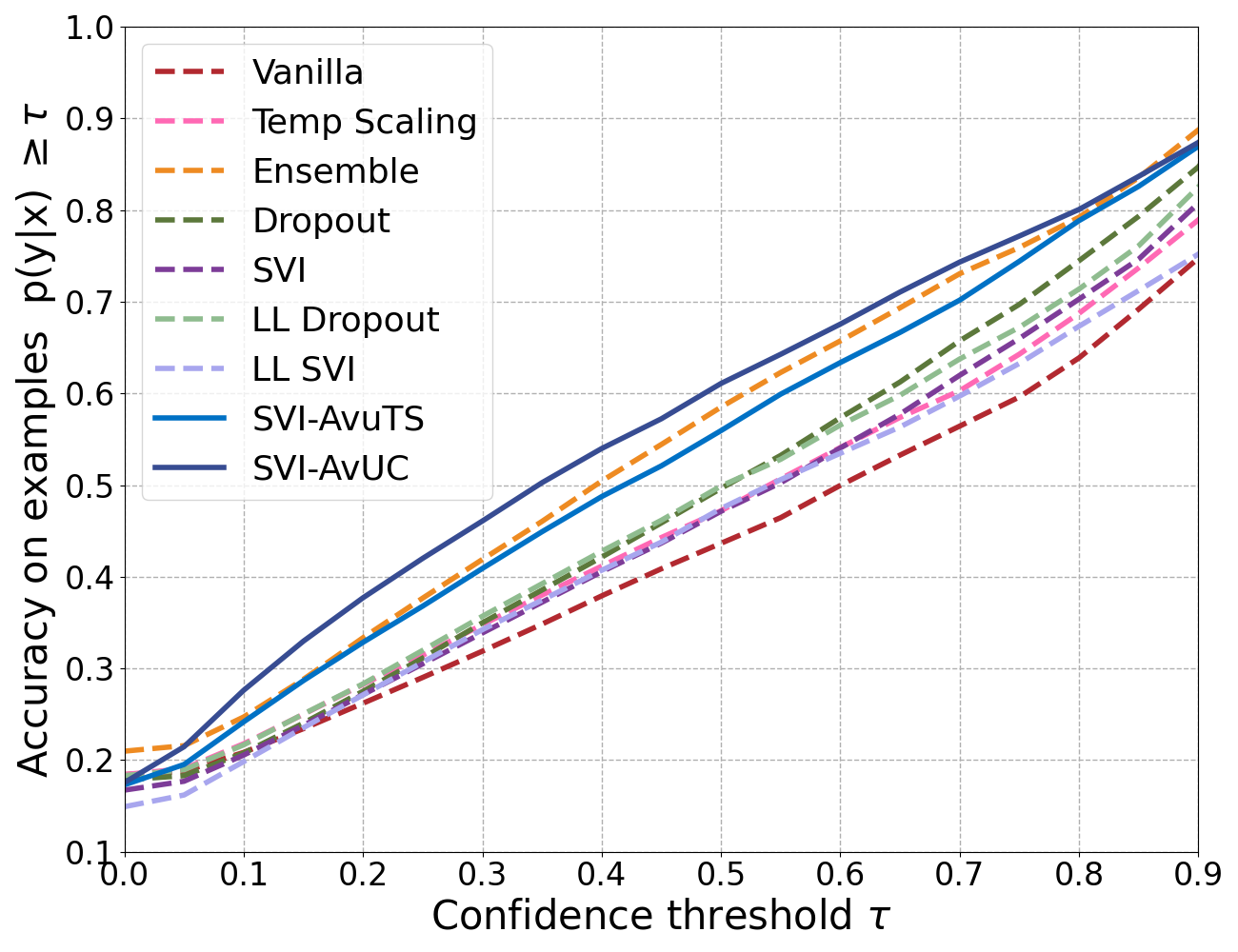}
    	\hfill
    	\includegraphics[width=0.325\linewidth]{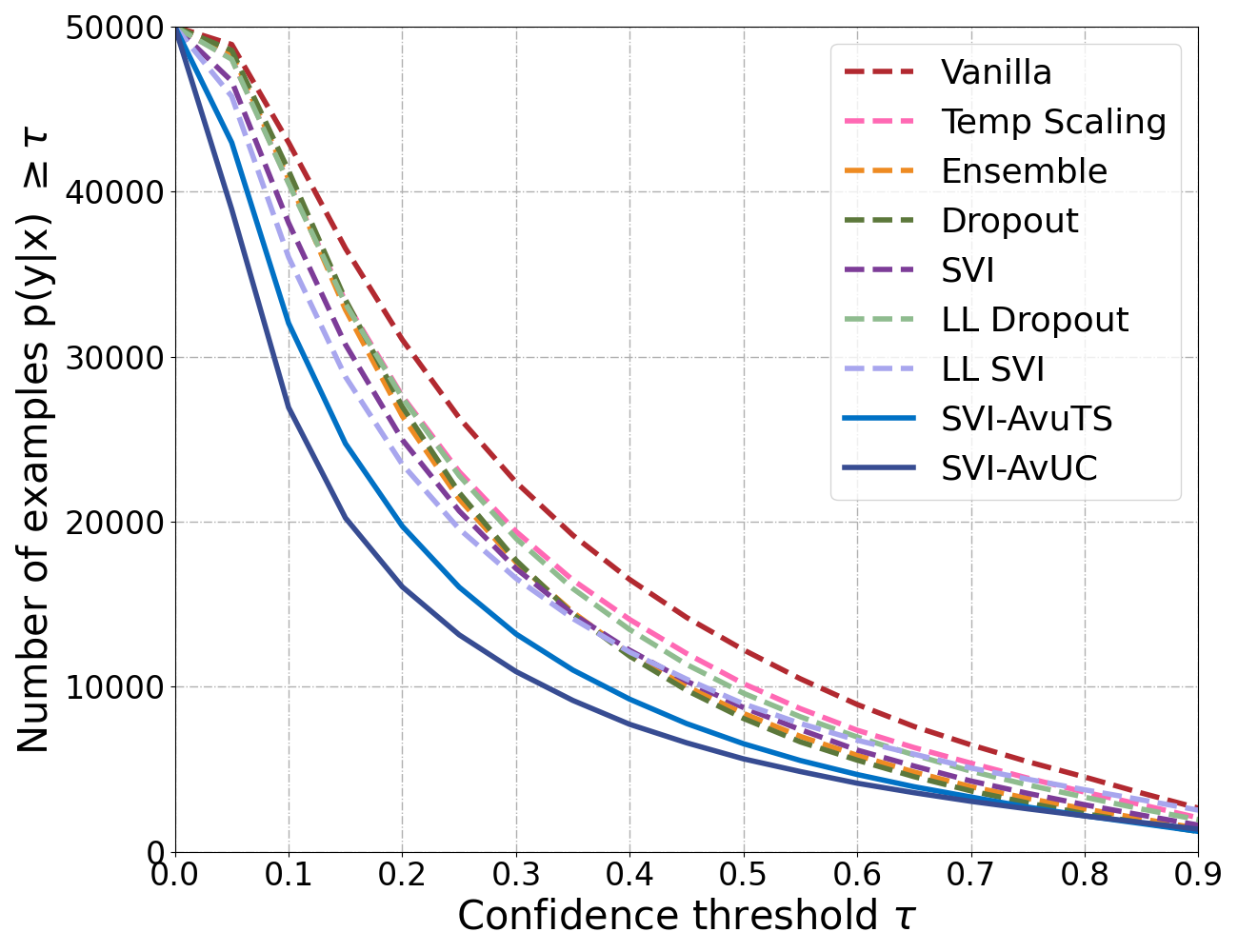}
    	\hfill
    	\includegraphics[width=0.325\linewidth]{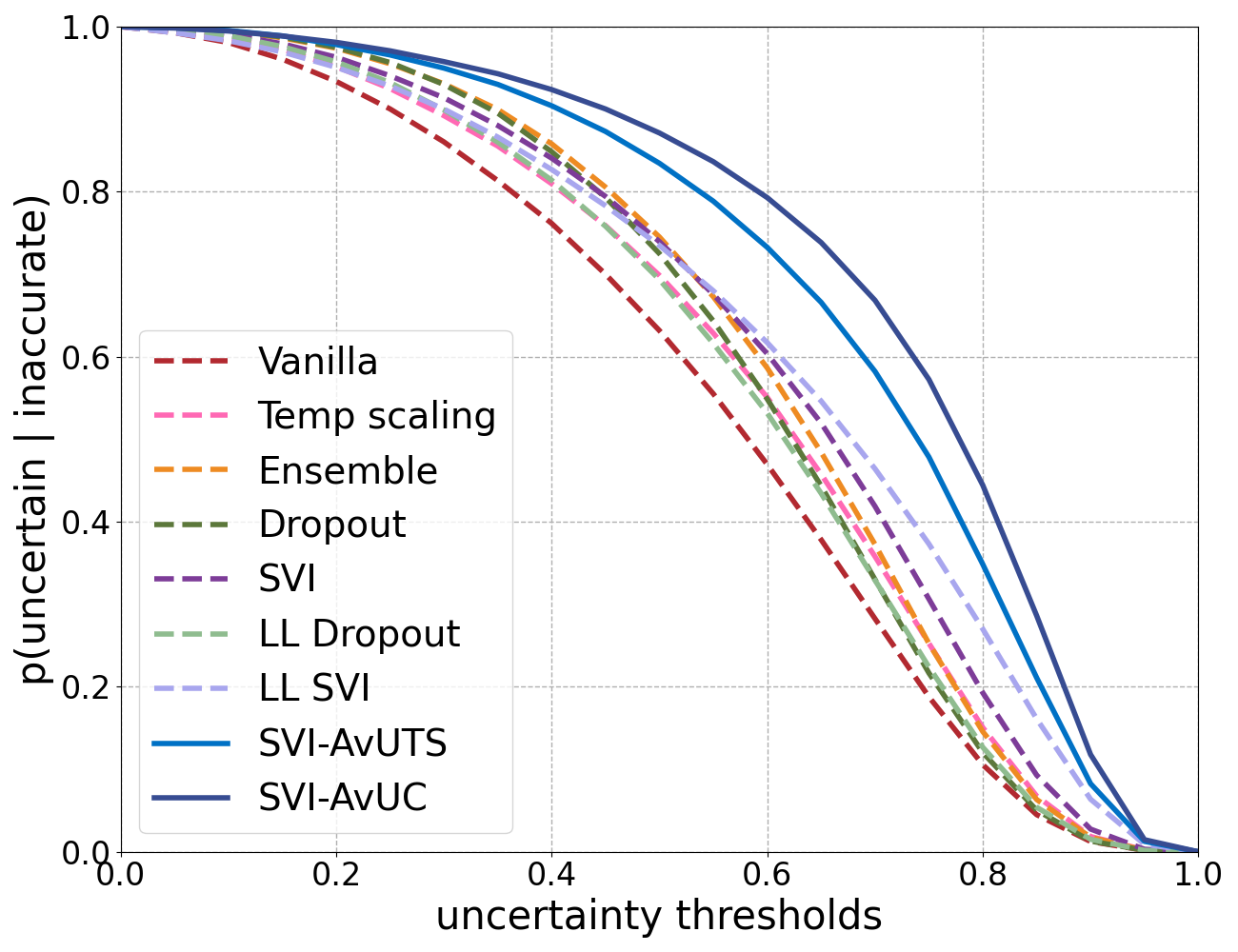}
    	\hfill
    	\caption{\small Glass blur}
    \end{subfigure}
	\caption{\small ImageNet: Model confidence and uncertainty evaluation under distributional shift (defocus blur and glass blur of intensity 3). Column 1: accuracy as a function of confidence. We expect a reliable model to be more accurate at higher confidence values; Column 2: number of examples above given confidence value. We expect a reliable model to have lesser number of examples with higher confidence as accuracy is significantly degraded under distributional shift; Column 3: probability of model being uncertain when making inaccurate predictions. We expect a reliable model to be more uncertain when it is inaccurate. Normalized uncertainty thresholds $\mathrm{t\in[0,1]}$ are shown in plots as the uncertainty range varies for different methods. All the plots show SVI-AvUC outperforms other methods.}
\end{figure}
\begin{figure}[ht!]
	\centering
	\vspace{-3mm}
	\begin{subfigure}[b]{\textwidth}
		\centering
		\includegraphics[width=0.325\linewidth]{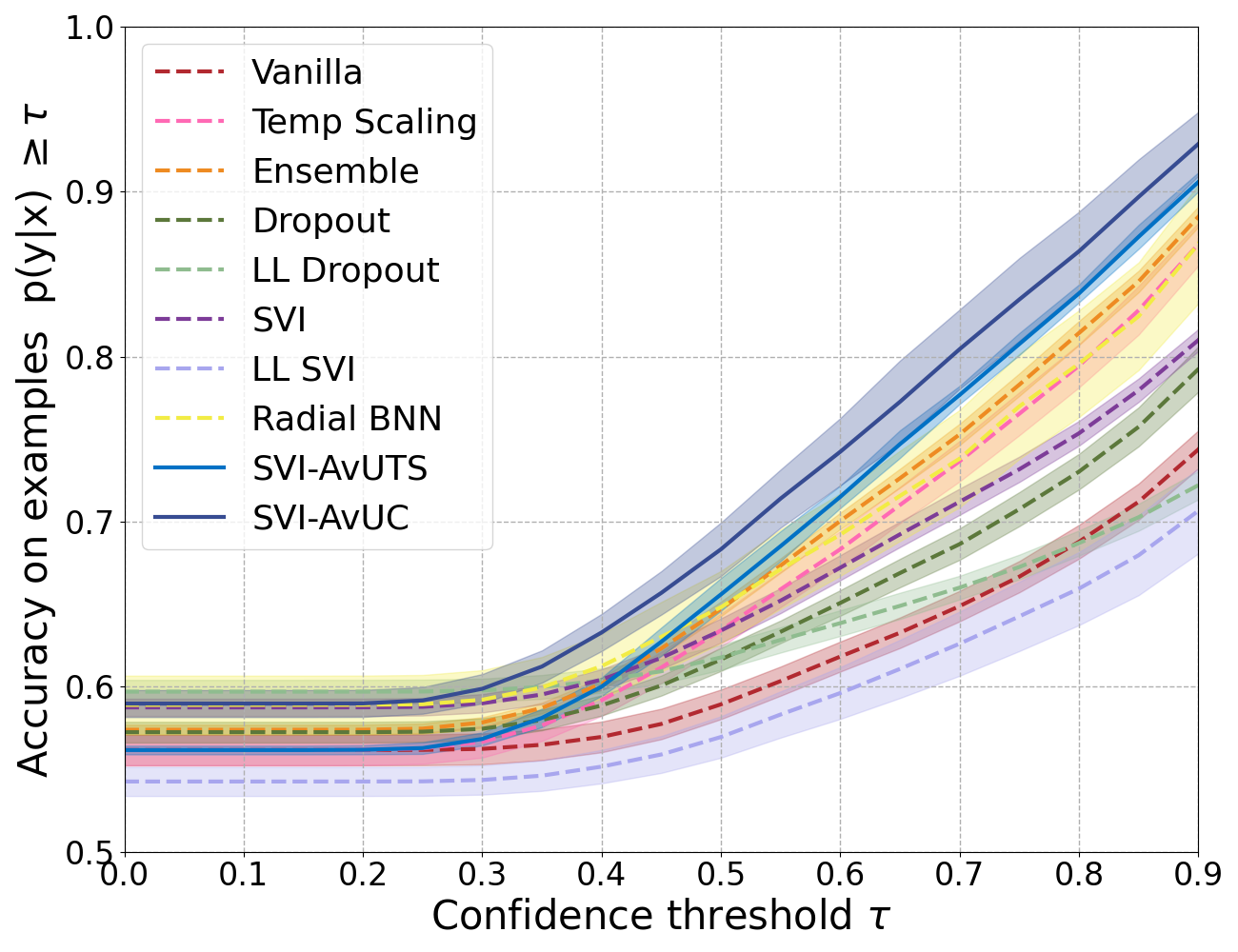}
		\hfill
		\includegraphics[width=0.325\linewidth]{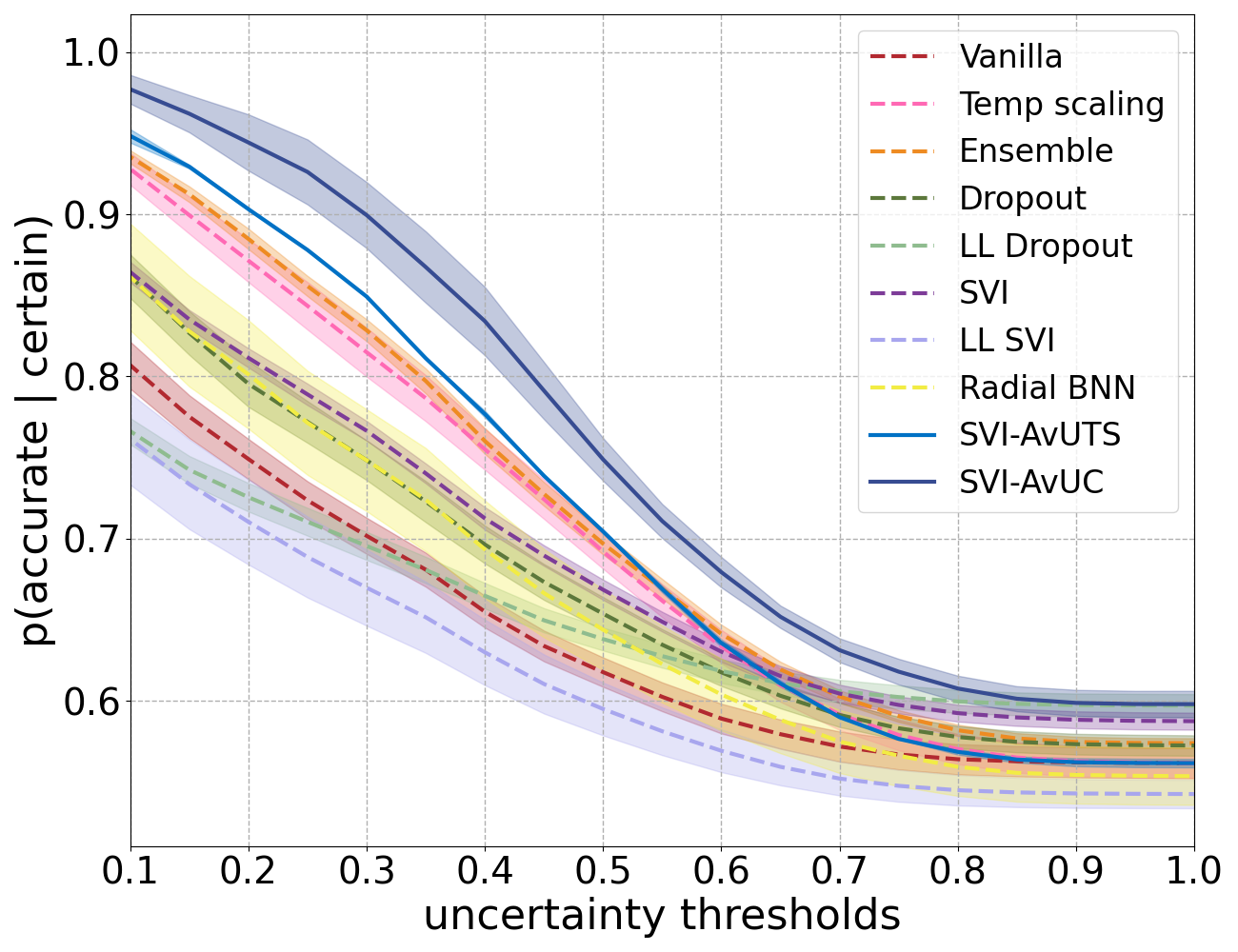}
		\hfill
		\includegraphics[width=0.325\linewidth]{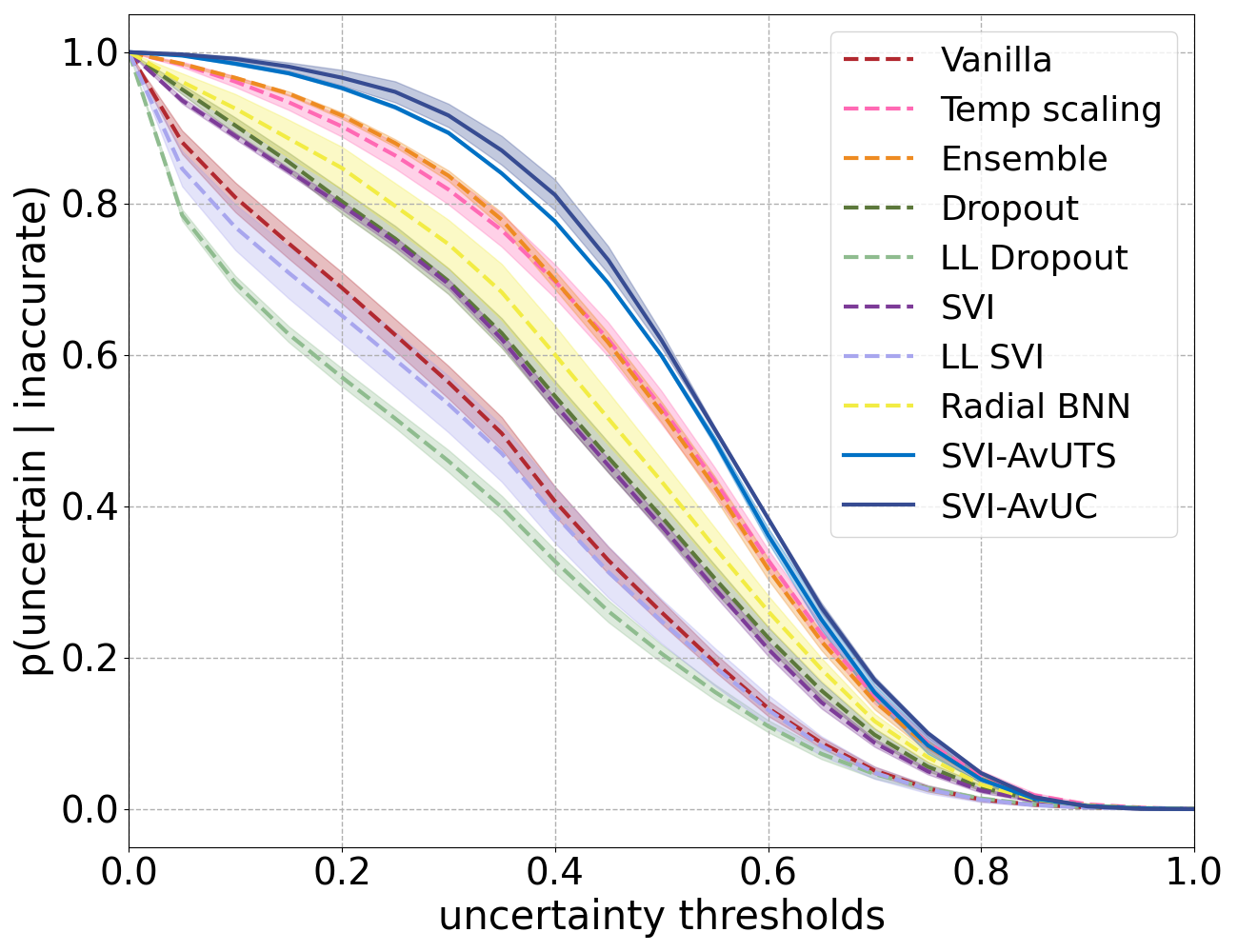}
		\hfill
		\caption{\small Speckle noise}
	\end{subfigure}
	\begin{subfigure}[b]{\textwidth}
		\centering
		\includegraphics[width=0.325\linewidth]{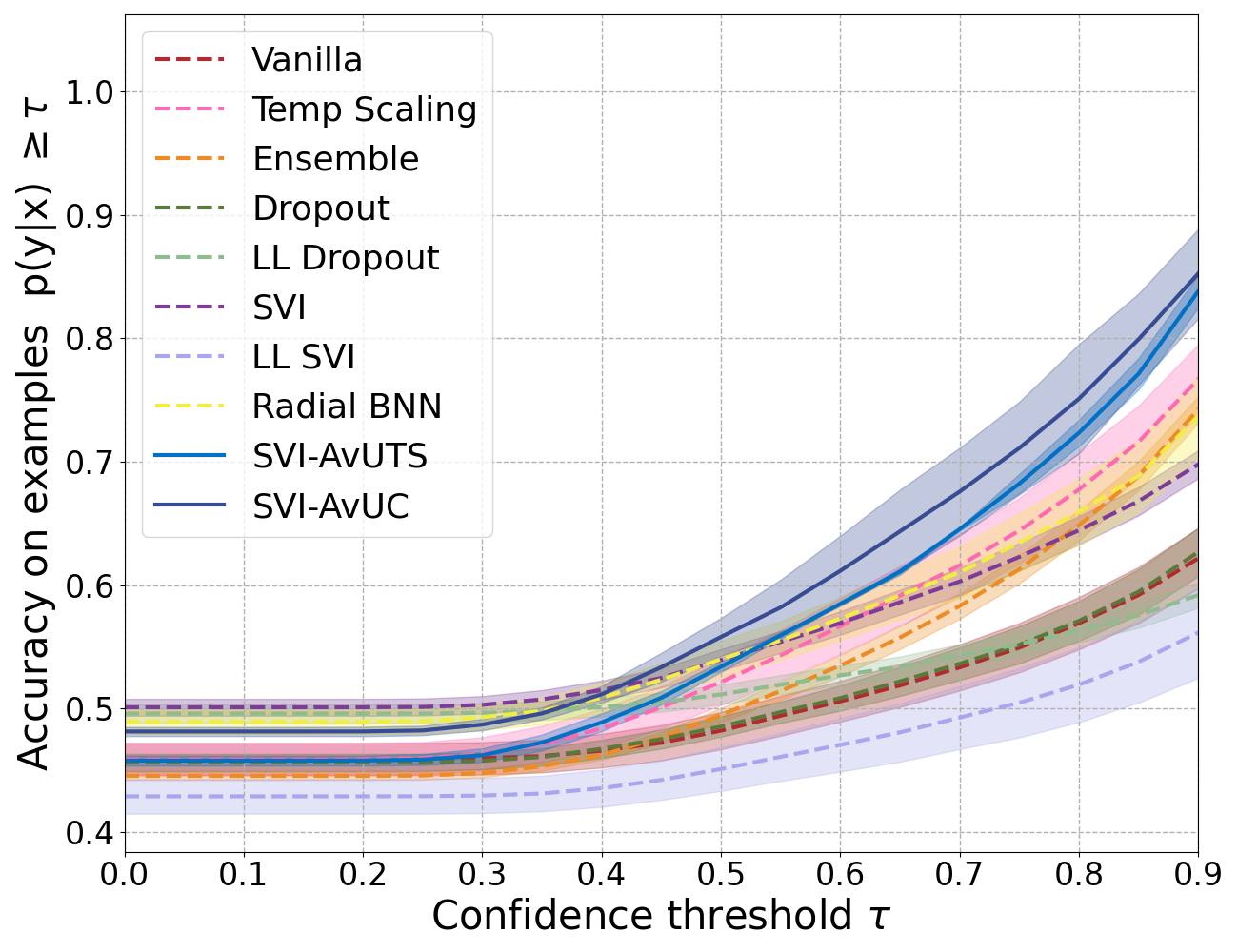}
		\hfill
		\includegraphics[width=0.325\linewidth]{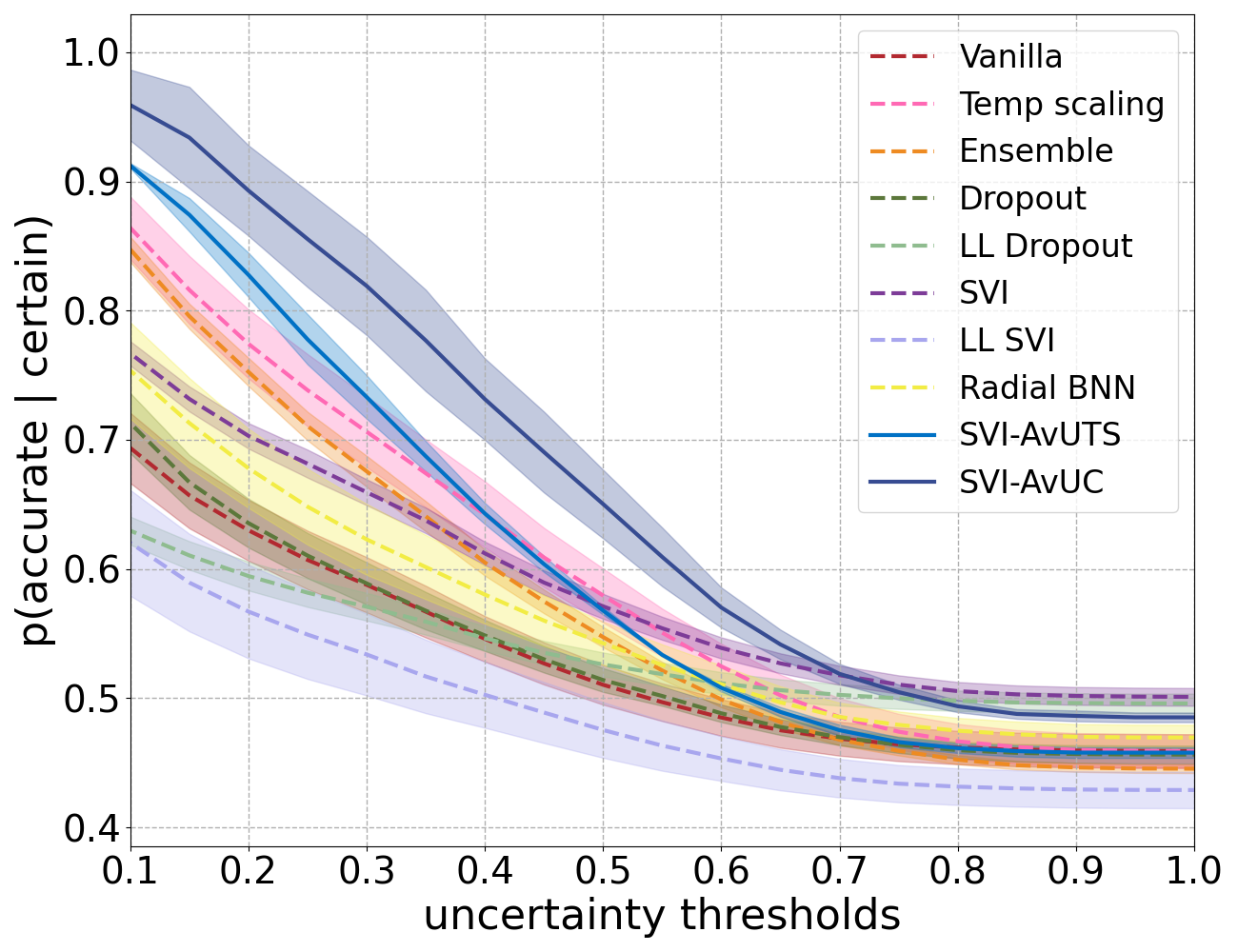}
		\hfill
		\includegraphics[width=0.325\linewidth]{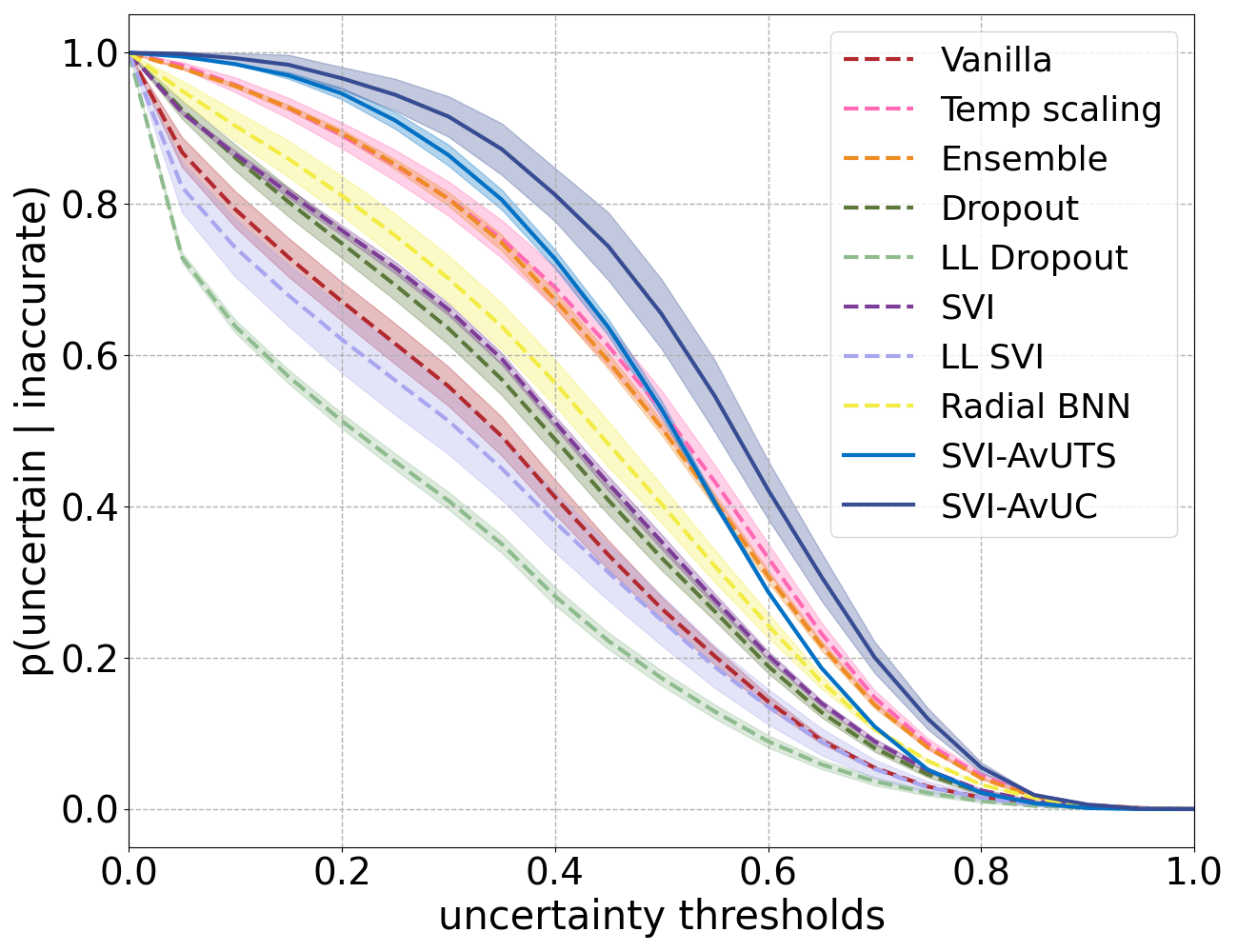}
		\hfill
		\caption{\small Shot noise}
	\end{subfigure}
	\caption{\small CIFAR: Model confidence and uncertainty evaluation under distributional shift (speckle noise and shot noise of intensity 3). Column 1: accuracy as a function of confidence; Column 2: probability of model being accurate on its predictions when it is certain; Column 3: probability of model being uncertain when making inaccurate predictions. Normalized uncertainty thresholds $\mathrm{t\in[0,1]}$ are shown in plots as the uncertainty range varies for different methods. All the plots show SVI-AvUC outperforms other methods.}
\end{figure}

\clearpage
\subsection{Comparing AUC of accuracy vs uncertainty (AvU) measures}
\begin{figure*}[hb!]
	\small
	\centering
	\begin{subfigure}{\textwidth}
		\includegraphics[width=1.0\linewidth]{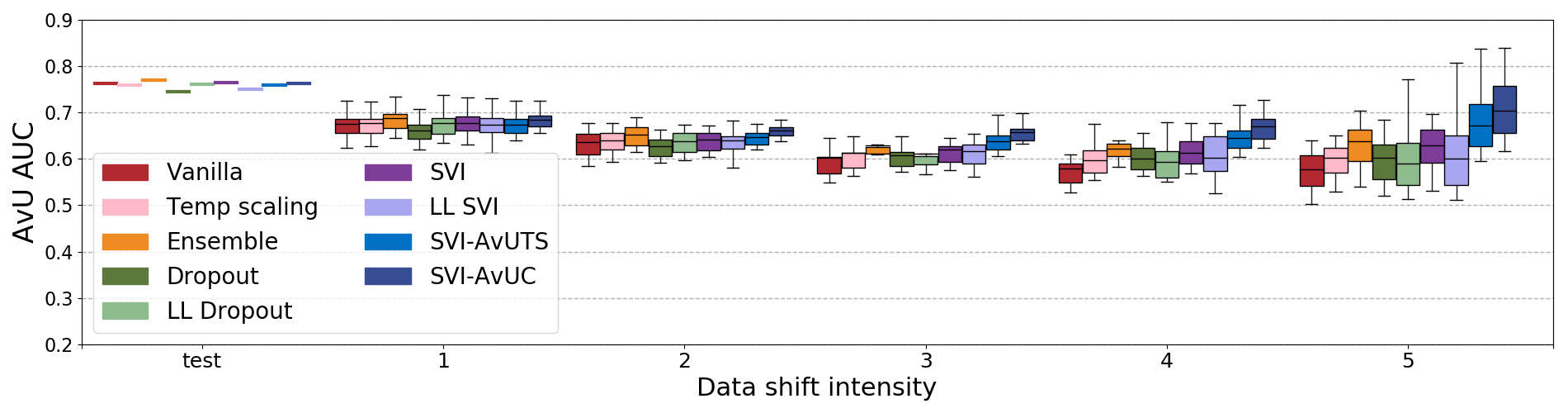}
		\label{fig:avu_imagenet}
	\end{subfigure}
	\caption{\small ImageNet: AvU AUC$\uparrow$ on in-distribution (test) and under dataset shift at different levels of shift intensities (1-5). We expect a well-calibrated model to consistently provide higher AvU AUC score even at increased levels of datashift. At each shift intensity level, the boxplot summarizes the results across 16 different datashift types showing the min, max and quartiles. SVI-AvUC and SVI-AvUTS yields higher area under the curve of AvU (AvU AUC) computed across various uncertainty thresholds at increased data shift intensity.}
	\label{fig:supp_boxplot_avu}
\end{figure*}

\begin{table}[h]
	\centering
	\caption{\small Spearman rank-order correlation coefficient assessing the relationship between AvU-AUC and data shift intensity. Spearman's $\rho$ indicates that AUC of AvU degrades with increased data shift for all the methods with comparatively SVI-AvUC being robust ($\rho$ value near to 0 is better).}
	\begin{adjustbox}{width=1\linewidth}
		\begin{tabular}{@{}cccccccccccc@{}}
			\toprule
			\multirow{2}{*}{\begin{tabular}[c]{@{}c@{}}Spearman's $\rho$\\ rank-order correlation coeff\\ wrt data shift intensity\end{tabular}} & \multirow{3}{*}{} & \multicolumn{10}{c}{Method}                                                                                                                               \\ \cmidrule(l){3-12} 
			&                   & Vanilla & \begin{tabular}[c]{@{}c@{}}Temp \\ scaling\end{tabular} & Ensemble & Dropout & LL Droput & SVI   & LL SVI & SVI-TS & SVI-AvUTS & SVI-AvUC       \\ \cmidrule(r){1-1} \cmidrule(l){3-12} 
			$\rho_{AvUAUC}$                                                                                                                        &                   & -1.0    & -0.94                                                   & -0.82    & -0.94   & -1.0      & -0.82 & -1.0   & -0.82  & -0.6      & \textbf{-0.25} \\ \bottomrule
		\end{tabular}
	\end{adjustbox}
	\label{tab:spearmanavu}
\end{table}

\subsection{Addition results for distributional shift detection}

\begin{figure}[h]
	\small
	\begin{subfigure}{0.24\textwidth}
		\centering
		\captionsetup{
			justification=centering}
		\includegraphics[scale=0.28]{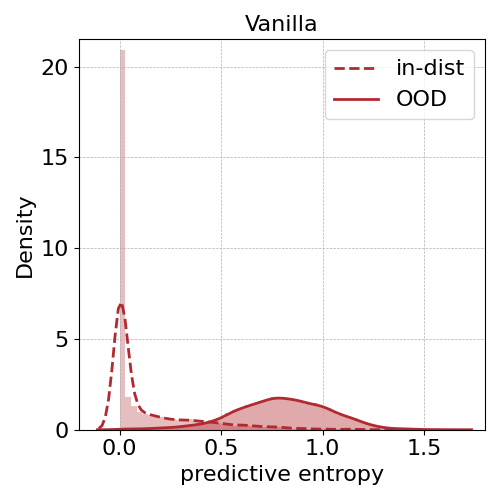}
	\end{subfigure}
	\begin{subfigure}{0.24\textwidth}
		\centering
		\captionsetup{
			justification=centering}
		\includegraphics[scale=0.28]{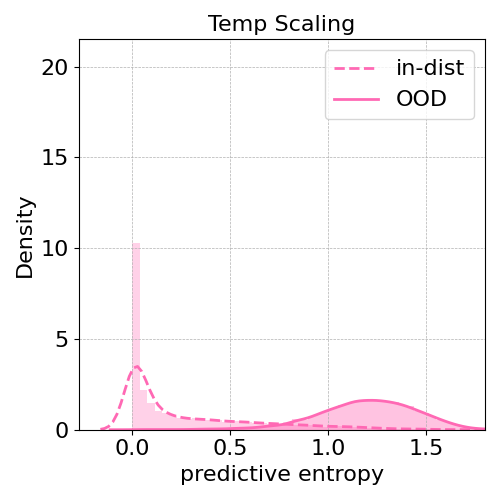}
	\end{subfigure}
	\begin{subfigure}{0.24\textwidth}
		\centering
		\captionsetup{
			justification=centering}
		\includegraphics[scale=0.28]{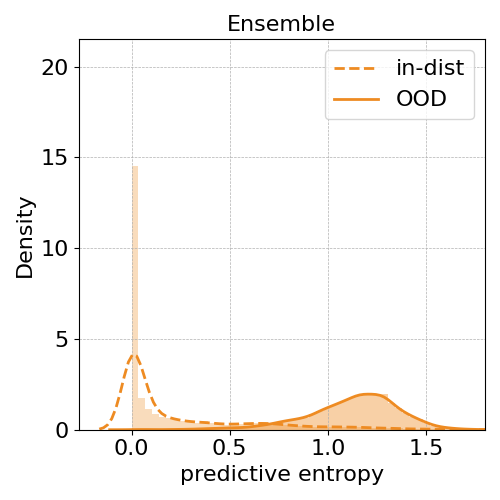}
	\end{subfigure}
	\begin{subfigure}{0.24\textwidth}
		\centering
		\captionsetup{
			justification=centering}
		\includegraphics[scale=0.28]{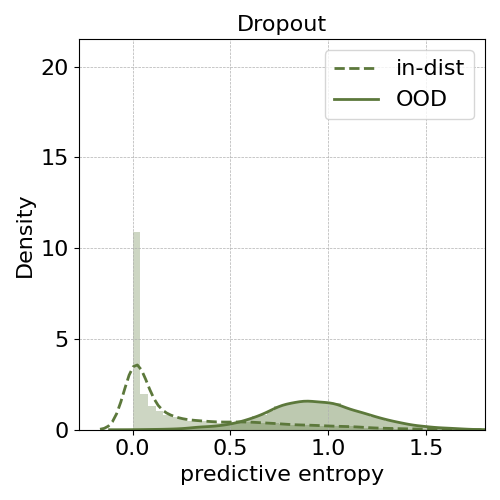}
	\end{subfigure}	
    \begin{subfigure}{0.245\textwidth}
    	\centering
    	\captionsetup{
    		justification=centering}
    	\includegraphics[scale=0.28]{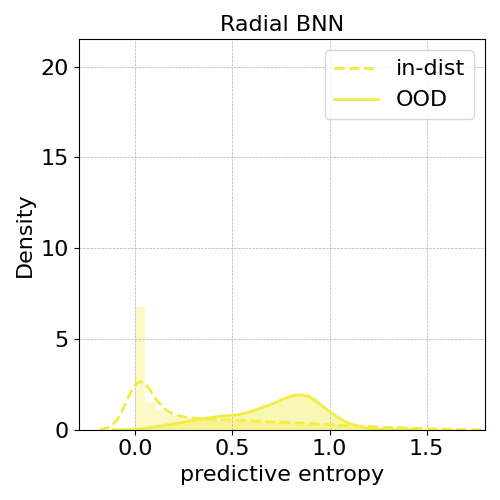}
    \end{subfigure}
	\begin{subfigure}{0.245\textwidth}
		\centering
		\captionsetup{
			justification=centering}
		\includegraphics[scale=0.28]{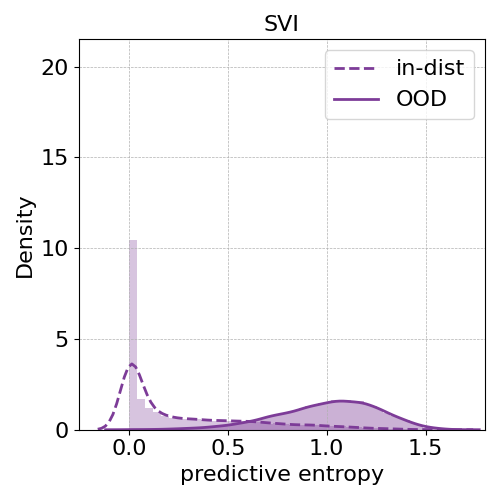}
	\end{subfigure}
	\begin{subfigure}{0.245\textwidth}
		\centering
		\captionsetup{
			justification=centering}
		\includegraphics[scale=0.28]{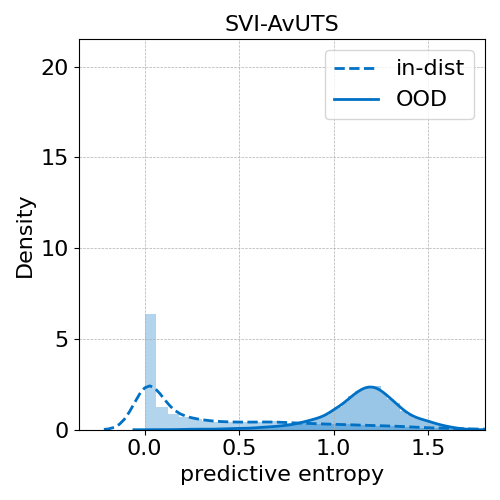}
	\end{subfigure}
	\begin{subfigure}{0.245\textwidth}
		\centering
		\captionsetup{
			justification=centering}
		\includegraphics[scale=0.28]{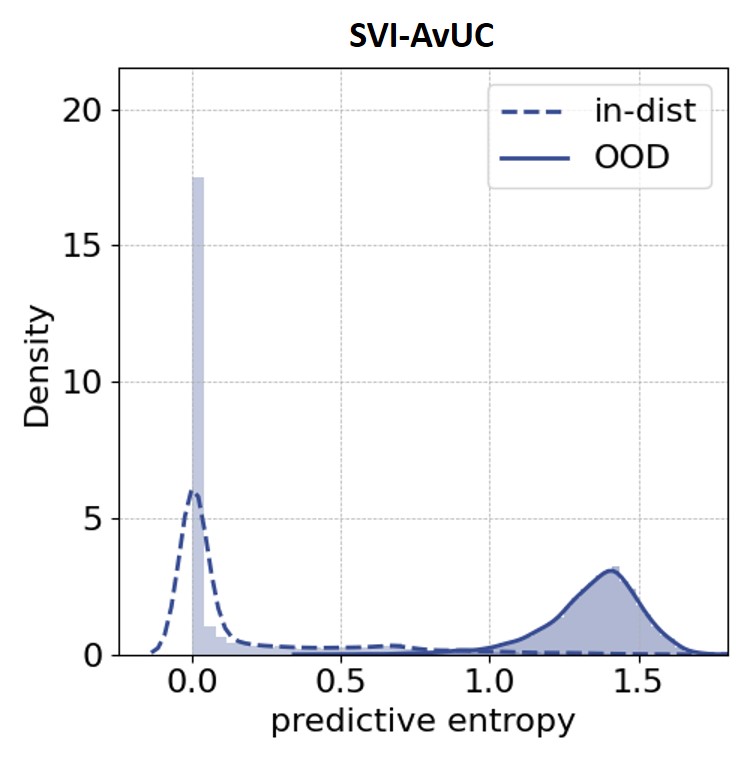}
	\end{subfigure}
	\caption{\small Out-of-distribution SVHN: Density histograms of predictive entropy on SVHN as OOD and CIFAR10 as in-distribution (ResNet-20 trained with CIFAR10). SVI-AvUC shows best separation of entropy densities between in-distribution and OOD as quantified by Wasserstein distance in Table~\ref{tab:wassdistoodcifar}.}
	\label{figappdx:oodrentropyplots}
\end{figure}

Figure~\ref{figappdx:oodrentropyplots} shows the density histogram plots of predictive uncertainty estimates obtained from different methods on SVHN dataset (out-distribution) and CIFAR10 test set (in-distribution) with ResNet-20 model that trained with CIFAR-10. These plots correspond to the out-of-distribution detection results presented in Table~\ref{tab:shiftdetection} of Section~\ref{sec:results}.
 
\begin{minipage}[b]{.45\textwidth}
	\centering
	\captionof{table}{\small Wasserstein distance between the distribution of predictive uncertainties on CIFAR10 test data (in-distribution) and SVHN data (out-of-distribution).}
	\begin{tabular}{@{}llc@{}}
		\toprule
		Method       &  & \begin{tabular}[c]{@{}c@{}}Wasserstein\\ distance\end{tabular} \\ \midrule
		Vanilla      &  & 0.6703                                                         \\
		Temp scaling &  & 0.9350                                                         \\
		Ensemble     &  & 0.9043                                                         \\
		Dropout      &  & 0.6767                                                         \\
		LL Dropout   &  & 0.4905                                                         \\
		Radial BNN   &  & 0.3933                                                         \\
		SVI          &  & 0.7480                                                         \\
		LL SVI       &  & 0.6367                                                         \\
		SVI-TS       &  & 0.7874                                                         \\
		SVI-AvUTS    &  & 0.8469                                                         \\
		SVI-AvUC     &  & \textbf{1.2021}                  \\ \bottomrule                             
	\end{tabular}
	\label{tab:wassdistoodcifar}
\end{minipage}\qquad \qquad 
\begin{minipage}[b]{.45\textwidth}
	\centering
	\captionof{table}{\small Wasserstein distance between the distribution of predictive uncertainties on ImageNet test data(in-distribution) and data shifted with defocus blur at intensity 5.}
		\begin{tabular}{@{}llc@{}}
			\toprule
			Method       &  & \begin{tabular}[c]{@{}c@{}}Wasserstein\\ distance\end{tabular} \\ \midrule
			Vanilla      &  & 3.0173                                                         \\
			Temp scaling &  & 3.1866                                                         \\
			Ensemble     &  & 3.2473                                                         \\
			Dropout      &  & 3.2605                                                         \\
			LL Dropout   &  & 3.3676                                                         \\
			SVI          &  & 3.6339                                                         \\
			LL SVI       &  & 2.9897                                                         \\
			SVI-TS       &  & 3.6851                                                         \\
			SVI-AvUTS    &  & 3.9466                                                         \\
			SVI-AvUC     &  & \textbf{4.2043}     \\ \bottomrule                                          
		\end{tabular}
		\label{tab:wassdistdefocus}
\end{minipage}

\begin{figure}[hb]
	\small
		\begin{subfigure}{0.24\textwidth}
			\centering
			\captionsetup{
				justification=centering}
		    	\includegraphics[scale=0.28]{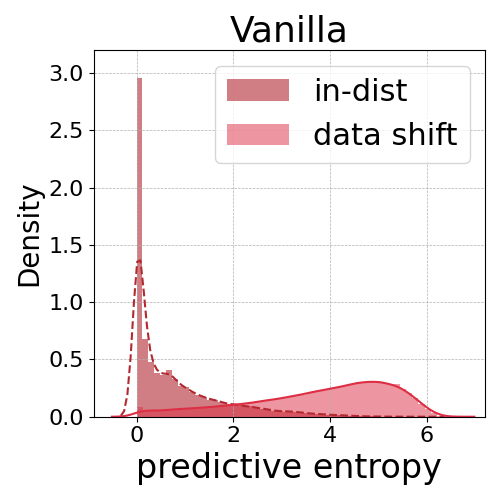}
		\end{subfigure}
		\begin{subfigure}{0.24\textwidth}
			\centering
			\captionsetup{
				justification=centering}
			\includegraphics[scale=0.28]{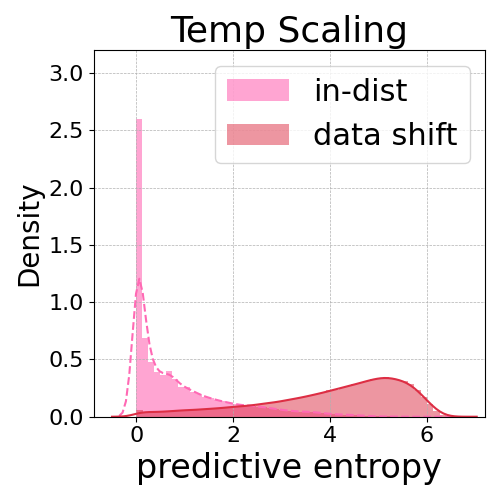}
		\end{subfigure}
	\begin{subfigure}{0.24\textwidth}
		\centering
		\captionsetup{
			justification=centering}
		\includegraphics[scale=0.28]{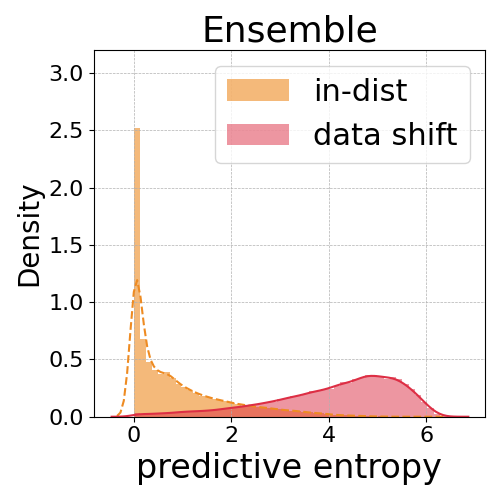}
	\end{subfigure}
    \begin{subfigure}{0.24\textwidth}
    	\centering
    	\captionsetup{
    		justification=centering}
    	\includegraphics[scale=0.28]{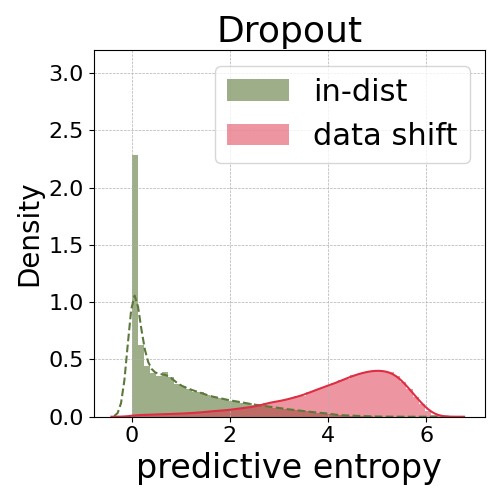}
   \end{subfigure}	
    \begin{subfigure}{0.246\textwidth}
    	\centering
    	\captionsetup{
    		justification=centering}
    	\includegraphics[scale=0.28]{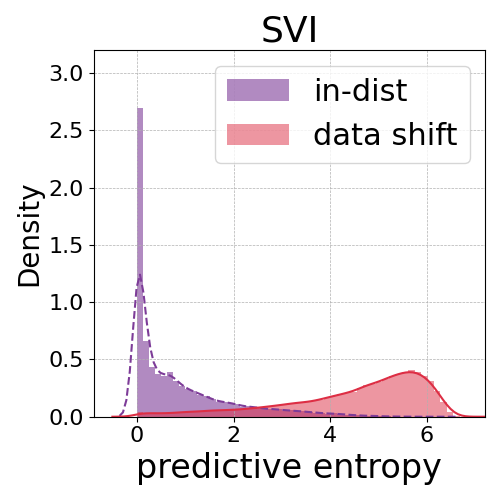}
    \end{subfigure}
     \begin{subfigure}{0.245\textwidth}
     	\centering
     	\captionsetup{
     		justification=centering}
     	\includegraphics[scale=0.28]{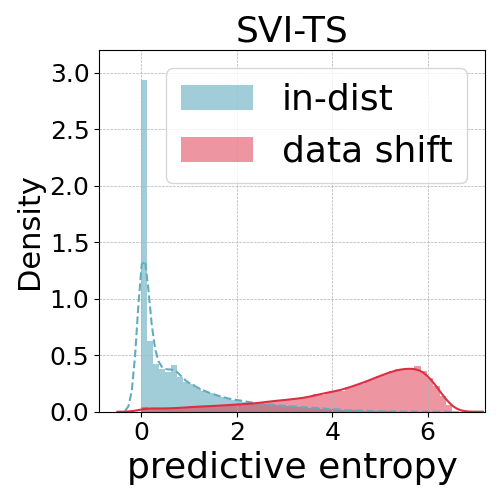}
     \end{subfigure}
     \begin{subfigure}{0.245\textwidth}
    	\centering
    	\captionsetup{
    		justification=centering}
    	\includegraphics[scale=0.28]{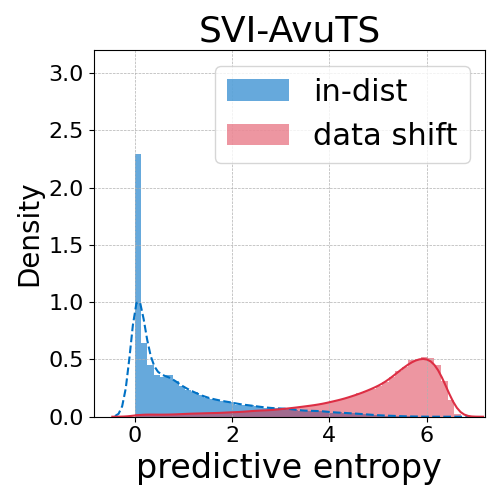}
    \end{subfigure}
    \begin{subfigure}{0.245\textwidth}
    	\centering
    	\captionsetup{
    		justification=centering}
    	\includegraphics[scale=0.28]{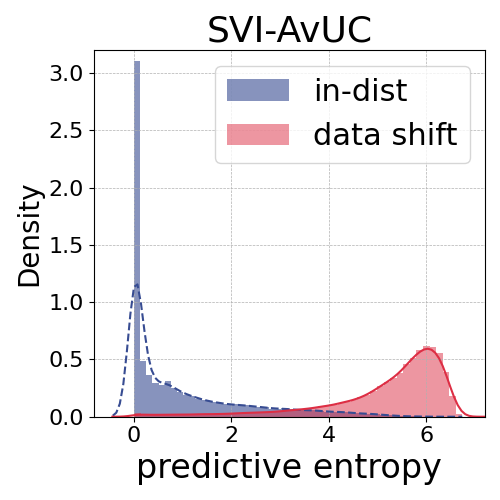}
    \end{subfigure}
	\caption{\small Data shift on ImageNet (defocus blur): Density histograms of predictive entropy on ImageNet in-distribution test set and data shifted with defocus blur (ResNet-50 trained with clean ImageNet). SVI-AvUC shows best separation of entropy densities between in-distribution and data-shift. SVI-AvUC shows best separation of predictive uncertainty densities between in-distribution and shifted data as quantified by Wasserstein distance in Table~\ref{tab:wassdistdefocus}.}
	\label{fig:defocusblurentropy}
\end{figure}

\vspace{4mm} 
Figure~\ref{fig:defocusblurentropy} shows the density histogram plots of predictive uncertainty estimates obtained from different methods on ImageNet test set (in-dist) and defocus blur of intensity 5 (data shift) with ResNet-50 model that was trained with clean ImageNet.

\vspace{4mm}
 
 Table~\ref{tab:comprehensiveshiftdetection} provides comprehensive distributional shift detection performance evaluation of different methods across 16 different types of datashift at intensity 5 on ImageNet as described in Section~\ref{appdx:datasetshift}. We observe SVI-AvUC performing best in detecting most of the shift types, and Ensemble perform best on few of the shift types.

\clearpage
\begin{table}[ht]
	\centering
	\caption{ImageNet: Distributional shift detection using predictive entropy. We compare distributional shift detection performance on 16 different types of dataset shift(each type contains 50k shifted test images). All values are in percentages and best results are indicated in bold.}
	\begin{adjustbox}{width=1\linewidth}
	\begin{tabular}{@{}llllccccccccc@{}}
		\toprule
		\multicolumn{1}{c}{\multirow{2}{*}{\begin{tabular}[c]{@{}c@{}}Dataset\\ shift type\end{tabular}}} & \multirow{2}{*}{} & \multicolumn{1}{c}{\multirow{2}{*}{\begin{tabular}[c]{@{}c@{}}Detection \\ evaluation\\ metric\;\contour{black}{$\uparrow$}\end{tabular}}} & \multirow{2}{*}{}    & \multicolumn{9}{c}{Methods}                                                                                                                                  \\ \cmidrule(l){5-13} 
		\multicolumn{1}{c}{}                                                                              &                   & \multicolumn{1}{c}{}                                                                                          &                      & Vanilla & \begin{tabular}[c]{@{}c@{}}Temp \\ scaling\end{tabular} & Ensemble       & Dropout & LL Dropout & SVI   & LL SVI & SVI-AvUTS      & SVI-AvUC       \\ \midrule
		\multirow{4}{*}{\textbf{\begin{tabular}[c]{@{}l@{}}Gaussian\\ blur\end{tabular}}}                 &                   & AUROC                                                                                                         &                      & 93.36   & 93.71                                                   & 95.49          & 96.38   & 96.04      & 96.40 & 93.58  & 96.89          & \textbf{97.60} \\  
		&                   & Det. accuracy                                                                                                 &                      & 86.08   & 86.47                                                   & 88.82          & 89.98   & 89.68      & 90.03 & 86.93  & 90.93          & \textbf{92.07} \\
		&                   & AUPR-in                                                                                                       &                      & 92.82   & 93.21                                                   & 95.31          & 96.16   & 95.63      & 95.97 & 92.06  & 96.58          & \textbf{97.39} \\
		&                   & AUPR-out                                                                                                      &                      & 93.71   & 94.01                                                   & 95.64          & 96.67   & 96.40      & 96.83 & 94.02  & 97.19          & \textbf{97.85} \\ \midrule
		\multirow{4}{*}{\textbf{Brightness}}                                                              &                   & AUROC                                                                                                         &                      & 70.58   & 71.02                                                   & 71.97          & 73.73   & 71.17      & 72.77 & 69.24  & \textbf{75.08} & 74.61          \\ 
		&                   & Det. accuracy                                                                                                 &                      & 65.03   & 65.45                                                   & 66.15          & 67.44   & 65.36      & 66.61 & 64.16  & \textbf{68.44} & 67.58          \\
		&                   & AUPR-in                                                                                                       &                      & 68.28   & 68.62                                                   & 70.57          & 72.42   & 68.96      & 70.93 & 65.41  & \textbf{73.12} & 73.54          \\
		&                   & AUPR-out                                                                                                      &                      & 70.80   & 71.26                                                   & 71.62          & 73.75   & 71.48      & 73.34 & 69.60  & \textbf{75.93} & 75.56          \\ \midrule
		\multirow{4}{*}{\textbf{Contrast}}                                                                &                   & AUROC                                                                                                         &                      & 98.82   & 98.96                                                   & 99.40          & 99.41   & 99.32      & 98.92 & 98.73  & 99.45          & \textbf{99.48} \\  
		&                   & Det. accuracy                                                                                                 &                      & 94.70   & 95.06                                                   & 96.27          & 96.22   & 96.06      & 94.87 & 94.59  & 96.52          & \textbf{96.69} \\
		&                   & AUPR-in                                                                                                       &                      & 98.75   & 98.91                                                   & 99.39          & 99.41   & 99.28      & 98.85 & 98.64  & 99.44          & \textbf{99.46} \\
		&                   & AUPR-out                                                                                                      &                      & 98.91   & 99.04                                                   & 99.42          & 99.43   & 99.37      & 99.02 & 98.85  & 99.48          & \textbf{99.52} \\ \midrule
		\multirow{4}{*}{\textbf{\begin{tabular}[c]{@{}l@{}}Defocus\\ blur\end{tabular}}}                  &                   & AUROC                                                                                                         &                      & 94.04   & 94.37                                                   & 95.74          & 96.26   & 95.97      & 95.88 & 93.69  & 96.68          & \textbf{97.18} \\ 
		&                   & Det. accuracy                                                                                                 &                      & 86.79   & 87.13                                                   & 89.06          & 89.79   & 89.52      & 89.35 & 86.98  & 90.51          & \textbf{91.40} \\
		&                   & AUPR-in                                                                                                       &                      & 93.34   & 93.70                                                   & 95.40          & 96.03   & 95.44      & 95.37 & 92.01  & 96.29          & \textbf{96.91} \\
		&                   & AUPR-out                                                                                                      &                      & 94.66   & 94.94                                                   & 96.11          & 96.58   & 96.43      & 96.39 & 94.28  & 97.05          & \textbf{97.50} \\ \midrule
		\multirow{4}{*}{\textbf{\begin{tabular}[c]{@{}l@{}}Elastic\\ transform\end{tabular}}}             &                   & AUROC                                                                                                         &                      & 88.15   & 88.81                                                   & \textbf{91.03} & 87.73   & 89.20      & 89.63 & 86.73  & 90.84          & 90.82          \\ 
		&                   & Det. accuracy                                                                                                 &                      & 80.43   & 81.16                                                   & \textbf{83.59} & 80.19   & 81.69      & 82.12 & 79.44  & 83.28          & 83.06          \\
		&                   & AUPR-in                                                                                                       &                      & 88.45   & 89.10                                                   & \textbf{91.43} & 88.57   & 89.56      & 89.99 & 86.08  & 91.08          & 91.29          \\
		&                   & AUPR-out                                                                                                      &                      & 87.18   & 87.84                                                   & 90.06          & 85.97   & 88.01      & 88.58 & 86.19  & 90.07          & \textbf{90.08} \\ \midrule
		\multirow{4}{*}{\textbf{Fog}}                                                                     &                   & AUROC                                                                                                         &                      & 89.15   & 89.74                                                   & 91.45          & 91.83   & 90.03      & 90.20 & 87.40  & \textbf{93.01} & 91.46          \\  
		&                   & Det. accuracy                                                                                                 &                      & 81.12   & 81.79                                                   & 83.78          & 84.00   & 82.03      & 82.48 & 79.75  & \textbf{85.47} & 83.44          \\
		&                   & AUPR-in                                                                                                       &                      & 88.75   & 89.30                                                   & 91.39          & 92.04   & 89.85      & 89.90 & 85.78  & \textbf{92.84} & 91.13          \\
		&                   & AUPR-out                                                                                                      &                      & 89.22   & 89.83                                                   & 91.34          & 91.61   & 89.99      & 90.26 & 87.67  & \textbf{93.14} & 91.90          \\ \midrule
		\multirow{4}{*}{\textbf{Frost}}                                                                   &                   & AUROC                                                                                                         &                      & 88.67   & 89.19                                                   & 90.90          & 90.53   & 88.56      & 90.60 & 87.69  & 91.74          & \textbf{92.19} \\ 
		&                   & Det. accuracy                                                                                                 &                      & 80.87   & 81.40                                                   & 83.23          & 82.64   & 80.65      & 82.84 & 80.07  & 83.99          & \textbf{84.31} \\
		&                   & AUPR-in                                                                                                       &                      & 87.95   & 88.46                                                   & 90.56          & 90.44   & 87.80      & 89.91 & 86.09  & 91.03          & \textbf{91.63} \\
		&                   & AUPR-out                                                                                                      &                      & 89.03   & 89.55                                                   & 91.06          & 90.56   & 88.98      & 91.20 & 88.10  & 92.41          & \textbf{92.88} \\ \midrule
		\multirow{4}{*}{\textbf{\begin{tabular}[c]{@{}l@{}}Glass\\ blur\end{tabular}}}                    &                   & AUROC                                                                                                         &                      & 94.96   & 95.29                                                   & 96.48          & 96.06   & 96.02      & 96.90 & 95.14  & 97.37          & \textbf{97.85} \\ 
		&                   & Det. accuracy                                                                                                 &                      & 87.86   & 88.31                                                   & 90.15          & 89.41   & 89.40      & 90.71 & 88.68  & 91.58          & \textbf{92.51} \\
		&                   & AUPR-in                                                                                                       &                      & 94.71   & 95.06                                                   & 96.32          & 95.94   & 95.76      & 96.68 & 94.28  & 97.20          & \textbf{97.70} \\
		&                   & AUPR-out                                                                                                      &                      & 95.24   & 95.54                                                   & 96.66          & 96.24   & 96.26      & 97.17 & 95.52  & 97.57          & \textbf{98.05} \\ \midrule
		\multirow{4}{*}{\textbf{\begin{tabular}[c]{@{}l@{}}Gaussian \\ noise\end{tabular}}}               &                   & AUROC                                                                                                         &                      & 92.36   & 92.84                                                   & \textbf{97.78} & 91.27   & 93.87      & 95.83 & 91.00  & 96.37          & 97.46          \\  
		&                   & Det. accuracy                                                                                                 &                      & 85.25   & 85.92                                                   & \textbf{92.92} & 85.84   & 87.31      & 89.29 & 84.60  & 90.10          & 91.73          \\
		&                   & AUPR-in                                                                                                       &                      & 92.66   & 93.16                                                   & \textbf{97.97} & 93.39   & 94.60      & 95.91 & 91.75  & 96.44          & 97.46          \\
		&                   & AUPR-out                                                                                                      &                      & 91.20   & 91.62                                                   & 97.42          & 86.10   & 92.70      & 95.76 & 89.03  & 96.28          & \textbf{97.52} \\ \midrule
		\multirow{4}{*}{\textbf{\begin{tabular}[c]{@{}l@{}}Impulse\\ noise\end{tabular}}}                 &                   & AUROC                                                                                                         &                      & 92.15   & 92.63                                                   & \textbf{97.64} & 92.10   & 93.77      & 95.39 & 91.68  & 96.01          & 97.14          \\  
		&                   & Det. accuracy                                                                                                 &                      & 85.03   & 85.69                                                   & \textbf{92.76} & 86.81   & 87.10      & 88.73 & 85.04  & 89.56          & 91.17          \\
		&                   & AUPR-in                                                                                                       &                      & 92.59   & 93.09                                                   & \textbf{97.91} & 94.01   & 94.44      & 95.51 & 92.27  & 96.10          & 97.20          \\
		&                   & AUPR-out                                                                                                      &                      & 90.75   & 91.17                                                   & 97.15          & 86.95   & 92.67      & 95.25 & 90.05  & 95.87          & \textbf{97.17} \\ \midrule
		\multirow{4}{*}{\textbf{Pixelate}}                                                                &                   & AUROC                                                                                                         &                      & 81.52   & 81.88                                                   & 87.80          & 88.03   & 87.01      & 87.98 & 79.85  & 87.19          & \textbf{90.04} \\  
		&                   & Det. accuracy                                                                                                 &                      & 74.37   & 74.71                                                   & 80.23          & 80.64   & 79.50      & 80.24 & 73.17  & 79.27          & \textbf{81.98} \\
		&                   & AUPR-in                                                                                                       &                      & 80.02   & 80.39                                                   & 87.16          & 87.94   & 86.03      & 87.07 & 76.98  & 86.10          & \textbf{89.48} \\
		&                   & AUPR-out                                                                                                      &                      & 81.34   & 81.66                                                   & 87.56          & 86.91   & 86.93      & 88.27 & 79.56  & 87.52          & \textbf{90.48} \\ \midrule
		\multirow{4}{*}{\textbf{Saturate}}                                                                &                   & \multicolumn{1}{c}{AUROC}                                                                                     & \multicolumn{1}{c}{} & 74.37   & 74.83                                                   & 76.70          & 75.70   & 74.19      & 77.21 & 73.26  & 78.05          & \textbf{78.71} \\ 
		&                   & \multicolumn{1}{c}{Det. accuracy}                                                                             & \multicolumn{1}{c}{} & 68.32   & 68.79                                                   & 70.37          & 69.60   & 68.22      & 70.65 & 67.52  & 71.41          & \textbf{71.57} \\
		&                   & \multicolumn{1}{c}{AUPR-in}                                                                                   & \multicolumn{1}{c}{} & 71.66   & 72.04                                                   & 74.53          & 74.21   & 71.54      & 74.95 & 69.38  & 75.84          & \textbf{77.31} \\
		&                   & \multicolumn{1}{c}{AUPR-out}                                                                                  & \multicolumn{1}{c}{} & 73.84   & 74.29                                                   & 75.90          & 73.75   & 73.24      & 77.07 & 72.94  & 77.76          & \textbf{78.82} \\ \midrule
		\multirow{4}{*}{\textbf{\begin{tabular}[c]{@{}l@{}}Shot\\ noise\end{tabular}}}                    &                   & AUROC                                                                                                         &                      & 90.38   & 90.92                                                   & \textbf{97.15} & 90.31   & 93.25      & 95.17 & 90.29  & 95.57          & 96.72          \\  
		&                   & Det. accuracy                                                                                                 &                      & 83.11   & 83.79                                                   & \textbf{91.98} & 84.88   & 86.74      & 88.49 & 84.02  & 89.02          & 90.47          \\
		&                   & AUPR-in                                                                                                       &                      & 90.72   & 91.29                                                   & \textbf{97.41} & 92.55   & 94.07      & 95.27 & 91.03  & 95.68          & 96.77          \\
		&                   & AUPR-out                                                                                                      &                      & 88.86   & 89.32                                                   & 96.53          & 84.79   & 91.68      & 95.00 & 87.45  & 95.38          & \textbf{96.75} \\ \midrule
		\multirow{4}{*}{\textbf{Spatter}}                                                                 &                   & AUROC                                                                                                         &                      & 84.23   & 84.92                                                   & \textbf{88.01} & 84.87   & 84.60      & 86.01 & 83.41  & 87.00          & 86.34          \\  
		&                   & Det. accuracy                                                                                                 &                      & 76.74   & 77.49                                                   & \textbf{80.78} & 77.95   & 77.66      & 78.75 & 76.15  & 79.53          & 78.73          \\
		&                   & AUPR-in                                                                                                       &                      & 84.06   & 84.69                                                   & \textbf{88.05} & 85.73   & 84.66      & 85.81 & 81.38  & 86.59          & 86.61          \\
		&                   & AUPR-out                                                                                                      &                      & 82.90   & 83.62                                                   & \textbf{86.66} & 81.50   & 82.38      & 84.90 & 83.21  & 86.36          & 85.12          \\ \midrule
		\multirow{4}{*}{\textbf{\begin{tabular}[c]{@{}l@{}}Speckle \\ noise\end{tabular}}}                &                   & AUROC                                                                                                         &                      & 87.32   & 87.83                                                   & \textbf{93.17} & 88.54   & 88.54      & 90.28 & 87.13  & 90.58          & 91.84          \\  
		&                   & Det. accuracy                                                                                                 &                      & 80.05   & 80.64                                                   & \textbf{86.40} & 82.00   & 82.00      & 82.87 & 80.39  & 83.10          & 84.02          \\
		&                   & AUPR-in                                                                                                       &                      & 87.57   & 88.09                                                   & \textbf{93.32} & 89.88   & 89.88      & 90.17 & 86.89  & 90.41          & 91.88          \\
		&                   & AUPR-out                                                                                                      &                      & 85.38   & 85.84                                                   & \textbf{92.25} & 84.48   & 84.48      & 89.78 & 84.78  & 90.12          & 91.70          \\ \midrule
		\multirow{4}{*}{\textbf{\begin{tabular}[c]{@{}l@{}}Zoom\\ blur\end{tabular}}}                     &                   & AUROC                                                                                                         &                      & 89.92   & 90.48                                                   & 92.12          & 90.47   & 90.77      & 90.65 & 88.65  & 91.56          & \textbf{93.87} \\  
		&                   & Det. accuracy                                                                                                 &                      & 82.29   & 82.92                                                   & 84.79          & 82.85   & 83.32      & 83.11 & 81.36  & 84.14          & \textbf{86.62} \\
		&                   & AUPR-in                                                                                                       &                      & 88.84   & 89.40                                                   & 91.49          & 90.18   & 89.86      & 89.82 & 86.31  & 91.01          & \textbf{93.41} \\
		&                   & AUPR-out                                                                                                      &                      & 90.39   & 90.93                                                   & 92.36          & 90.27   & 91.07      & 91.04 & 89.01  & 91.68          & \textbf{94.40} \\ \bottomrule
	\end{tabular}
	\end{adjustbox}
	\label{tab:comprehensiveshiftdetection}
\end{table}

\clearpage
\subsection{AvUTS applied to Vanilla DNN (Comparison with Temp scaling using NLL)}
\label{appdx:avutssec}
We evaluate AvUTS (AvU Temperature Scaling) by performing post-hoc calibration on vanilla DNN with \textit{accuracy versus uncertainty calibration} (AvUC) loss and compare with conventional  temperature scaling~\cite{guo2017calibration} that optimizes negative log-likelihood loss. We use entropy of softmax as uncertainty for AvUC loss computation. 
\begin{figure*}[h]
	\small
	\centering
	\begin{subfigure}{\textwidth}
		\centering
		\includegraphics[width=0.98\linewidth]{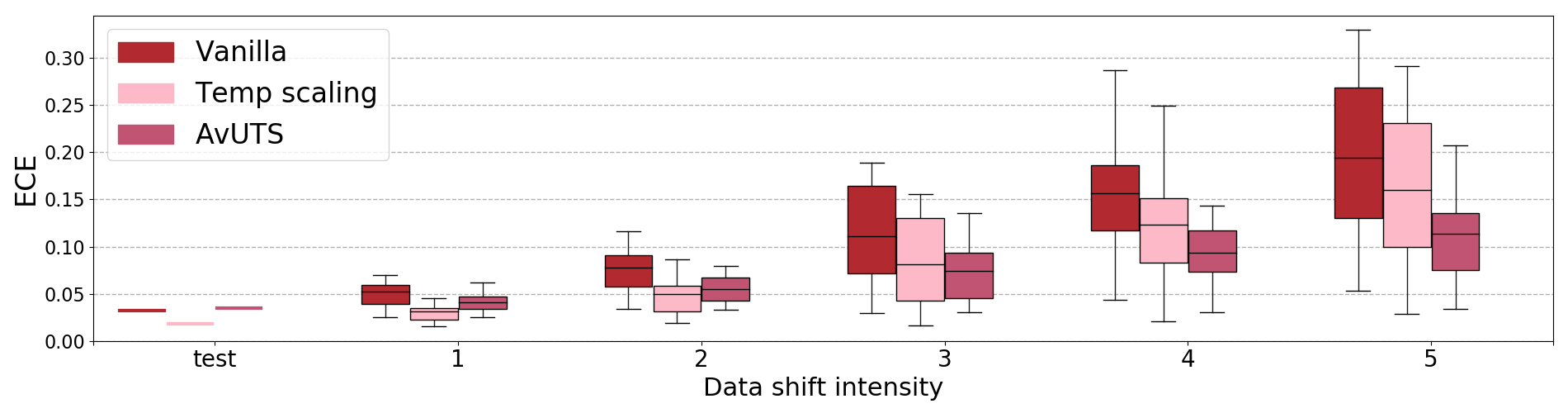}
	\end{subfigure}
	\begin{subfigure}{\textwidth}
		\centering
		\includegraphics[width=0.98\linewidth]{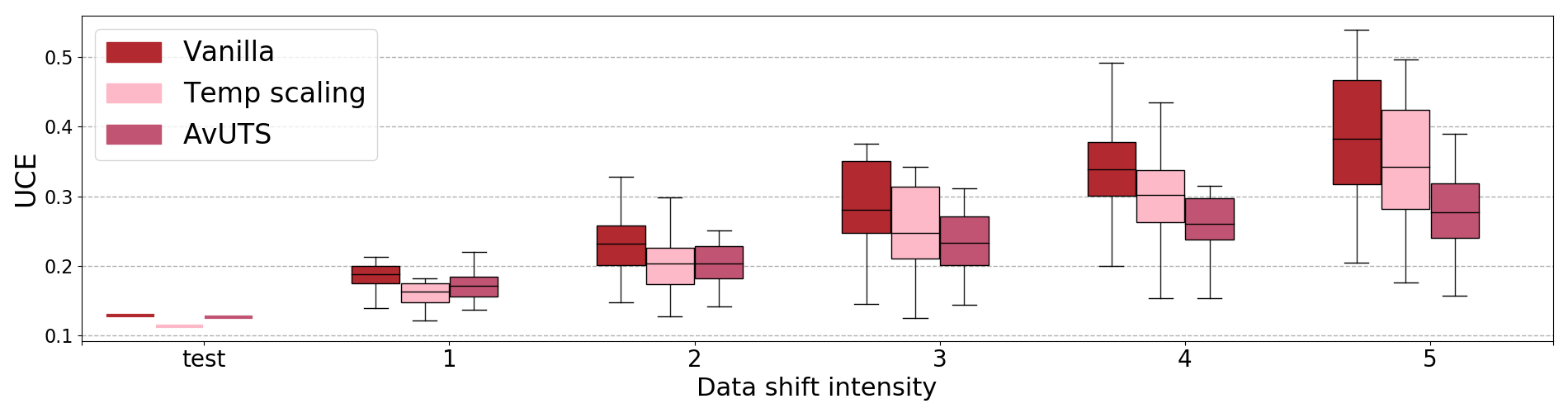}
	\end{subfigure}
	\begin{subfigure}{\textwidth}
		\centering
		\includegraphics[width=0.98\linewidth]{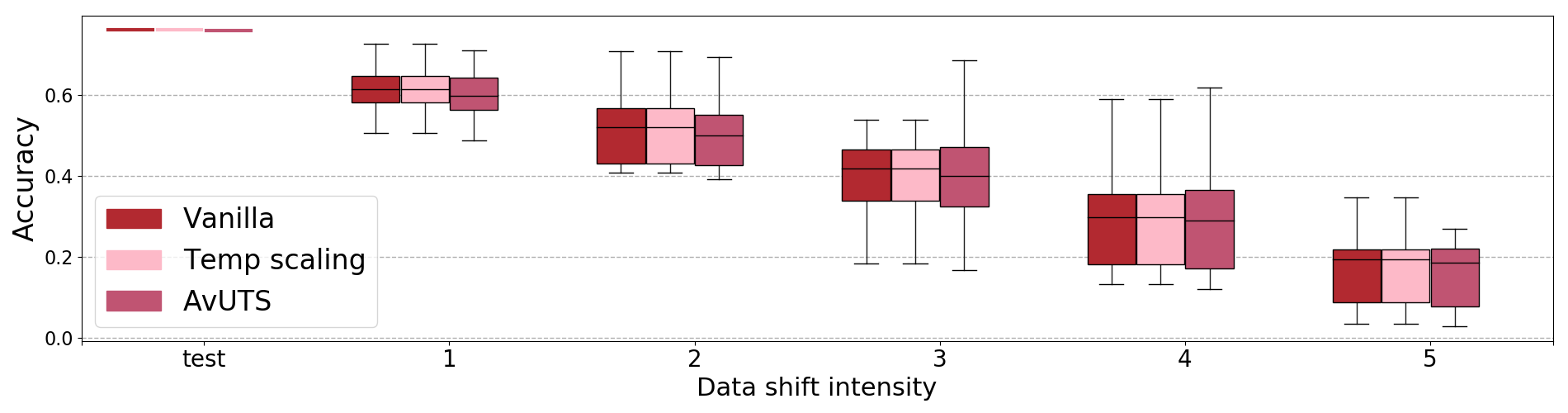}
	\end{subfigure}
	\caption{AvUTS on Vanilla ResNet-50: Model calibration comparison of AvUTS with conventional Temp Scaling and Vanilla baselines using ECE$\downarrow$ and UCE$\downarrow$ on ImageNet under in-distribution (test) and dataset shift at different levels of shift intensities (1-5). A well-calibrated model should provide lower calibration errors even at increased levels of datashift, though accuracy may degrade with data shift. At each shift intensity level, the boxplot summarizes the results across 16 different datashift types showing the min, max and quartiles. We can see that AvUTS provides significantly lower model calibration errors (ECE and UCE) than Vanilla and Temp scaling methods at increased distributional shift intensity, while providing comparable accuracy.}
	\label{fig:supp_boxplot_avuts}
\end{figure*}

\vspace{-6mm}
\section{Ablation study for $\beta$ weight factor in SVI-AvUC}
\begin{figure}[h]
	\vspace{-2mm}
	\centering
	\begin{subfigure}[b]{\textwidth}
		\centering
		\includegraphics[width=0.42\linewidth]{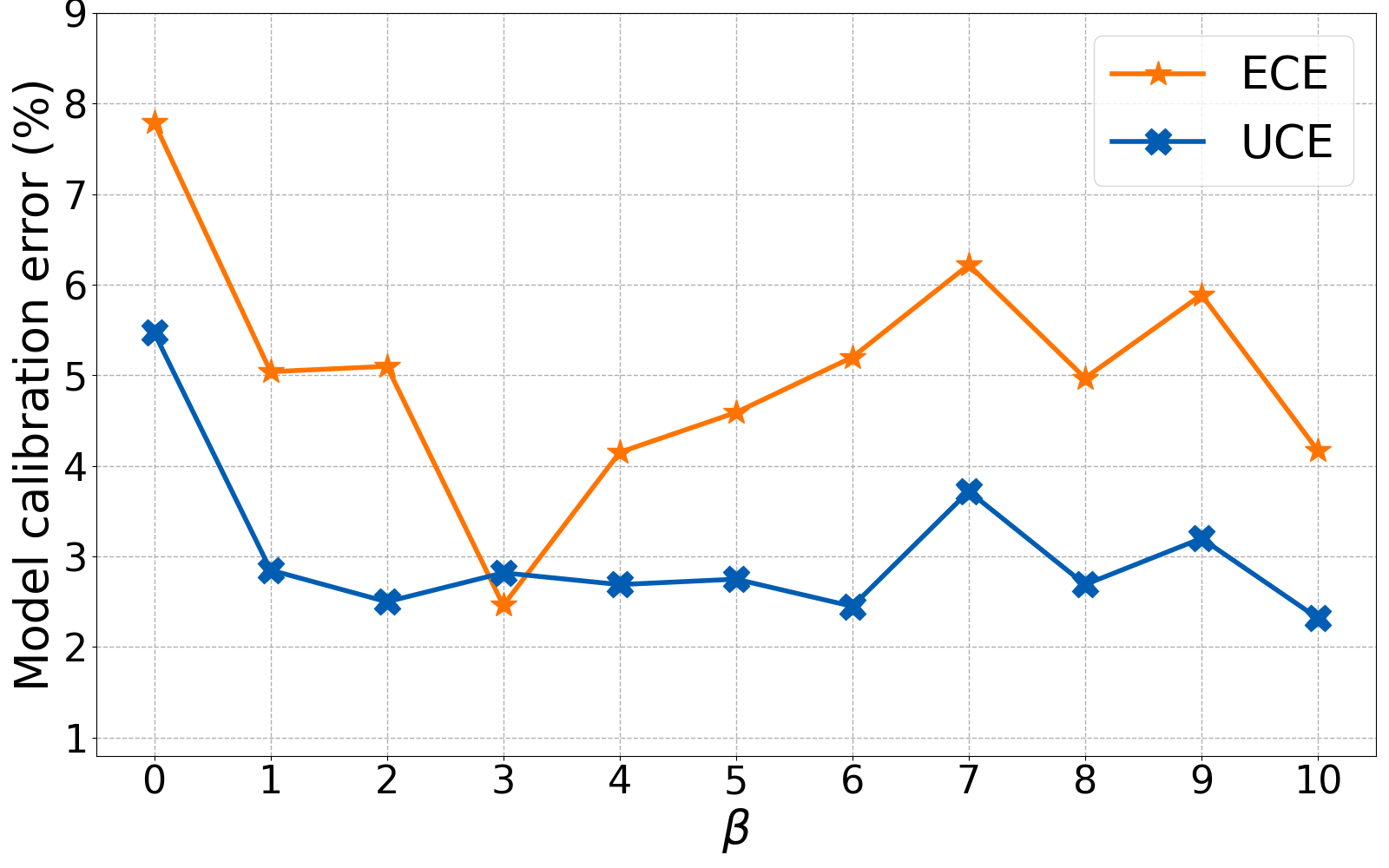}
		\hfill
		\includegraphics[width=0.42\linewidth]{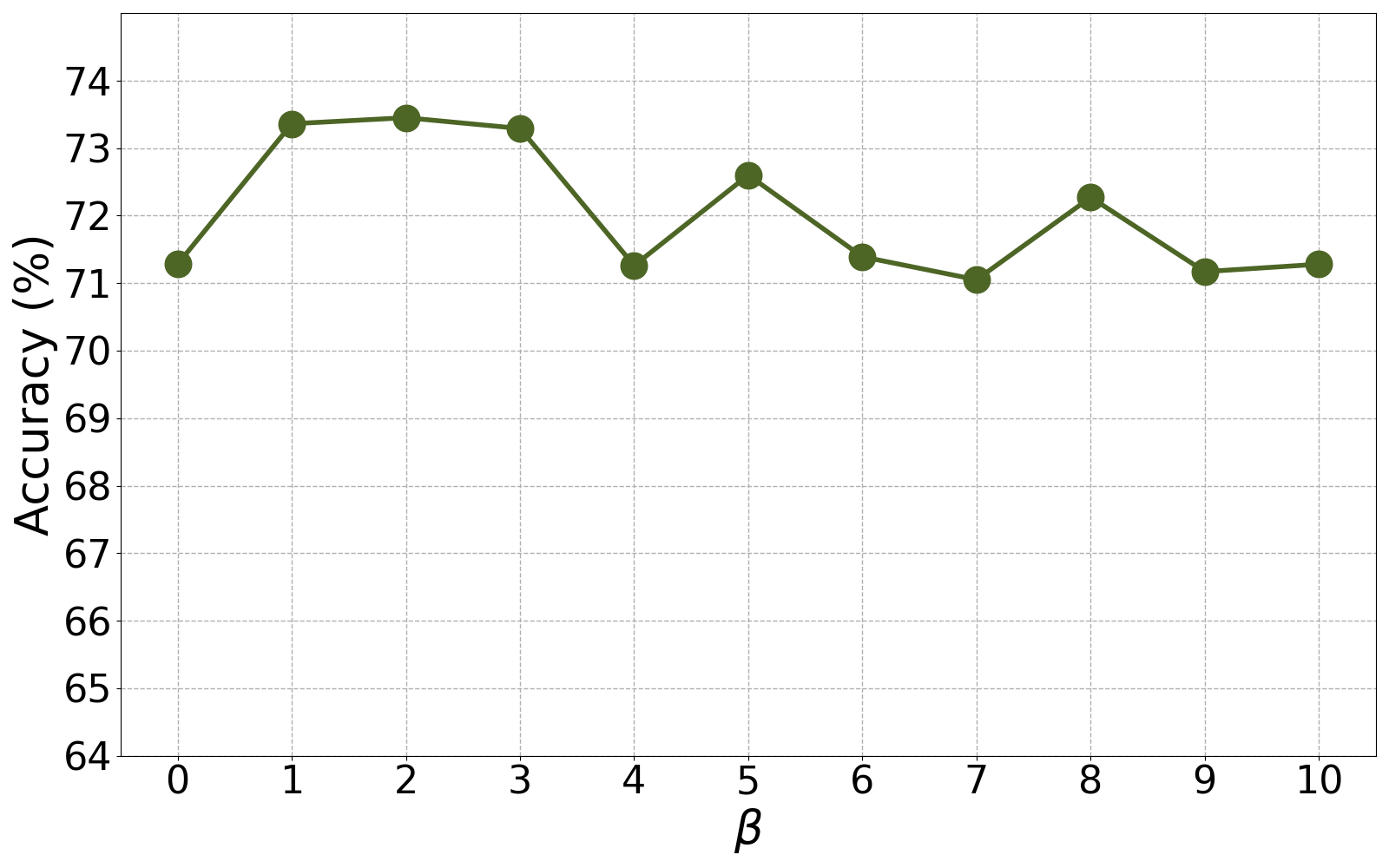}
		\hfill
	\end{subfigure}
	\caption{\small Model calibration errors (ECE, UCE) and accuracy at different values of $\beta$ in Equation~\ref{eqn:total_loss}}
	\label{fig:ablation}
\end{figure}

We evaluate SVI-AvUC method on ResNet-20 model with different values of $\beta$ in Equation~\ref{eqn:total_loss}. Figure~\ref{fig:ablation} shows the effect of different values of $\beta$ on the model calibration errors (ECE and UCE) and model test accuracy on test data shifted with Gaussian blur at intensity 3. We observe that the accuracy curve remains almost flat with different $\beta$ values, ECE decreases initially and increases above $\beta$=3, UCE decreases initially with $\beta$ and then remains almost flat.

\section{Optimizing Area under the curve of AvU}
\label{appdx:auavuc}
\begin{figure*}[h]
	\small
	\begin{subfigure}{\textwidth}
		\centering
		\includegraphics[width=1\linewidth]{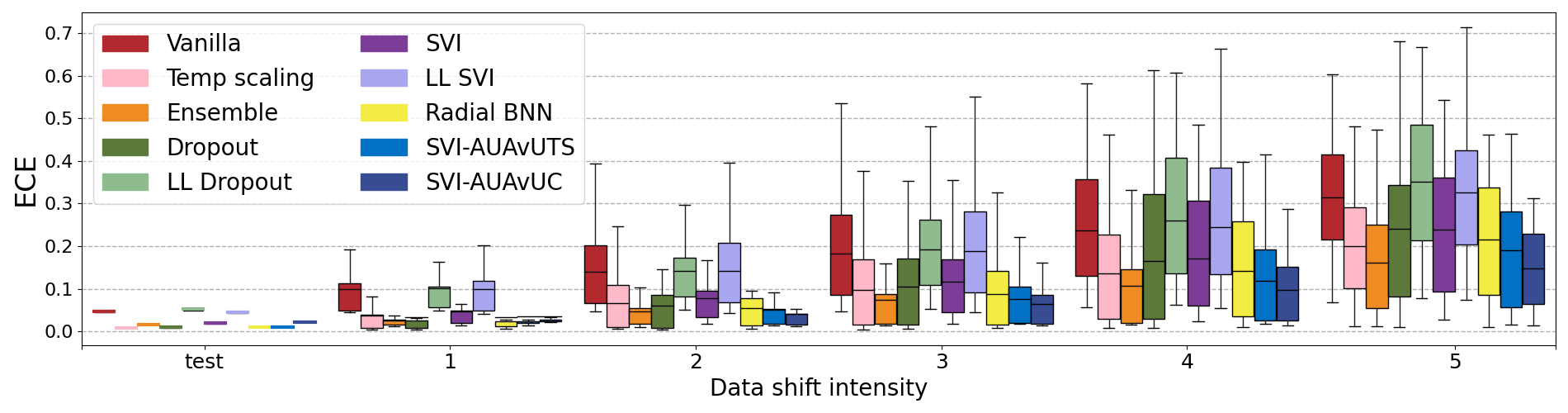}
	\end{subfigure}
	\begin{subfigure}{\textwidth}
		\centering
		\includegraphics[width=1\linewidth]{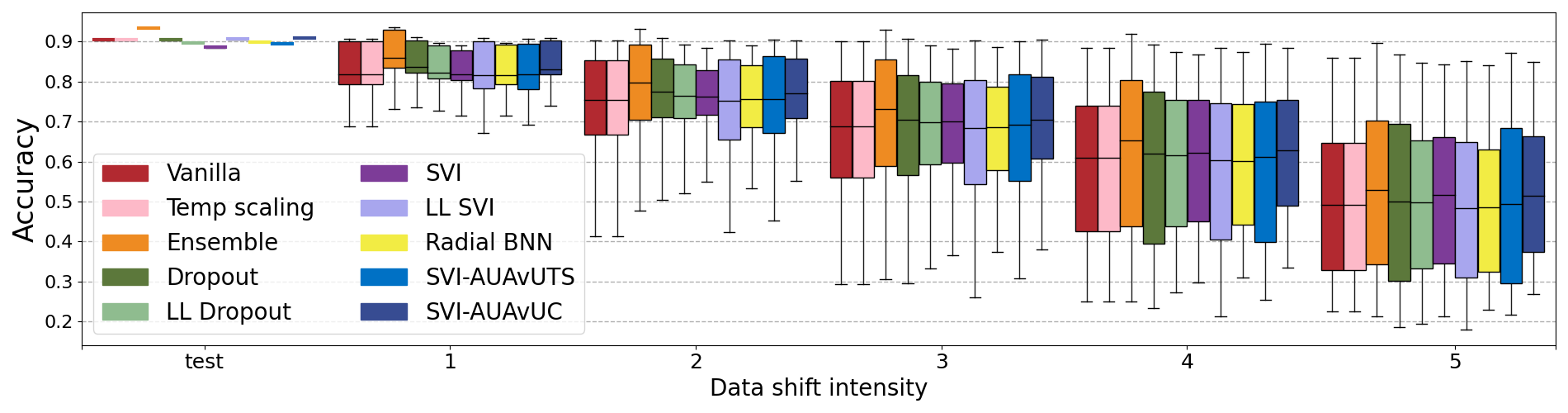}
	\end{subfigure}
	\caption{AUC of AvU optimized ResNet-20/CIFAR10: Model calibration comparison using ECE$\downarrow$ on CIFAR10 under in-distribution (test) and dataset shift at different levels of shift intensities (1-5). A well-calibrated model should consistently provide lower calibration error even at increased levels of datashift, though accuracy may degrade with increased datashift. At each shift intensity level, the boxplot summarizes the results across 16 different datashift types showing the min, max and quartiles.}
	\label{fig:supp_boxplot_auavu}
\end{figure*}

We optimized area under the curve of AvU across various uncertainty thresholds towards a threshold free mechanism. This method is compute intensive during training as we need to compute AvU at different thresholds \( u_{t h}=u_{\min }+\left(t\left(u_{\max }-u_{\min }\right)\right) \) with $t\in[0,1]$. We applied this method to both training the model and post-hoc calibration on SVI (SVI-AUAvUC and SVI-AUAvUTS), results are shown in Figure~\ref{fig:supp_boxplot_auavu}. The results are similar to SVI-AvUC and SVI-AvUTS as presented in Figure~\ref{fig:supp_boxplot_cifar}.

\end{document}